\documentclass[pdflatex,sn-nature]{sn-jnl}

\usepackage{graphicx}
\usepackage{multirow}
\usepackage{amsmath,amssymb,amsfonts}
\usepackage{amsthm}
\usepackage{mathrsfs}
\usepackage[title]{appendix}
\usepackage{xcolor}
\usepackage{textcomp}
\usepackage{booktabs}
\usepackage{url}
\usepackage{hyperref}

\begin{document}

\title[SonoBase]{Open ultrasound foundation model for robust segmentation and clinical measurement across heterogeneous settings}

\author*[1]{\fnm{Chao} \sur{Qin}}\email{chao.qin@mbzuai.ac.ae}
\author[1]{\fnm{Fahad Shahbaz} \sur{Khan}}\email{fahad.khan@mbzuai.ac.ae}
\author[1]{\fnm{Salman} \sur{Khan}}\email{salman.khan@mbzuai.ac.ae}
\author[2]{\fnm{Sarim} \sur{Ather}}\email{sather@seha.ae}
\author[3,4]{\fnm{Siddiq} \sur{Anwar}}\email{siddiq.anwar@kch.ae}
\author[1]{\fnm{Rao Muhammad} \sur{Anwer}}\email{rao.anwer@mbzuai.ac.ae}
\author*[4]{\fnm{Shadab} \sur{Khan}}\email{shadab@alumni.harvard.edu}

\affil[1]{\orgname{Mohamed bin Zayed Uni. of Artificial Intelligence}, \orgaddress{\city{Abu Dhabi}, \country{UAE}}}

\affil[2]{\orgname{Sheikh Tahnoon Bin Mohammed Medical City (STMC)}, \orgaddress{\city{Al Ain}, \country{UAE}}}

\affil[3]{\orgname{King's College Hospital London - Dubai}, \orgaddress{\city{Dubai}, \country{UAE}}}

\affil[4]{\orgname{ADIA Lab}, \orgaddress{\city{Abu Dhabi}, \country{UAE}}}

\abstract{
Ultrasound is the most widely deployed imaging modality worldwide, yet clinical AI remains fragmented into narrow single-task models that fail when device, operator, or anatomy changes. Here we present SonoCorpus, an open resource unifying 456,963 images and 1,626,085 expert masks from 53 public datasets spanning 24 clinical applications and 17 countries, and SonoBase, an interactive segmentation foundation model pretrained on it. Across fifteen evaluation datasets introducing new organs, devices, operators, and geographies, SonoBase outperforms SAM2, MedSAM2, and the concept-promptable MedSAM3 on every dataset and matches per-dataset specialist models trained on the same data; on fully external data it exceeds the accuracy these baselines achieve on their own in-distribution benchmarks. Ejection fraction derived from its segmentations falls within inter-observer variability (6.63\% error), with fewer misclassifications at the defibrillator-candidacy threshold than either promptable baseline (13\% versus 18--42\%); fetal head-circumference (1.81~mm) and gestational-age (1.2 days) errors fall below inter-observer variability. Where a baseline fails outright, one in four test cases, SonoBase recovers a usable segmentation in 81\% of them, including on handheld probes operated by minimally trained users in two low- and middle-income countries (Sierra Leone and Tanzania). Five labeled examples can help the model adapt to a new setting, and the identical training protocol transfers well to newer models such as SAM3, locating the advantage in ultrasound-specific pretraining rather than any single architecture. To ensure reproducibility and enable the community to build on SonoBase as a platform, we release all checkpoints, optimizer states, data-split indices, deduplication hashes, and starter code.
}

\keywords{Ultrasound, foundation model, segmentation, clinical measurement, domain shift, open science}

\maketitle


Ultrasound is the most widely used diagnostic imaging modality worldwide, spanning cardiology, obstetrics, emergency medicine, musculoskeletal assessment, and procedural guidance~\cite{Wang2020ultrasound}. Its advantages are well established: real-time acquisition, absence of ionizing radiation, portability, and low cost. These properties make ultrasound uniquely suited to point-of-care use and to settings where CT and MRI are not accessible. Segmentation of anatomical structures in ultrasound images is upstream of nearly all quantitative clinical measurements, including cardiac ejection fraction, fetal biometry, organ volumetry, and lesion characterization. The accuracy of these measurements determines whether patients receive appropriate treatment, whether pregnancies are correctly dated, and whether cancers are detected at actionable stages.

Current ultrasound AI is organized around narrow, single-task models trained on individual datasets for specific anatomies and clinical applications. These models achieve strong performance within their training distribution but degrade substantially when acquisition conditions change. A cardiac segmentation model trained on one scanner vendor may lose 25\% of its Dice score when evaluated on point-of-care devices from different manufacturers~\cite{cardiacFMvsDomainSpecific2025}. This fragility reflects shortcut learning from acquisition-specific artifacts rather than genuine anatomical understanding~\cite{shortcutLearning2024}. Public ultrasound datasets remain fragmented: individually small, inconsistently annotated, and rarely accompanied by the metadata (scanner type, site, patient demographics) needed for rigorous generalization evaluation. The result is an ecosystem of hundreds of narrow ultrasound AI models, none of which transfers reliably across the clinical heterogeneity that characterizes real-world ultrasound practice.

Foundation models offer an alternative: a single pretrained representation that generalizes across tasks, anatomies, and acquisition conditions, and that can be interactively prompted and corrected by clinicians. This approach has produced strong results in pathology~\cite{TITAN2024}, dermatology~\cite{PanDerm2025}, and ophthalmology~\cite{EyeFM2024}, where specialty-specific pretraining on large, diverse corpora yields representations that transfer broadly within the clinical domain. General-purpose promptable segmentation models (SAM~\cite{SAM2023}, SAM2~\cite{SAM2_2024}, SAM3~\cite{SAM3_2025}) and their medical adaptations (MedSAM~\cite{MedSAM2024}, MedSAM2~\cite{MedSAM2_2024}, MedSAM3~\cite{MedSAM3_2025}) provide interactive segmentation capability, but they are trained on natural images or broad multi-modal medical data, and their performance on ultrasound remains substantially below clinical utility. UltraSam~\cite{UltraSam2025}, a fine-tune of the first-generation SAM on 43 public datasets, showed that pooled public masks improve prompt-based segmentation of 2D ultrasound images, but it was evaluated on held-in test splits with overlap metrics only and, built on a base without the video memory of the SAM2 lineage, is not among the baselines compared here. Ultrasound presents distinctive challenges for foundation models: extreme variation in target scale (from millimeter nerve fascicles to centimeter-scale organs), high operator dependence, pervasive acoustic artifacts (shadowing, speckle, reverberation), and data spanning three distinct formats (2D images, video sequences, 3D volumes). Concurrent work has begun to address these challenges through generative modeling for breast ultrasound~\cite{BUSGen2026}, but these challenges demand an ultrasound-specific foundation model built on ultrasound-specific data at scale, spanning the full breadth of clinical applications.

Here we present SonoCorpus and SonoBase. SonoCorpus is the largest open-source ultrasound segmentation resource assembled to date, aggregating 456,963 images and frames with 1,626,085 expert segmentation masks from 53 public datasets spanning 24 clinical applications, three data modalities, and acquisition sites across 17 countries. For each dataset, we curated acquisition metadata including scanner vendor, acquisition site and country, operator context, and image quality, and used it to construct training and evaluation partitions in which distribution shifts in vendor, geography, operator expertise, image quality, and target organ can be studied deliberately rather than incidentally. SonoBase is an interactive segmentation foundation model that adapts SAM2 with an image-pyramid hybrid encoder: a Hiera transformer branch for global context at low resolution and two ConvNeXt branches for fine anatomical detail at higher resolutions, connected by cross-branch attention at each stage. We evaluate SonoBase across four axes designed to test distinct aspects of generalization. First, cross-modality generalization on eight held-out benchmark datasets spanning 2D, video, and 3D formats, including a head-to-head against task-specific specialist models trained on the same splits. Second, robustness to distribution shift on seven completely external datasets deliberately chosen to isolate specific shift types (geographic, device, operator, quality). Third, clinical measurement validity through derived ejection fraction, head circumference, gestational age, abdominal circumference, and prostate volume estimates compared against expert measurements and inter-observer variability benchmarks. We further evaluate catastrophic failure resolution, interactive workflow efficiency, downstream dense prediction transfer, cross-species generalization, subgroup fairness, and few-shot domain adaptation. Together, these constitute, to our knowledge, the broadest generalization evaluation reported for a medical imaging foundation model, spanning fifteen evaluation datasets, three data formats, two species, and two backbone generations. SonoBase is developed and reported in accordance with the CLAIM 2024 and REFINE consensus checklists~\cite{CLAIM2024,CLAIMupdate2024,REFINE2026}, and all artifacts needed to reproduce or extend it including model checkpoints, optimizer states, data-split indices, deduplication hashes, and fine-tuning starter code is publicly released, making SonoCorpus and SonoBase a platform on which ultrasound AI can be built rather than a single frozen model.


\subsection*{SonoCorpus and SonoBase establish an ultrasound foundation-model platform}

\begin{figure*}[p]
  \centering
  \includegraphics[width=\textwidth,height=0.86\textheight,keepaspectratio]{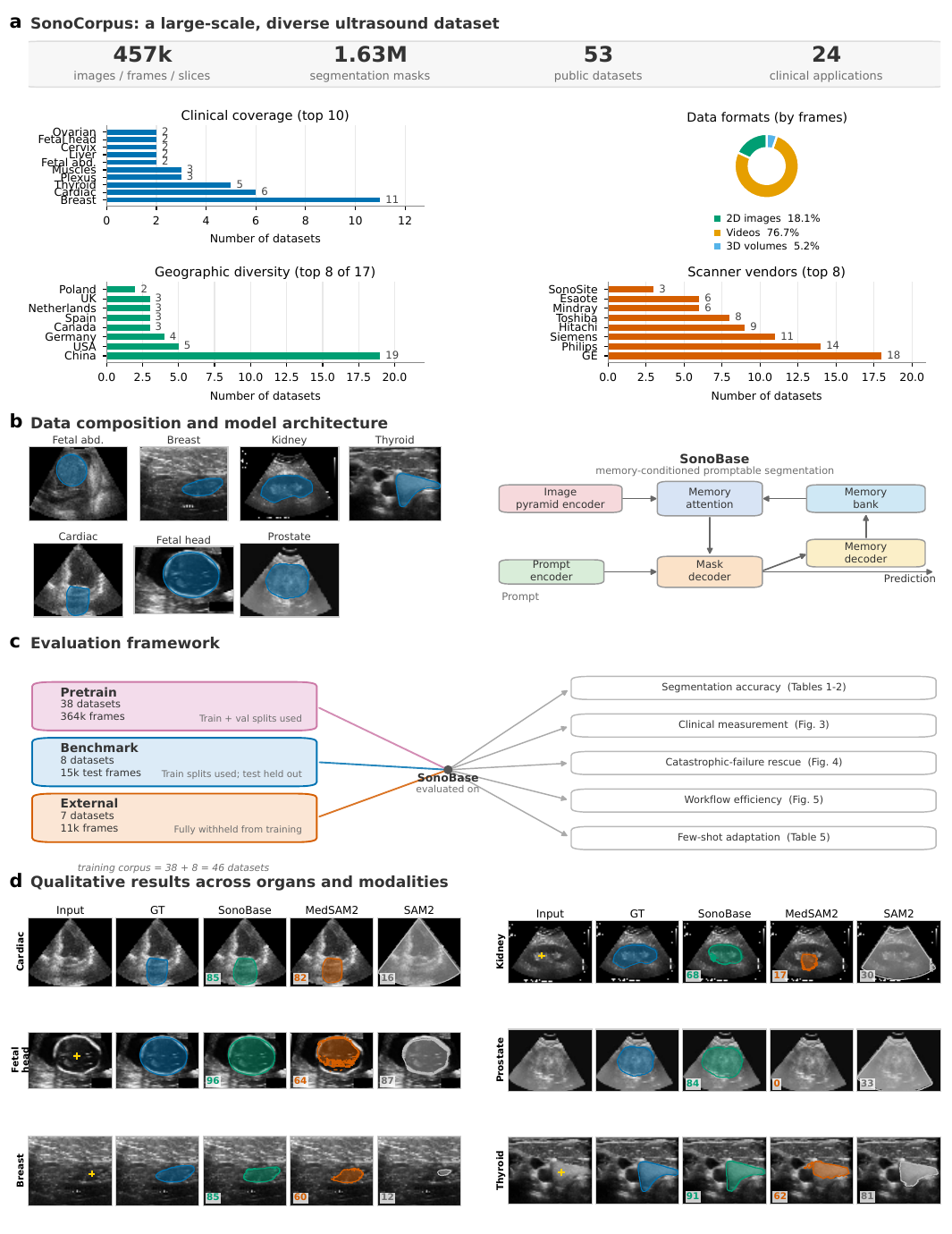}
  \vspace{-2pt}\caption{\textbf{SonoCorpus and SonoBase establish an ultrasound foundation-model platform.}
  \textbf{a,} SonoCorpus at a glance: 53 public datasets spanning 24 clinical applications and 17 countries, with 456,963 images/frames and 1,626,085 expert masks after deduplication. Charts summarize clinical coverage, data formats (by frame count), geographic diversity, and scanner vendors.
  \textbf{b,} Data composition and model architecture. Representative images across anatomies feed SonoBase, a memory-conditioned promptable segmentation model built on SAM2 that replaces its image encoder with an image-pyramid hybrid encoder (Methods) while retaining the prompt encoder, mask decoder, and memory mechanism for interactive segmentation and temporal propagation.
  \textbf{c,} The three-tier evaluation framework. Pretrain (38 datasets) contributes training and validation splits only; Benchmark (8 datasets) contributes training splits while its test partitions are held out for cross-modality evaluation; External (7 datasets) is fully withheld from training and tests domain shift. The training corpus is therefore 38 + 8 = 46 datasets, and the two evaluation tiers feed every downstream analysis reported in this paper.
  \textbf{d,} Representative segmentation results across six anatomies, with the point prompt marked in yellow where the displayed frame is the prompted one. For each dataset we show the case whose SonoBase IoU is the median over all scored images, so each row is a typical result rather than the most favorable one available.}
  \label{fig:overview}
\end{figure*}

SonoCorpus consolidates 53 public ultrasound datasets into a unified resource for pretraining and evaluation (Fig.~\ref{fig:overview}a). The collection spans 24 clinical applications across cardiac, fetal, breast, thyroid, kidney, prostate, nerve, muscle, and vascular imaging, with data from 17 countries acquired on scanners from all major manufacturers. Three data modalities are represented: 40 datasets of 2D images, 8 video datasets, and 5 volumetric 3D datasets. We curated metadata including acquisition site, scanner manufacturer and model, patient demographics (when available), and image quality annotations, enabling evaluation under controlled distribution shifts rather than ad hoc data splits. Deduplication verified zero overlap between training and test partitions across all 53 datasets, and splits were made at the patient level wherever patient identifiers are available (Methods).

We partitioned SonoCorpus into three evaluation tiers (Fig.~\ref{fig:overview}c). The Pretrain set provides the bulk of training data with 95\%/5\% train/validation splits. The Benchmark set consists of eight representative datasets (BUSI~\cite{BUSI}, Brachial-Plexus~\cite{Brachial-Plexus}, C-TRUS~\cite{C-TRUS}, CAMUS~\cite{CAMUS}, HC18~\cite{HC18}, PFUS~\cite{PFUS}, RegPro~\cite{RegPro}, TG3K~\cite{TG3K}) spanning all three modalities, with held-out test sets for evaluation. The External set comprises seven datasets (ACOUSLIC~\cite{ACOUSLIC}, BUS-BRA~\cite{BUS-BRA}, DDTI~\cite{DDTI}, FUGC~\cite{FUGC}, KidneyUS~\cite{KidneyUS}, LUMINOUS~\cite{LUMINOUS}, MMOTU-3d~\cite{MMOTU-2d}) completely withheld from training to assess generalization under natural domain shift. Each external dataset was chosen to isolate a specific shift type: ACOUSLIC represents combined geographic, device, and operator shift (smartphone-connected handheld probes, a blind-sweep protocol, and operators with one hour of training, at public health units in Sierra Leone and Tanzania); KidneyUS represents multi-vendor shift (nine scanner models from multiple manufacturers); DDTI and BUS-BRA represent geographic and scanner shift (Colombia and Brazil); FUGC represents device and acquisition shift (China, GE Voluson scanners). For each external dataset, 90\% of cases were reserved for testing and 10\% for few-shot adaptation experiments.

SonoBase adapts SAM2 by replacing its image encoder with an image-pyramid hybrid encoder while retaining the prompt encoder, mask decoder, and memory module that enable interactive segmentation and temporal propagation (Fig.~\ref{fig:overview}b). The hybrid encoder processes each input at three resolutions through complementary branches: a Hiera transformer backbone at low resolution for global semantic context, and two ConvNeXt convolutional backbones at medium and high resolution for local texture and boundary detail. Cross-branch attention units exchange information between branches at each of four hierarchical stages, and stage-wise fusion produces a unified multi-scale feature pyramid compatible with the rest of SAM2 (Methods). For single images, the memory pathway is disabled; for videos and 3D volumes, the memory mechanism enables prompt-once segmentation where a single user interaction on one frame propagates to the entire sequence.

\subsection*{SonoBase generalizes across held-out benchmarks and completely external datasets}

\begin{figure*}[!t]
  \centering
  \includegraphics[width=\textwidth]{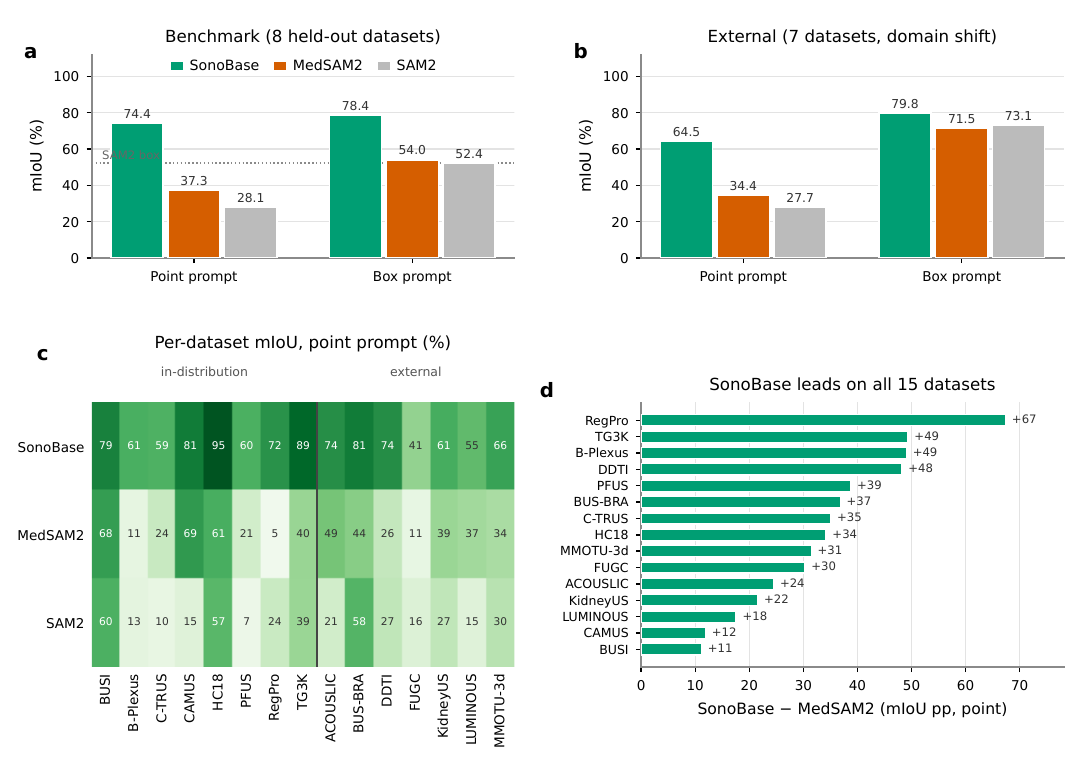}
  \caption{\textbf{SonoBase generalizes across held-out benchmarks and completely external datasets.}
  \textbf{a,} Average mIoU on eight benchmark datasets under point and box prompts. SonoBase with a single point prompt (74.4) outperforms SAM2 with a bounding box (52.4), demonstrating that ultrasound-specific pretraining provides more spatial information than a geometric prior in a generic model.
  \textbf{b,} Average mIoU on seven external datasets under natural domain shift. SonoBase's external box performance (79.8) exceeds what baselines achieve on in-distribution benchmark data.
  \textbf{c,} Per-dataset performance heatmap showing consistent SonoBase advantage across all 15 evaluation datasets, with external datasets grouped at right.
  \textbf{d,} Pairwise comparison of SonoBase versus MedSAM2 shows SonoBase leads on all 15 datasets, with the largest point-prompt advantages on RegPro (+67.4 pp), TG3K (+49.3 pp), Brachial-Plexus (+49.0 pp), and DDTI (+48.2 pp).}
  \label{fig:generalization}
\end{figure*}

SonoBase substantially outperforms both SAM2 (pretrained on natural images, no ultrasound fine-tuning) and MedSAM2 (fine-tuned on multi-modal medical data including some ultrasound) on all benchmark and external datasets under both point and box prompt settings (Fig.~\ref{fig:generalization}). Throughout, ``baselines'' denotes these promptable models (and MedSAM3 where evaluated); ``specialists'' denotes task-specific models trained on a single dataset, against which SonoBase is compared at the end of this section.

On eight held-out benchmark datasets, SonoBase achieves 74.4 mIoU with a single point prompt and 78.4 mIoU with a bounding-box prompt, compared to 28.1/52.4 for SAM2 and 37.3/54.0 for MedSAM2 (point/box mIoU; paired values follow this order throughout; Table~\ref{tab:benchmark}). The improvement over SAM2 (+46.3 percentage points on point prompts) represents the largest reported gain of any SAM adaptation for ultrasound. SonoBase with a single point click (74.4 mIoU) substantially surpasses SAM2 with a full bounding box (52.4 mIoU), demonstrating that ultrasound-specific pretraining provides more spatial information than the strong geometric prior of a bounding box in a generic model. Per-dataset results confirm consistent improvement: SonoBase leads on every benchmark dataset (Supplementary Table~S2), with gains most pronounced on challenging datasets such as Brachial-Plexus (60.5 versus 12.6 point mIoU for SAM2), C-TRUS (59.3 versus 10.3), and RegPro (72.2 versus 4.8 for MedSAM2, which fails almost completely on 3D transrectal volumes).

On seven completely external datasets, SonoBase maintains strong performance: 64.5 mIoU (point) and 79.8 mIoU (box), compared to 34.4/71.5 for MedSAM2 and 27.7/73.1 for SAM2 (Table~\ref{tab:external}). External performance exceeds what baselines achieve even on in-distribution benchmark data, indicating that SonoBase generalizes to unseen acquisition conditions rather than memorizing training-distribution shortcuts. The external datasets represent named, interpretable distribution shifts. On ACOUSLIC (smartphone-connected handheld probes, blind sweeps by minimally trained operators, Sierra Leone and Tanzania), SonoBase achieves 73.7/77.2 point/box mIoU, the most extreme combined shift in our evaluation. On KidneyUS (nine scanner models from multiple manufacturers at a single Canadian center), SonoBase reaches 61.0/87.9, demonstrating vendor-agnostic generalization. On DDTI (thyroid, Colombia) and BUS-BRA (breast, Brazil), the model handles geographic and scanner shift with 74.0/82.6 and 81.1/84.4 mIoU respectively. FUGC (cervix, China) remains the most challenging external dataset (41.1/63.0), likely due to the combination of a difficult anatomy with limited training representation; it is also where few-shot adaptation yields the largest relative gains among our adaptation experiments (see ``Few-shot adaptation demonstrates superior sample efficiency''). Full per-dataset breakdowns are provided in Supplementary Tables~S2--S3; the MedSAM3 and SB-SAM3 per-dataset tables are S33 and S34.

Prompted SonoBase gives up no accuracy to per-dataset specialists. We trained one nnU-Net ResEnc-M~\cite{nnUNet2021,nnUNetRevisited2024} per Benchmark dataset on the same training split SonoBase saw and scored it, unprompted, on the identical test rows with the same code (Methods). The specialist reaches 76.3 macro mIoU against 78.4 for box-prompted SonoBase and 74.4 with a single point (Table~\ref{tab:benchmark}). Paired on identical rows, box-prompted SonoBase leads on five of the eight datasets (BUSI, C-TRUS, CAMUS, PFUS, RegPro; 2--12 IoU points, all BH-FDR-adjusted $q < 0.001$), is within a point on HC18 (paired difference $0.0$ [$-0.8$, $0.5$] under the deterministic prompt) and, at the video level, on Brachial-Plexus ($+1.1$ [$-3.7$, $5.6$] for the specialist, driven by the needle, which first-frame box propagation loses), and trails only on TG3K, whose public split is frame-level within 16 videos, so a single-dataset model memorizes near-duplicate frames (Supplementary Table~S3b). The specialists' worst cases are localization failures, extra or wrong regions, which is what a prompt helps resolve. A single point, far less spatial information than a box, can approximate a dedicated model.

The hybrid encoder design contributes beyond fine-tuning alone. The full three-branch hybrid architecture (84.0 mIoU) outperforms the best single-branch variant (Hiera-B+, 81.5 mIoU) by 2.5 percentage points, and outperforms homogeneous three-branch designs (82.0--82.5 on an ablation subset) by 1.5--2.0 points (Supplementary Table~S24). Ablation of cross-branch attention blocks confirms monotonic improvement with block count, plateauing at 9--12 blocks (Supplementary Table~S25). These results indicate that the gain comes from complementary feature extraction across transformer and convolutional branches, consistent with recent findings that heterogeneous architectures outperform homogeneous designs in medical imaging~\cite{GenSeg2025, MINIM2025}.

\begin{table*}[!t]
  \centering
  \scriptsize
  \setlength{\tabcolsep}{4pt}
  \caption{\textbf{Segmentation performance on the held-out test partitions of the eight Benchmark datasets (mIoU, \%).} Point: a point drawn uniformly from the ground-truth foreground; values are the mean of three prompt seeds. Box: single bounding-box prompt with standard jitter. SonoBase is fine-tuned on SonoCorpus; SAM2 uses original pretrained weights without ultrasound fine-tuning; MedSAM2 uses the publicly released Hiera-Tiny checkpoint. SonoBase with a single point click outperforms SAM2 with a full bounding box (SonoBase rows in bold). The specialist block reports an unprompted nnU-Net ResEnc-M~\cite{nnUNet2021,nnUNetRevisited2024} trained separately on each dataset's training split and scored on the identical test rows (single run; paired differences in Supplementary Table~S3b). Per-dataset seed spread and Dice values are provided in Supplementary Table~S2.}
  \label{tab:benchmark}
  \resizebox{\linewidth}{!}{
  \begin{tabular}{lcccccccc|c}
    \toprule
    \textbf{Model (Prompt)} & \textbf{BUSI} & \textbf{B-Plexus} & \textbf{C-TRUS} & \textbf{CAMUS} & \textbf{HC18} & \textbf{PFUS} & \textbf{RegPro} & \textbf{TG3K} & \textbf{Avg} \\
    \midrule
    \multicolumn{10}{l}{\textit{Point prompt}} \\
    SAM2 (Hiera-B+, no ft) & 59.6 & 12.6 & 10.3 & 15.1 & 56.6 & 7.5 & 23.9 & 39.1 & 28.1 \\
    MedSAM2 (Hiera-T)      & 67.6 & 11.5 & 24.2 & 69.2 & 60.6 & 21.1 & 4.8 & 39.7 & 37.3 \\
    \textbf{SonoBase (Hybrid B+/S/T)} & \textbf{78.7} & \textbf{60.5} & \textbf{59.3} & \textbf{81.1} & \textbf{94.8} & \textbf{59.8} & \textbf{72.2} & \textbf{89.0} & \textbf{74.4} \\
    \midrule
    \multicolumn{10}{l}{\textit{Box prompt}} \\
    SAM2 (Hiera-B+, no ft) & 80.3 & 41.5 & 49.8 & 56.2 & 81.3 & 37.9 & 2.8 & 69.0 & 52.4 \\
    MedSAM2 (Hiera-T)      & 81.2 & 26.2 & 46.4 & 77.9 & 90.8 & 48.0 & 1.7 & 60.1 & 54.0 \\
    \textbf{SonoBase (Hybrid B+/S/T)} & \textbf{84.0} & \textbf{62.1} & \textbf{67.1} & \textbf{84.3} & \textbf{94.9} & \textbf{70.9} & \textbf{75.4} & \textbf{88.5} & \textbf{78.4} \\
    \midrule
    \multicolumn{10}{l}{\textit{Specialist, unprompted (one model per dataset, trained on the same training split)}} \\
    nnU-Net ResEnc-M (2D) & 76.8 & 66.7 & 59.5 & 83.7 & 95.7 & 66.1 & 63.5 & 98.5 & 76.3 \\
    \bottomrule
  \end{tabular}
  }
\end{table*}

\begin{table*}[!t]
  \centering
  \scriptsize
  \setlength{\tabcolsep}{4pt}
  \caption{\textbf{Segmentation performance on completely external datasets under natural domain shift (mIoU, \%).} All external datasets were withheld from training. Each dataset isolates a named distribution shift: ACOUSLIC (handheld device, blind sweeps by minimally trained operators, Sierra Leone and Tanzania), KidneyUS (multi-vendor, nine scanner models), BUS-BRA (Brazil, scanner shift), DDTI (Colombia, scanner and geographic shift), FUGC (China, device and acquisition shift), LUMINOUS (Canada, anatomy and geographic shift), MMOTU-3d (China, ovarian). SonoBase's external performance exceeds what baselines achieve on in-distribution benchmark data. Per-dataset seed spread and Dice values are provided in Supplementary Table~S3.}
  \label{tab:external}
  \resizebox{\linewidth}{!}{
  \begin{tabular}{lccccccc|c}
    \toprule
    \textbf{Model (Prompt)} & \textbf{ACOUSLIC} & \textbf{BUS-BRA} & \textbf{DDTI} & \textbf{FUGC} & \textbf{KidneyUS} & \textbf{LUMINOUS} & \textbf{MMOTU-3d} & \textbf{Avg} \\
    \midrule
    \multicolumn{9}{l}{\textit{Point prompt}} \\
    SAM2 (Hiera-B+, no ft) & 21.0 & 57.6 & 27.2 & 16.1 & 27.2 & 15.0 & 29.7 & 27.7 \\
    MedSAM2 (Hiera-T)      & 49.3 & 44.3 & 25.8 & 10.8 & 39.4 & 37.0 & 34.5 & 34.4 \\
    \textbf{SonoBase (best hybrid)} & \textbf{73.7} & \textbf{81.1} & \textbf{74.0} & \textbf{41.1} & \textbf{61.0} & \textbf{54.6} & \textbf{65.9} & \textbf{64.5} \\
    \midrule
    \multicolumn{9}{l}{\textit{Box prompt}} \\
    SAM2 (Hiera-B+, no ft) & 70.3 & 83.9 & 78.9 & 56.1 & 81.6 & 67.3 & 73.9 & 73.1 \\
    MedSAM2 (Hiera-T)      & 61.8 & 76.1 & 76.2 & 54.8 & 79.7 & 76.4 & 75.7 & 71.5 \\
    \textbf{SonoBase (best hybrid)} & \textbf{77.2} & \textbf{84.4} & \textbf{82.6} & \textbf{63.0} & \textbf{87.9} & \textbf{80.4} & \textbf{82.9} & \textbf{79.8} \\
    \bottomrule
  \end{tabular}
  }
\end{table*}

\subsection*{The SonoCorpus platform extends to new base models}

Promptable segmentation models now appear at the pace of the general vision field: SAM3~\cite{SAM3_2025} introduced concept prompting within a year of SAM2, and its medical adaptation MedSAM3~\cite{MedSAM3_2025} followed within weeks. Each release poses the same question for a clinical team: not which architecture is newest, but whether an ultrasound model built on it would be better, which can only be answered with curated data, controlled splits, and a training recipe held fixed across bases. SonoCorpus provides this scaffolding. To demonstrate it, we retrained the complete SonoBase recipe, the same 46 training datasets, the same splits, the same schedule and loss, on a SAM3.1 base, producing \textbf{SB-SAM3}, and evaluated all models under a matched box-prompt protocol scored by a single Dice implementation (Methods). Throughout, ``SonoBase'' refers to the released SAM2-based model; SB-SAM3 is a controlled comparison, not a second released model.

\paragraph{Ultrasound pretraining, not backbone generation, drives the gain.} Zero-shot, the two bases are indistinguishable on ultrasound: 0.823 macro Dice for both SAM2 and SAM3.1 (Table~\ref{tab:headtohead}), so SAM3's architectural advances do not, by themselves, transfer to this modality. After identical fine-tuning on SonoCorpus both improve dramatically, and the SAM2-based SonoBase remains ahead: 0.897 versus 0.869, leading on 13 of 15 datasets. The recipe, corpus, and splits therefore transfer to a second backbone and reproduce most of the gain; the advantage is a property of the ultrasound-specific training pipeline rather than of any backbone generation. Freezing the SB-SAM3 encoder costs a further 0.008 Dice, confirming that full-parameter adaptation of the ultrasound encoder is where the gain originates (Supplementary Section~S15).

\paragraph{The platform yields a model that surpasses the current state of the art.} The same scaffolding supports a fair comparison against MedSAM3, the strongest available medical adaptation of SAM3, evaluated on the fourteen datasets free of training-data overlap (BUSI appears in MedSAM3's training corpus and is excluded; Methods). Because MedSAM3 cannot be driven by a box alone, it received its native text-plus-box prompt --- strictly more information than the box-only SonoBase (Methods). SonoBase nevertheless leads on all fourteen datasets, with mean Dice 0.896 versus 0.776 (Table~\ref{tab:medsam3}). The margin is largest on video datasets, where per-frame concept prompting has no temporal memory to exploit, CAMUS ($+0.279$), PFUS ($+0.185$), and on the hardest 2D targets, C-TRUS ($+0.203$) and FUGC ($+0.187$); it is smallest where both models approach the annotation ceiling (HC18, $+0.006$). MedSAM3 is a substantially stronger baseline than MedSAM2, it scores 0.831 Dice on RegPro, where MedSAM2 collapses entirely, and SonoBase's lead on every dataset indicates the advantage tracks ultrasound-specific pretraining rather than baseline vintage.

\paragraph{Ultrasound fine-tuning unlocks text prompting.} The SAM3 lineage adds an interaction mode SAM2 lacks: open-vocabulary text prompts. Zero-shot, this capability is close to unusable on ultrasound (0.138 macro Dice), largely because the target is never detected at all. Fine-tuning on SonoCorpus raises text-only performance five-fold to 0.681, with detection recall reaching 0.811, although text prompting does not yet match box-prompted accuracy, since a box still resolves which object is meant (Supplementary Section~S16, Table~S35). Capabilities introduced by each new base model thus become clinically viable through ultrasound-specific adaptation on the platform. Because the corpus, splits, recipe, and evaluation harness are all released, this loop of taking a new base, adapting it, and comparing under one protocol can be repeated as future backbones appear, enabling informed selection of a base model for a target clinical application rather than adoption by novelty.

\begin{table*}[!t]
  \centering
  \small
  \caption{\textbf{Matched box-prompt head-to-head across five models (macro Dice, 15 datasets).} All fine-tuned models use the identical recipe, corpus, and splits; all values are computed per frame from archived per-dataset outputs with a single Dice implementation under a matched per-frame protocol (Methods). The two zero-shot bases tie on ultrasound; after identical fine-tuning the SAM2-based SonoBase leads SB-SAM3 on 13 of 15 datasets, losing BUS-BRA and tying KidneyUS. Per-dataset values in Supplementary Table~S34.}
  \label{tab:headtohead}
  \begin{tabular}{llccc}
    \toprule
    \textbf{Model} & \textbf{Base} & \textbf{All 15} & \textbf{Benchmark (8)} & \textbf{External (7)} \\
    \midrule
    \multicolumn{5}{l}{\textit{Zero-shot (no ultrasound fine-tuning)}} \\
    SAM2 (no ft)            & SAM2   & 0.823 & 0.791 & 0.861 \\
    SAM3.1 (no ft)          & SAM3.1 & 0.823 & 0.802 & 0.847 \\
    \midrule
    \multicolumn{5}{l}{\textit{After identical ultrasound fine-tuning on SonoCorpus}} \\
    SB-SAM3 (frozen encoder) & SAM3.1 & 0.861 & 0.843 & 0.881 \\
    SB-SAM3 (full)           & SAM3.1 & 0.869 & 0.862 & 0.878 \\
    \textbf{SonoBase}        & SAM2   & \textbf{0.897} & \textbf{0.892} & \textbf{0.904} \\
    \bottomrule
  \end{tabular}
\end{table*}

\begin{table*}[!t]
  \centering
  \scriptsize
  \setlength{\tabcolsep}{3pt}
  \caption{\textbf{SonoBase versus MedSAM3 on fourteen leakage-clean datasets (Dice).} BUSI is excluded because it appears in MedSAM3's training data. MedSAM3 receives its native text-plus-box prompt, strictly more prompt information than SonoBase, which is given the box only, and all models are prompted per frame under a matched protocol (Methods). SonoBase leads on all fourteen datasets.}
  \label{tab:medsam3}
  \resizebox{\linewidth}{!}{
  \begin{tabular}{lccccccc|c}
    \toprule
    \textbf{Dataset} & \textbf{C-TRUS} & \textbf{HC18} & \textbf{TG3K} & \textbf{B-Plexus} & \textbf{CAMUS} & \textbf{PFUS} & \textbf{RegPro} & \multicolumn{1}{c}{} \\
    \midrule
    MedSAM3            & 0.599 & 0.972 & 0.811 & 0.652 & 0.641 & 0.646 & 0.831 & \multicolumn{1}{c}{} \\
    \textbf{SonoBase}  & \textbf{0.803} & \textbf{0.978} & \textbf{0.947} & \textbf{0.829} & \textbf{0.920} & \textbf{0.831} & \textbf{0.910} & \multicolumn{1}{c}{} \\
    $\Delta$           & +0.203 & +0.006 & +0.136 & +0.177 & +0.279 & +0.185 & +0.079 & \multicolumn{1}{c}{} \\
    \midrule
    \textbf{Dataset} & \textbf{ACOUSLIC} & \textbf{BUS-BRA} & \textbf{DDTI} & \textbf{FUGC} & \textbf{KidneyUS} & \textbf{LUMINOUS} & \textbf{MMOTU-3d} & \textbf{Mean} \\
    \midrule
    MedSAM3            & 0.896 & 0.897 & 0.865 & 0.595 & 0.875 & 0.797 & 0.784 & 0.776 \\
    \textbf{SonoBase}  & \textbf{0.948} & \textbf{0.920} & \textbf{0.921} & \textbf{0.782} & \textbf{0.938} & \textbf{0.904} & \textbf{0.912} & \textbf{0.896} \\
    $\Delta$           & +0.052 & +0.023 & +0.055 & +0.187 & +0.063 & +0.107 & +0.128 & \textbf{+0.120} \\
    \bottomrule
  \end{tabular}
  }
\end{table*}

\subsection*{Segmentation quality enables clinically meaningful measurements with expert-level agreement}

\begin{figure*}[!t]
  \centering
  \includegraphics[width=\textwidth]{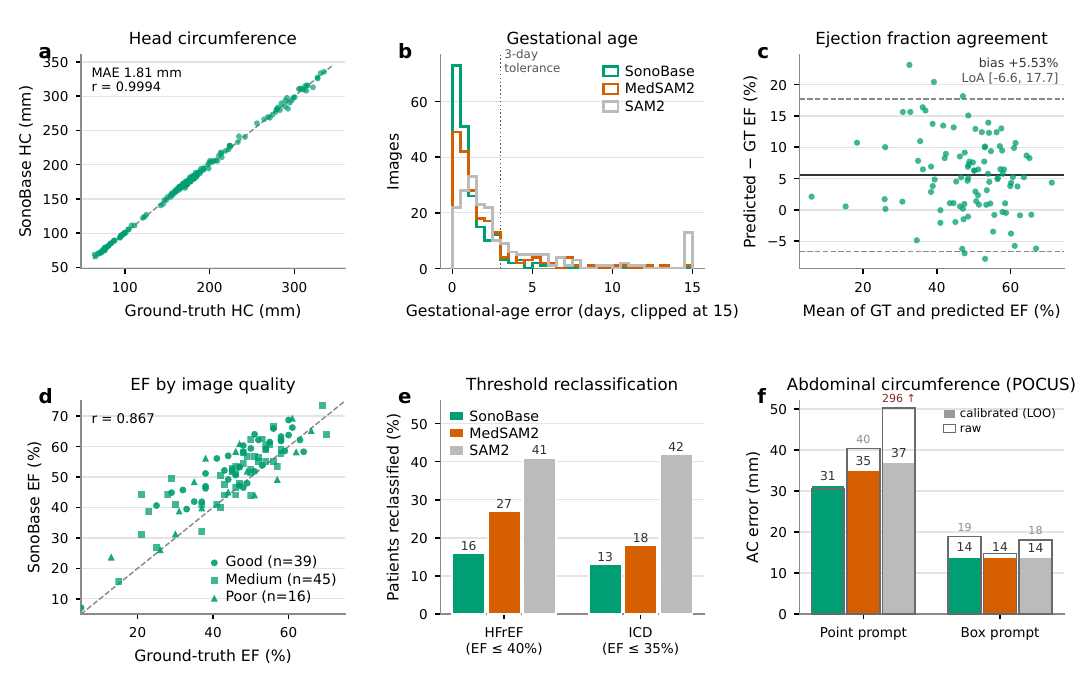}
  \caption{\textbf{SonoBase-derived clinical measurements achieve near-expert agreement.}
  \textbf{a,} Head circumference (HC) scatter on HC18 ($n=201$). SonoBase box-prompt error of 1.81~mm falls below inter-observer variability (3--5~mm) and approaches the 1.37~mm ellipse-fitting floor, with $r = 0.9994$.
  \textbf{b,} Gestational age (GA) error distribution. SonoBase: 93.0\% within 3 days (box), 0\% false workup; point-prompt baselines exceed 52\% false workup.
  \textbf{c,} Bland-Altman analysis of ejection fraction (EF) derived from SonoBase segmentations on CAMUS ($n=100$), ASE disc-pairing primary. Positive bias ($+5.53\%$ box) with 95\% limits of agreement $[-6.65, +17.71]\%$.
  \textbf{d,} SonoBase-derived versus ground-truth EF by image quality grade; the dashed line marks identity. Correlation ($r = 0.867$ box) is maintained across all quality grades including ``poor'' ($r = 0.849$), against a ground-truth-mask extraction ceiling of $r = 0.943$.
  \textbf{e,} Clinical reclassification rates at HFrEF (40\%) and ICD candidacy (35\%) thresholds. SonoBase box prompt reclassifies 13.0\% of patients at the 35\% threshold, compared to 18.0\% (MedSAM2) and 42.0\% (SAM2).
  \textbf{f,} Abdominal circumference (AC) under extreme distribution shift (ACOUSLIC, handheld point-of-care probe, Sierra Leone and Tanzania). Solid bars: MAE after held-out linear recalibration of all three models (leave-one-out; Supplementary Table~S9d); outlines: raw MAE. Under point prompts SonoBase is best before and after calibration; under box prompts the raw MedSAM2 advantage ($p = 7.8 \times 10^{-6}$) is a scale offset that calibration removes (13.9, 13.9, and 13.8~mm; $p = 0.998$).}
  \label{fig:clinical}
\end{figure*}

The purpose of ultrasound segmentation is to derive the quantitative measurements that inform clinical decisions: ejection fraction for heart failure management, head circumference for gestational dating, abdominal circumference for fetal growth assessment, and organ volume for disease staging, among others. We evaluated SonoBase-derived measurements against expert ground truth on four clinical endpoints, comparing measurement error to established inter-observer variability benchmarks (Fig.~\ref{fig:clinical}).

\paragraph{Fetal head circumference and gestational age.} On the HC18 dataset ($n=201$ images), SonoBase head circumference (HC) error measured 2.49~mm (point prompt) and 1.81~mm (box prompt), compared to a ground-truth ellipse-fitting floor of 1.37~mm (Supplementary Tables~S10--S11). These values fall below the 3--5~mm inter-observer variability for HC measurement at 20 weeks~\cite{Sarris2012}. With box prompts, 96.0\% of SonoBase predictions fell within 5~mm clinical tolerance and 80.6\% within 3~mm. Pearson correlation reached $r = 0.999$, indicating near-perfect linear agreement with ground truth. Two specialists trained on the HC18 training split and scored unprompted on the same 201 images were less accurate, a DeepLabV3+~\cite{DeepLabV3plus2018} at 2.30~mm and nnU-Net at 2.83~mm ($q < 10^{-3}$ for both against box-prompted SonoBase; 72.6\% and 71.6\% within 3~mm), despite comparable overlap: their worst cases are localization failures that a prompt removes (Supplementary Table~S11).

HC determines gestational age (GA) through established formulae~\cite{Hadlock1984}. SonoBase-derived GA estimates showed 1.20-day mean absolute error (box prompt) and 1.65 days (point), with 93.0\% and 85.1\% of cases within 3 days of true GA (Supplementary Table~S12). False workup rates for clinically significant discrepancies ($>14$ days) were 0.0\% for both prompts. Baseline models produced GA errors of 33--63 days (MedSAM2 and SAM2 point prompts), triggering unnecessary workup referrals in over 52\% of cases.

\paragraph{Cardiac ejection fraction.} On the CAMUS dataset ($n=100$ patients), SonoBase-derived ejection fraction (EF) estimates via Simpson's biplane method showed mean absolute error of 8.86\% (point prompt) and 6.63\% (box prompt), with Pearson correlation $r=0.861$ and $r=0.867$ respectively --- 92\% of the $r = 0.943$ ceiling attainable from the ground-truth masks themselves (Supplementary Tables~S4--S5). The box-prompt error falls within the reported inter-observer variability of 5.9--6.9\% for echocardiographic EF measurement~\cite{Hoffmann2005}; the point-prompt error exceeds it. A CAMUS-trained nnU-Net, unprompted, reaches 6.28\% on the same patients, statistically indistinguishable from box-prompted SonoBase ($q = 0.39$), and an EchoNet-Dynamic segmentation network fine-tuned on the CAMUS training split reaches 7.43\% (Supplementary Table~S5). Bland-Altman analysis showed a systematic positive bias ($+8.37\%$ point, $+5.53\%$ box) with 95\% limits of agreement of $[-4.10, 20.84]$ and $[-6.65, 17.71]$ respectively; the bias is directional rather than random (Supplementary Tables~S4--S5).

The clinical consequence of EF accuracy is measured at treatment-relevant thresholds. At the EF $\leq 40\%$ threshold for heart failure with reduced ejection fraction (HFrEF)~\cite{McDonagh2021}, SonoBase misclassified 16.0\% of patients (box prompt), compared to 27.0\% for MedSAM2 and 41.0\% for SAM2. At the more consequential EF $\leq 35\%$ threshold for implantable cardioverter-defibrillator (ICD) candidacy~\cite{Heidenreich2022}, SonoBase box-prompt misclassification dropped to 13.0\% (13 of 100 patients), compared to 18.0\% for MedSAM2, 42.0\% for SAM2, and 11.0\% for the CAMUS-trained nnU-Net. In the clinically ambiguous gray zone (EF 30--50\%, $n=52$), SonoBase box prompt achieved 6.74\% MAE with 13.5\% misclassification at the 35\% threshold (Supplementary Table~S8). SonoBase maintained robust EF accuracy across image quality grades ($r > 0.84$ even on images rated ``poor'': $r = 0.849$ box, $r = 0.853$ point), while MedSAM2 fell to $r = 0.732$ on poor-quality images under point prompts and degraded further under box prompts, and SAM2's point-prompt correlation was near zero at every grade (Supplementary Tables~S6--S7).

Uncorrected Cohen's kappa at the 35\% threshold was 0.457 for SonoBase versus 0.558 for MedSAM2, a difference that traces to SonoBase's directional bias rather than to closer baseline agreement. Because the bias is directional, a simple linear recalibration recovers threshold agreement: fitting the correction on one random half of the cohort and evaluating on the held-out half (2,000 splits), SonoBase's kappa rises to 0.546 at the 35\% threshold and to 0.639 at 40\%, while mean absolute error falls from 6.63\% to 4.95\%, well inside the inter-observer band. The same procedure lowers MedSAM2's kappa at 35\% from 0.558 to 0.237, indicating that its nominally higher uncorrected agreement reflected cancellation between a large negative bias and the class balance rather than closer agreement (interval estimates in Supplementary Table~S9b).

SAM2 without ultrasound fine-tuning produced catastrophic EF estimates: 45.01\% mean absolute error on point prompts ($r = 0.094$), with 80.0\% of patients misclassified at the 35\% threshold. Box prompts partially rescued SAM2 (17.41\% MAE, $r = 0.368$), but the error remained clinically dangerous. MedSAM2 achieved 12.80\% MAE (point) and 10.35\% (box), functional but outside inter-observer variability and with substantially higher misclassification rates. SonoBase with box prompts, by contrast, delivers EF estimates within established inter-observer variability, supporting prospective evaluation as an assistive measurement tool at treatment-relevant thresholds.

\paragraph{Fetal abdominal circumference under extreme distribution shift.} On the ACOUSLIC dataset ($n=270$ sweeps from the ACOUSLIC-AI challenge, designed for operator-agnostic measurement in low-income countries~\cite{ACOUSLIC}: handheld probes, blind sweeps by minimally trained operators, Sierra Leone and Tanzania), SonoBase abdominal circumference (AC) error measured 30.50~mm (point) and 18.92~mm (box), compared to 40.34/14.77~mm for MedSAM2 and 296.34/18.11~mm for SAM2 (Supplementary Tables~S13--S14); the ground-truth ellipse-fitting pipeline itself has a 7.39~mm floor on this dataset when all annotated frames are averaged. With point prompts, SonoBase significantly outperformed both baselines (Wilcoxon $p = 2.6 \times 10^{-6}$ versus MedSAM2). With box prompts, MedSAM2 records the lowest raw error and the difference is significant ($p = 7.8 \times 10^{-6}$), but this is a scale offset rather than lower contour accuracy: SonoBase's box-prompted contours have the highest per-frame overlap and the lowest per-frame circumference error, yet are about 3.5\% inside the expert boundary, and the ellipse perimeter inherits that offset while an overlap score does not. After identical held-out linear recalibration of all three models, the same procedure applied to ejection fraction above, the models are indistinguishable (13.9, 13.9, and 13.8~mm; $p = 0.998$; Supplementary Table~S9d), and under the challenge organizers' optimal-plane protocol they tie (13.1, 12.8, and 12.9~mm)~\cite{ACOUSLIC}. The challenge's three winning solutions, retrained under subject-level five-fold cross-validation on the 300 public sweeps and scored on the identical frames and videos, mark what in-distribution training enables on this task. The third-placed nnU-Net hybrid (A3) reaches 89.4 mIoU and 10.8~mm AC (zero-shot SonoBase under the same deterministic prompt: 77.6 and 18.9~mm; $q < 10^{-3}$), whereas the other two winners tie or trail zero-shot SonoBase on this frame-averaged protocol (20.0 and 18.2~mm) and show the same inward offset ($-19$ and $-17$~mm; Supplementary Table~S14b). Five labeled examples from the target domain reverse the raw deficit against MedSAM2: after decoder-only few-shot adaptation, SonoBase reaches 17.17~mm AC error versus 29.80~mm for identically adapted MedSAM2 ($p = 2.6 \times 10^{-20}$), and thirty examples bring it to 13.7~mm (Supplementary Table~S32; see ``Few-shot adaptation demonstrates superior sample efficiency''). Throughout this most extreme distribution shift in our evaluation which combines a handheld probe, blind sweeps by minimally trained operators, and acquisition sites absent from training, SonoBase remains functional where unadapted SAM2 collapses entirely (296~mm MAE with point prompts).

\paragraph{Prostate volume.} On the RegPro 3D volume dataset ($n=8$ patients), SonoBase prostate volume error measured 6.28~mL (point) and 3.62~mL (box), with relative error of 15.95\% and 9.27\%; the box-prompt error falls within the inter-observer variability of 11.4\%~\cite{Tong1998} while the point-prompt error exceeds it. Pearson correlation reached $r = 0.960$ (point) and $r = 0.962$ (box), and bootstrap 95\% confidence intervals separate SonoBase from MedSAM2 by an order of magnitude (box: [2.28, 5.00]~mL versus [35.35, 47.77]~mL; Supplementary Table~S18). These results extend SonoBase's measurement utility to 3D volumetric endpoints.

All pairwise comparisons (Wilcoxon signed-rank tests) were corrected for multiple testing using the Benjamini-Hochberg procedure controlling the false discovery rate at $q = 0.05$ (Supplementary Section~S6). After correction, all comparisons remain significant except ACOUSLIC box-prompt SonoBase vs. SAM2 (corrected $p = 0.977$); the ACOUSLIC box-prompt SonoBase vs. MedSAM2 comparison is significant in MedSAM2's favor (corrected $p = 1.6 \times 10^{-5}$), the scale offset described above.

\subsection*{SonoBase resolves catastrophic baseline failures and reduces interaction burden}

\begin{figure*}[!t]
  \centering
  \includegraphics[width=\textwidth]{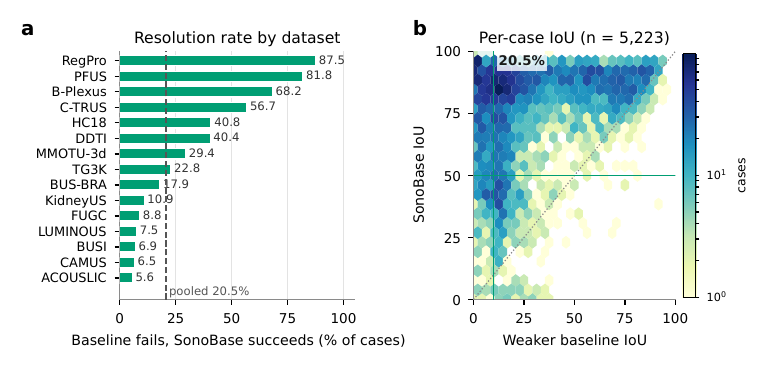}\\[2pt]
  \includegraphics[width=\textwidth]{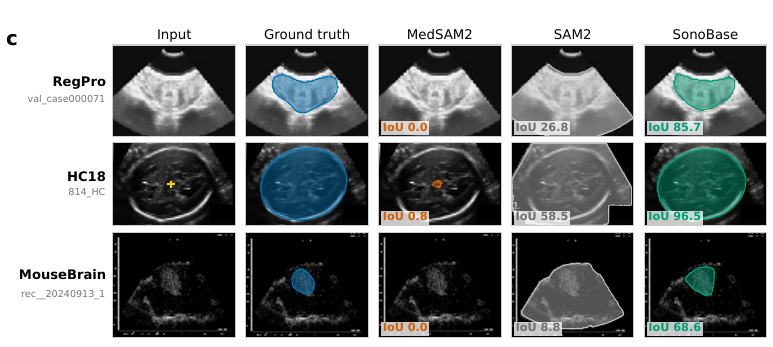}
  \caption{\textbf{SonoBase resolves catastrophic baseline failures across diverse anatomies and acquisition conditions.}
  A catastrophic baseline failure is a test case on which \emph{at least one} baseline scores IoU $< 10$ (no usable segmentation) under point prompting with no corrections; a failure is \emph{resolved} when SonoBase scores IoU $> 50$ (clinically useful) on the same case. Panels \textbf{a} and \textbf{b} are computed per case over all fifteen evaluation datasets (5,223 cases); for video and 3D datasets a case's IoU is the mean over its frames or slices, so the panels match Supplementary Table~S22 exactly.
  \textbf{a,} Per-dataset resolution rate. Dashed line marks the pooled rate of 20.5\%.
  \textbf{b,} Joint distribution of per-case IoU. The horizontal axis is the weaker of the two baselines, so the shaded upper-left quadrant contains exactly the cases counted in \textbf{a}. Color is case count on a log scale; the dotted line is parity. The mass above the diagonal at low baseline IoU is the asymmetry the rate summarizes.
  \textbf{c,} Representative resolved cases, rendered from the same predictions: two from the fifteen-dataset evaluation (RegPro, HC18) and one from the cross-species mouse brain dataset, illustrating that failure rescue extends beyond the core evaluation tiers. Among qualifying cases we show the one whose SonoBase IoU is the median, so these are typical resolved cases rather than best cases. Overlays are ground truth (blue) and each model's prediction; the yellow cross marks the point prompt where the displayed frame is the prompted frame. In two of the three rows only one baseline collapses, which is the general pattern: 91.7\% of failures are single-baseline, and joint failures are resolved at the same rate (81.8\%).}
  \label{fig:failure}
\end{figure*}

Foundation model utility extends beyond average-case accuracy to behavior on difficult cases. We define a catastrophic baseline failure as a test case on which a baseline model scores IoU below 10 under point prompting, effectively no usable segmentation; failure status is determined by the baseline scores alone. Across the 5,223 test cases of the fifteen evaluation datasets, at least one baseline failed catastrophically on 1,324 cases (25.3\%). SonoBase recovered a clinically useful segmentation (IoU above 50) on 1,073 of those 1,324 cases (81.0\%), corresponding to 20.5\% of all test cases (Fig.~\ref{fig:failure}a, Supplementary Table~S22). Both baselines failed together on 110 cases (2.1\%), of which SonoBase resolved 90 (81.8\%), the same rate as for single-baseline failures, so the rescue rate is not inflated by cases where only the weaker baseline collapsed. For approximately one in five test images, baseline models produce clinically unusable output while SonoBase produces a usable segmentation.

The resolved-failure rate varied by dataset, reaching 87.5\% on RegPro (3D prostate),
81.8\% on PFUS (pelvic-floor video), and 68.2\% on Brachial-Plexus (peripheral nerve), and falling
below 7\% on ACOUSLIC, CAMUS, and BUSI, where the baselines are already competent (Supplementary Table~S22). These are anatomies and acquisition settings where general-purpose models lack the feature representations to produce meaningful segmentations. Qualitative examples (Fig.~\ref{fig:failure}c, Supplementary Fig.~S3) show that baseline failures occur on images with low tissue contrast, acoustic shadowing, unusual anatomy, and severe domain shift. SonoBase's ultrasound-specific pretraining provides robust feature representations that handle these challenging conditions.

\begin{figure*}[!tp]
  \centering
  \includegraphics[width=\textwidth]{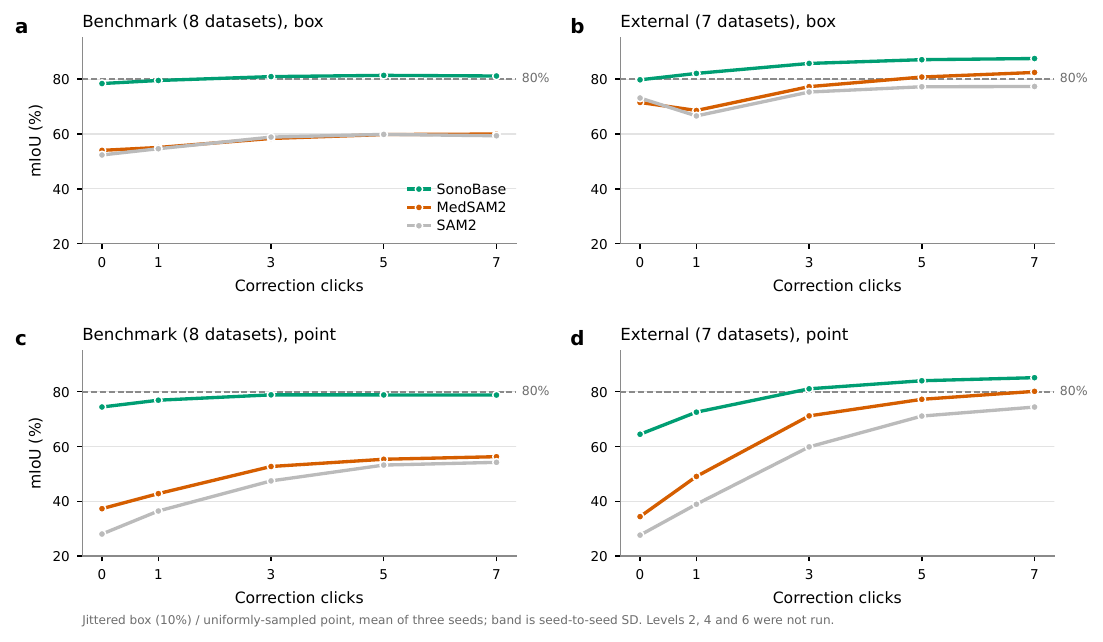}\\[6pt]
  \begin{minipage}{0.52\textwidth}
    \raggedright\textbf{e}\quad Memory propagation on CAMUS\\[3pt]
    \includegraphics[width=\textwidth]{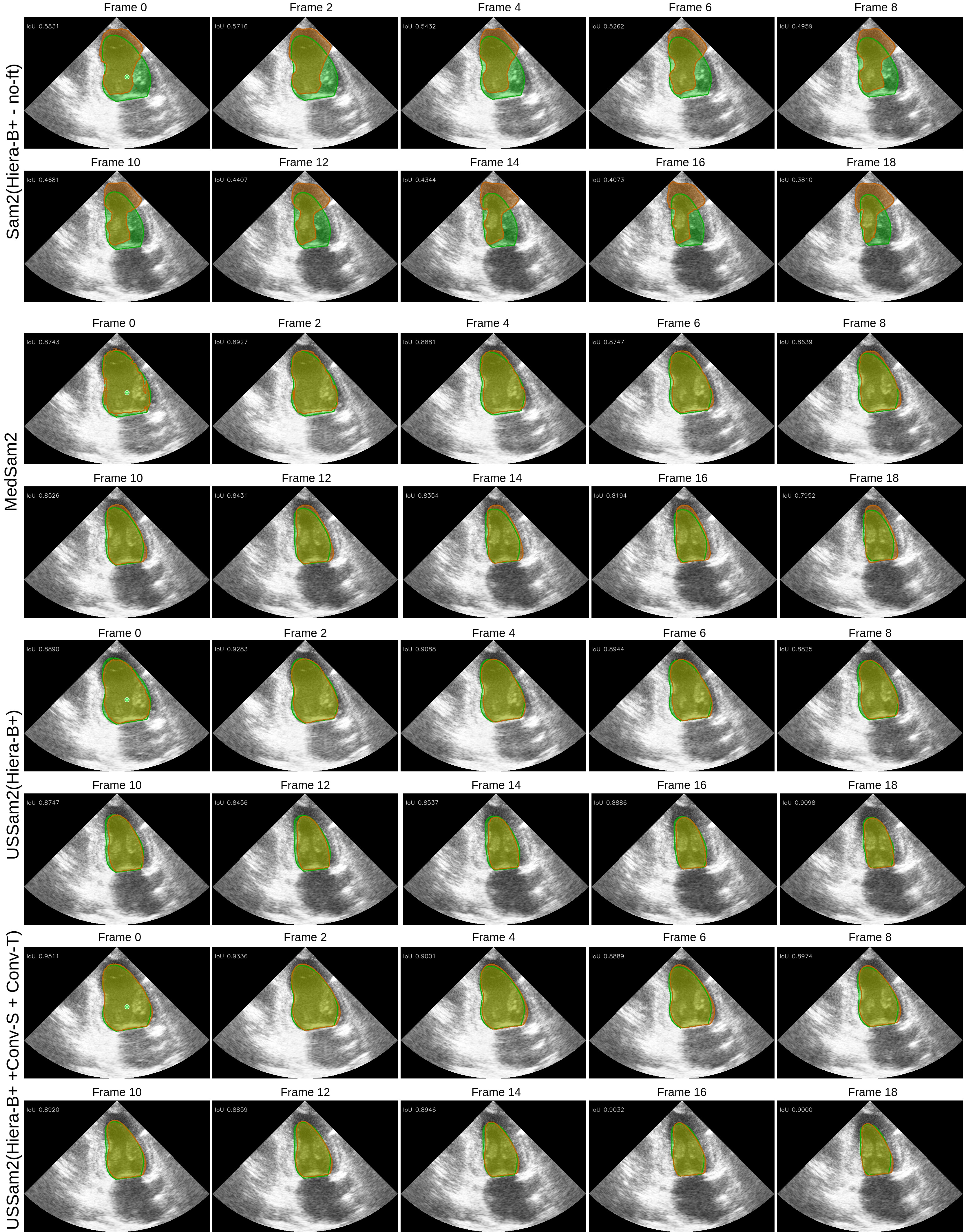}
  \end{minipage}
  \caption{\textbf{SonoBase reaches clinically usable quality with fewer corrective prompts.}
  \textbf{a--d,} Convergence of segmentation accuracy with oracle corrective clicks under the same prompt protocol as Tables~\ref{tab:benchmark}--\ref{tab:external} (10\% jittered box or uniformly sampled point; mean of three seeds, seed-to-seed s.d.\ $\leq 0.5$; measured at 0, 1, 3, 5, and 7 corrections). The dashed line marks the 80\% mIoU usability threshold.
  \textbf{a,} Benchmark, box prompt: SonoBase crosses within three measured clicks (interpolated crossing 1.67); neither baseline reaches the threshold at any budget (maxima 60.0 and 59.9).
  \textbf{b,} External, box prompt: SonoBase crosses with a single click; the first correction click degrades both baselines before they recover.
  \textbf{c,} Benchmark, point prompt: SonoBase plateaus at 78.8, just below the threshold; baselines plateau near 55.
  \textbf{d,} External, point prompt: SonoBase crosses in three clicks, MedSAM2 in seven, and SAM2 never does.
  \textbf{e,} Memory-based propagation enables a prompt-once workflow: a single interaction on one frame propagates through the video sequence (mean inter-frame IoU 0.964 on CAMUS), reducing per-frame annotation burden.}
  \label{fig:workflow}
\end{figure*}

Interactive segmentation enables clinicians to correct model predictions through additional point clicks. SonoBase reaches clinically usable segmentation quality (80\% mIoU) with fewer corrections than baselines under the same jittered, seed-averaged prompt protocol as Tables~\ref{tab:benchmark}--\ref{tab:external} (Fig.~\ref{fig:workflow}, Supplementary Table~S23). On external datasets with box initialization, SonoBase starts at 79.8 mIoU and crosses the threshold with a single corrective click (6~seconds of clinician time); MedSAM2 needs five clicks (14~seconds) and SAM2 never crosses at any tested budget. On benchmark datasets with box initialization, SonoBase crosses within three measured clicks (10~seconds; interpolated crossing 1.67), while neither baseline reaches the threshold under either prompt on this tier (box maxima 60.0 and 59.9). With point initialization on external data, SonoBase reaches the threshold in three corrections (8~seconds), whereas MedSAM2 requires seven (16~seconds) and SAM2 does not reach it at all. Notably, the first correction click \emph{degrades} both baselines under external box initialization, MedSAM2 falls from 71.5 to 68.6 mIoU and SAM2 from 73.1 to 66.6 before recovering by three clicks, while SonoBase improves monotonically: the ultrasound-adapted model integrates a mixed box-and-click prompt that the unadapted models do not. The efficiency advantage widens under domain shift, making SonoBase even more useful in new settings.

SonoBase's memory mechanism enables prompt-once propagation: a single user interaction on one frame of a video or one slice of a 3D volume propagates segmentation to the entire sequence (Fig.~\ref{fig:workflow}e). On CAMUS cardiac videos and ACOUSLIC fetal sequences, SonoBase maintained temporal consistency across sequences of 18--28 frames. This reduces per-frame interaction burden and aligns with realistic clinical workflows where clinicians expect to annotate a key frame and have the result propagate.

\subsection*{SonoBase learns transferable ultrasound representations beyond core clinical endpoints}

To establish that SonoBase learns reusable ultrasound representations rather than task-specific segmentation features, we evaluated three further dimensions of transfer: downstream dense prediction, cross-species generalization, and subgroup robustness.

\paragraph{Downstream dense prediction.} Using SonoBase's encoder as a frozen backbone for US-RF-DETR (a detection and instance segmentation framework), SonoCorpus pretraining yielded 54.5 macro detection mAP across eight datasets, against 36.5 for SAM2 and 34.3 for MedSAM2, leading on all eight (Fig.~\ref{fig:breadth}a, Supplementary Table~S26). Enabling the mask head, SonoBase reached 67.3 macro instance-segmentation mAP against 54.3 and 48.8, leading on five of the six datasets where all three backbones produced masks (losing BUS-BRA, 50.8 versus 53.9). The margin is largest where the generic backbones fail outright: on DDTI thyroid SonoBase reaches 38.5 detection and 54.6 segmentation mAP against 1.9/15.6 for SAM2 and 1.7/4.2 for MedSAM2, gaps that reflect the baselines never localizing the target rather than localizing it imprecisely. Large gains also appear on KidneyUS (63.1 detection, 80.3 segmentation) and LUMINOUS (63.7, 79.0). This confirms that SonoCorpus pretraining produces a general ultrasound feature backbone useful for tasks beyond the promptable-segmentation objective used during training, and that the benefit survives the change of output head (protocol and dataset exclusions in Supplementary Section~S10).

\paragraph{Cross-species generalization.} On a mouse brain tumor dataset~\cite{mouseBrainTumor2025} ($n=1{,}203$ frames), SonoBase achieved 47.0/64.2 mIoU (point/box), compared to 8.5/46.1
for SAM2 and 21.5/45.3 for MedSAM2 (Fig.~\ref{fig:breadth}b, Supplementary Table~S27). The 5.5-fold improvement over SAM2 on point prompts demonstrates that SonoCorpus pretraining captures fundamental ultrasound physics (boundary contrast, speckle texture, acoustic properties) that transfer across species, supporting preclinical research applications without retraining.

\paragraph{Subgroup robustness.} Ultrasound is absent from medical AI fairness literature, largely because public ultrasound datasets lack per-patient demographic metadata~\cite{fairnessMedIA2024}. We evaluated subgroup performance along the proxy fairness axes available in our datasets (Fig.~\ref{fig:breadth}c, Supplementary Table~S28). On CAMUS, SonoBase performance spread across image quality grades (good/medium/poor) was only 2.8 percentage points with point prompts (2.0 with box prompts), and the gender gap was 1.2 points. On KidneyUS, box-prompt performance was near-invariant across scanner manufacturers (1.4-point spread over the five vendors with sufficient samples for stratification). On BUSI, the benign-malignant gap (83.7 vs. 68.4 point mIoU) is clinically expected and consistent with the greater segmentation difficulty of infiltrative malignant lesions (Supplementary Table~S28). SonoCorpus's curated metadata set is designed to incorporate demographic annotations as they become available, enabling future demographic-stratified evaluation.

\subsection*{Few-shot adaptation demonstrates superior sample efficiency}
\label{sec:fewshot}

The ACOUSLIC box-prompt comparison (SonoBase 18.92~mm vs. MedSAM2 14.77~mm, $p = 7.8 \times 10^{-6}$) is the one zero-shot clinical endpoint on which a baseline leads. A foundation model's value, however, lies in how efficiently it leverages new labeled data from the target domain. We tested few-shot adaptation by fine-tuning only the decoder of each model (freezing the image encoder) on $N = \{1, 2, 5, 10, 20, 30\}$ labeled examples from each dataset's held-out adaptation pool (Methods), with 3 random seeds per $N$ value, and evaluating on the full test set. We ran this experiment on three external datasets: ACOUSLIC (primary, addressing that result), DDTI (where the zero-shot box-prompt advantage is modest, 3.7--6.4 pp over the baselines), and FUGC (where absolute performance is lowest). The pre-specified primary endpoint was ACOUSLIC box-prompt AC MAE at $N=5$ (SonoBase vs. MedSAM2), tested at $\alpha = 0.05$ without multiplicity adjustment. All secondary comparisons were corrected using the Benjamini-Hochberg procedure at FDR $q = 0.05$ (Methods).

\begin{figure*}[!tp]
  \centering
  \includegraphics[width=\textwidth]{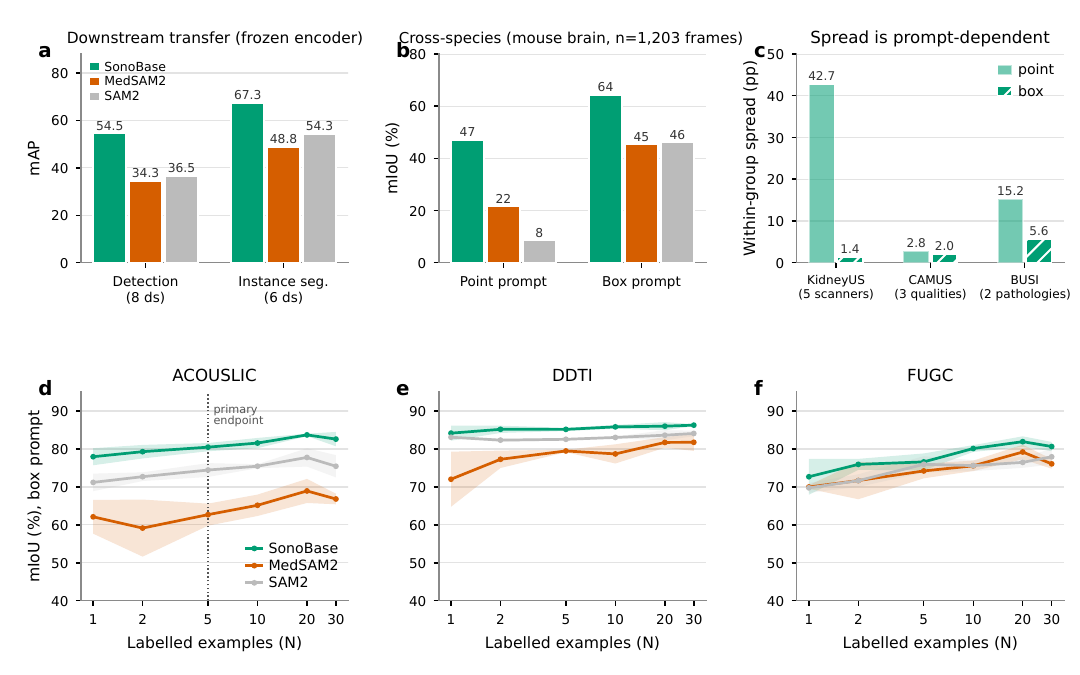}
  \caption{\textbf{SonoBase learns transferable representations and adapts efficiently to new domains.}
  \textbf{a,} Downstream dense-prediction transfer using the SonoBase encoder as a frozen backbone for US-RF-DETR, on the detection splits. Left: detection, macro bbox mAP 54.5 against 36.5 (SAM2) and 34.3 (MedSAM2), leading on all eight datasets. Right: instance segmentation with the mask head enabled, macro segm mAP 67.3 against 54.3 and 48.8, leading on five of the six datasets where all three backbones produced masks. Both axes are mAP in percentage points.
  \textbf{b,} Cross-species generalization on mouse brain tumor ($n = 1{,}203$ frames): SonoBase improves 5.5$\times$ over SAM2 under point prompting (47.0 versus 8.5) and retains an 18-point lead under box prompting (64.2 versus 46.1).
  \textbf{c,} Subgroup robustness along proxy axes. Box-prompted performance is near-invariant to image quality (2.0 pp) and scanner manufacturer (1.4 pp); point-prompted performance is not (42.7 pp across manufacturers), so invariance claims are prompt-dependent.
  \textbf{d--f,} Few-shot sample-efficiency curves on ACOUSLIC, DDTI, and FUGC, box prompt; lines are the mean over three seeds and shaded bands are $\pm$1 standard deviation across those seeds (not confidence intervals). SonoBase leads at every $N$ from a single example, and five labeled examples overturn the zero-shot baseline advantage on ACOUSLIC box prompts (17.2~mm versus 29.8~mm AC MAE, $p = 2.6\times10^{-20}$): a hospital labeling five to ten cases can unlock most of the available accuracy.}
  \label{fig:breadth}
\end{figure*}

The primary endpoint was met: at $N=5$ with box prompts on ACOUSLIC, SonoBase reached $17.17 \pm 5.52$~mm AC MAE against $29.80 \pm 16.83$~mm for identically adapted MedSAM2 ($p = 2.6 \times 10^{-20}$), converting the one zero-shot clinical endpoint on which a baseline led into a decisive advantage (Table~\ref{tab:fewshot}, Fig.~\ref{fig:breadth}d--f). The advantage is present from the first example: SonoBase leads MedSAM2 at every $N$ on ACOUSLIC and DDTI under both prompt types, and at $N=1$ already exceeds MedSAM2's $N=30$ performance in four of six dataset--prompt combinations --- a starting-point advantage inherited from ultrasound pretraining rather than a faster learning rate. Adaptation saturates by $N=5$--$10$ across models and datasets; practically, a site that labels five to ten cases captures nearly all of the available benefit. Segmentation gains do not fully propagate to derived scalar measurements: against fine-tuned SAM2 the same raw AC endpoint favors SAM2 ($12.79 \pm 0.74$~mm, $q = 7.5 \times 10^{-12}$) despite SonoBase's 6.0 mIoU lead, because ellipse fitting rewards extent rather than overlap; after identical held-out calibration the two models tie ($12.39 \pm 0.98$ versus $12.50 \pm 0.38$~mm, $p = 0.86$; Supplementary Table~S9d, Section~S13).

\begin{table*}[!t]
  \centering
  \scriptsize
  \setlength{\tabcolsep}{4pt}
  \caption{\textbf{Few-shot adaptation on three external datasets (mIoU, \%, mean $\pm$ s.d. across 3 seeds).} Mask-decoder-only fine-tuning; image encoder, prompt encoder and memory modules frozen. $N$ = number of labeled training examples drawn from each dataset's dedicated adaptation pool. All SonoBase-versus-MedSAM2 differences are significant after Benjamini-Hochberg correction except FUGC point at $N=1$. $N=0$ is the released checkpoint evaluated without adaptation under the same deterministic protocol as all other columns, which differs from the randomized-prompt protocol of Table~\ref{tab:external} (Methods).}
  \label{tab:fewshot}
  \resizebox{\linewidth}{!}{
  \begin{tabular}{llccccccc}
    \toprule
    \textbf{Dataset} & \textbf{Model} & $N=0$ & $N=1$ & $N=2$ & $N=5$ & $N=10$ & $N=20$ & $N=30$ \\
    \midrule
    \multicolumn{9}{l}{\textit{Point prompt}} \\
    ACOUSLIC & \textbf{SonoBase} & \textbf{72.9} & \textbf{72.7\,$\pm$\,1.9} & \textbf{75.4\,$\pm$\,2.3} & \textbf{77.3\,$\pm$\,1.7} & \textbf{76.8\,$\pm$\,2.9} & \textbf{80.4\,$\pm$\,0.5} & \textbf{79.8\,$\pm$\,1.5} \\
             & MedSAM2  & 49.7 & 45.2\,$\pm$\,9.8 & 47.1\,$\pm$\,13.2 & 51.4\,$\pm$\,0.5 & 56.9\,$\pm$\,3.1 & 57.5\,$\pm$\,5.8 & 59.4\,$\pm$\,2.3 \\
             & SAM2     & 21.4 & 30.3\,$\pm$\,6.4 & 46.9\,$\pm$\,12.3 & 64.6\,$\pm$\,0.8 & 63.7\,$\pm$\,1.3 & 69.3\,$\pm$\,0.1 & 65.8\,$\pm$\,3.0 \\
    DDTI     & \textbf{SonoBase} & \textbf{72.4} & \textbf{72.2\,$\pm$\,0.5} & \textbf{71.2\,$\pm$\,1.5} & \textbf{71.2\,$\pm$\,1.3} & \textbf{72.4\,$\pm$\,0.4} & \textbf{72.2\,$\pm$\,0.7} & \textbf{72.2\,$\pm$\,0.9} \\
             & MedSAM2  & 28.7 & 36.7\,$\pm$\,7.9 & 41.6\,$\pm$\,7.0 & 45.3\,$\pm$\,2.0 & 47.7\,$\pm$\,2.8 & 48.3\,$\pm$\,1.0 & 50.5\,$\pm$\,3.1 \\
             & SAM2     & 29.2 & 51.4\,$\pm$\,2.9 & 52.6\,$\pm$\,1.6 & 54.2\,$\pm$\,2.3 & 54.7\,$\pm$\,3.3 & 56.1\,$\pm$\,1.5 & 55.7\,$\pm$\,1.5 \\
    FUGC     & \textbf{SonoBase} & \textbf{45.0} & 49.9\,$\pm$\,2.4 & 51.2\,$\pm$\,6.3 & \textbf{61.3\,$\pm$\,2.5} & \textbf{61.0\,$\pm$\,0.9} & \textbf{66.5\,$\pm$\,0.9} & \textbf{65.6\,$\pm$\,1.8} \\
             & MedSAM2  & 16.1 & 50.1\,$\pm$\,5.0 & 52.0\,$\pm$\,0.9 & 53.1\,$\pm$\,6.0 & 53.0\,$\pm$\,2.1 & 60.6\,$\pm$\,1.6 & 54.7\,$\pm$\,4.5 \\
             & SAM2     & 16.6 & 19.3\,$\pm$\,2.2 & 21.2\,$\pm$\,4.1 & 53.1\,$\pm$\,1.3 & 53.5\,$\pm$\,0.9 & 59.2\,$\pm$\,2.1 & 57.5\,$\pm$\,4.0 \\
    \midrule
    \multicolumn{9}{l}{\textit{Box prompt}} \\
    ACOUSLIC & \textbf{SonoBase} & \textbf{77.6} & \textbf{78.0\,$\pm$\,2.2} & \textbf{79.3\,$\pm$\,1.8} & \textbf{80.5\,$\pm$\,1.1} & \textbf{81.6\,$\pm$\,1.4} & \textbf{83.7\,$\pm$\,0.3} & \textbf{82.6\,$\pm$\,1.9} \\
             & MedSAM2  & 62.7 & 62.1\,$\pm$\,4.4 & 59.2\,$\pm$\,7.5 & 62.7\,$\pm$\,2.9 & 65.2\,$\pm$\,2.8 & 68.9\,$\pm$\,3.2 & 66.8\,$\pm$\,1.4 \\
             & SAM2     & 71.4 & 71.2\,$\pm$\,2.3 & 72.7\,$\pm$\,1.1 & 74.5\,$\pm$\,1.8 & 75.5\,$\pm$\,0.5 & 77.8\,$\pm$\,2.4 & 75.4\,$\pm$\,2.9 \\
    DDTI     & \textbf{SonoBase} & \textbf{85.5} & \textbf{84.2\,$\pm$\,2.0} & \textbf{85.2\,$\pm$\,1.0} & \textbf{85.2\,$\pm$\,0.4} & \textbf{85.8\,$\pm$\,0.5} & \textbf{86.0\,$\pm$\,1.1} & \textbf{86.3\,$\pm$\,0.3} \\
             & MedSAM2  & 80.8 & 72.0\,$\pm$\,7.3 & 77.3\,$\pm$\,2.3 & 79.5\,$\pm$\,0.3 & 78.7\,$\pm$\,2.5 & 81.7\,$\pm$\,1.4 & 81.8\,$\pm$\,2.3 \\
             & SAM2     & 82.2 & 83.1\,$\pm$\,0.2 & 82.3\,$\pm$\,0.4 & 82.5\,$\pm$\,0.3 & 83.0\,$\pm$\,0.2 & 83.6\,$\pm$\,1.3 & 84.1\,$\pm$\,0.7 \\
    FUGC     & \textbf{SonoBase} & \textbf{65.8} & \textbf{72.7\,$\pm$\,4.7} & \textbf{75.9\,$\pm$\,1.5} & \textbf{76.6\,$\pm$\,2.2} & \textbf{80.1\,$\pm$\,0.9} & \textbf{81.9\,$\pm$\,1.4} & \textbf{80.7\,$\pm$\,1.2} \\
             & MedSAM2  & 57.2 & 70.0\,$\pm$\,0.5 & 71.7\,$\pm$\,4.9 & 74.2\,$\pm$\,1.9 & 75.6\,$\pm$\,1.4 & 79.2\,$\pm$\,1.9 & 76.1\,$\pm$\,1.3 \\
             & SAM2     & 56.9 & 69.8\,$\pm$\,1.3 & 71.7\,$\pm$\,1.7 & 76.0\,$\pm$\,2.1 & 75.6\,$\pm$\,1.4 & 76.5\,$\pm$\,1.4 & 77.9\,$\pm$\,2.0 \\
    \bottomrule
  \end{tabular}
  }
\end{table*}

\section*{Discussion}

An open, reusable ultrasound foundation model can support clinically meaningful segmentation and measurement across heterogeneous acquisition settings. SonoBase and SonoCorpus provide evidence for this claim across multiple complementary evaluation axes: broad segmentation accuracy, external generalization under named distribution shifts, clinical measurement agreement with expert-level inter-observer benchmarks, catastrophic failure resolution, interactive workflow efficiency, downstream transfer, and cross-species generalization, as well as few-shot adaptation for target tasks. The clinical significance of these results is measured at the treatment thresholds and measurement tolerances where errors have patient consequences: ICD candidacy, gestational dating, fetal growth assessment, and cancer risk stratification.

Ultrasound is the most deployed imaging modality globally yet has lacked a strong open foundation-model platform. Prior ultrasound AI efforts have followed a single-task, single-dataset paradigm that produces models unable to transfer across the acquisition variability inherent to clinical practice. General-purpose promptable models (SAM2, MedSAM2) provide architectural capability but lack the ultrasound-specific features needed for clinical-grade performance. SonoCorpus addresses the data gap by consolidating 53 fragmented public datasets with curated metadata enabling controlled evaluation. SonoBase addresses the model gap by combining ultrasound-specific pretraining at scale with a hybrid multi-scale architecture that handles the extreme scale variation characteristic of ultrasound targets. Together, they provide infrastructure that the ultrasound AI community can build on: checkpoints for immediate use, optimizer states for continual pretraining on new datasets without cold-start learning rate problems, split indices for exact reproducibility, and deduplication hashes for verifying data integrity when extending the resource.

We also find evidence from several converging lines to call SonoBase a foundation model. SonoBase performs broadly across 2D images, video sequences, and 3D volumes from 24 clinical applications. Performance is robust across scanner vendors, geographic regions, operator expertise levels, and image quality grades. The encoder transfers to downstream tasks (detection, instance segmentation) with large improvements over both SAM2 and MedSAM2, confirming that the learned representations are general rather than segmentation-specific. Cross-species transfer to preclinical imaging demonstrates that the representations capture fundamental ultrasound physics rather than only human anatomy. This breadth of evidence distinguishes a genuine foundation model from a strong task-specific model. The comparison against task-specific models makes the point directly: per-dataset nnU-Net specialists trained on the same splits do not outperform prompted SonoBase on segmentation or on the head-circumference and ejection-fraction endpoints, so generality costs no accuracy. The residual specialist advantage appears only where a dedicated model is trained on a target distribution the generalist SonoBase model was never trained on (ACOUSLIC), and this gap is removable by few-shot adaptation.

The clinical measurement results demonstrate the potential for clinical utility of the model. Ejection fraction estimates within inter-observer variability (6.63\% MAE, 4.95\% after held-out linear recalibration), with 13.0\% misclassification at the ICD-candidacy threshold, support prospective evaluation of SonoBase-derived EF as an assistive measurement tool. Head circumference accuracy (1.81~mm) with gestational age estimates accurate to 1.20 days and 0\% false workup rate confirms utility in obstetric dating. Catastrophic failure resolution in 20.5\% of test cases, 81\% of the cases where a baseline collapses, shifts the SonoBase's practical value  from incremental improvement to fundamental reliability. SonoBase scores higher on average, and also rescues cases where baseline models would produce clinically unusable output. The ACOUSLIC results demonstrate robust segmentation of blind-sweep data from handheld point-of-care devices at public health units in Sierra Leone and Tanzania, where unadapted SAM2 collapses; on the one endpoint where a baseline leads at zero shot, box-prompted abdominal circumference, the deficit is a systematic scale offset that identical held-out calibration removes, and five labeled cases establish a decisive advantage over MedSAM2. This supports the case for assistive fetal biometry in task-shifted antenatal care, where trained non-sonographer staff acquire the sweeps and sonographer time is scarce, a constraint of primary-care services in high- and low-income countries alike.

The evaluation is retrospective; prospective validation, scoped per application and clinical context, is the natural next step and is directly enabled by the released platform. The current corpus is B-mode, and extending SonoCorpus to Doppler and elastography is a continuation of the same curation pipeline. Demographic metadata is absent from nearly all public ultrasound datasets, a field-wide gap, and the SonoCorpus metadata schema already provisions the fields to close it as annotations become available.

SonoBase is designed as an interactive assistant that augments clinician measurement workflows, not an autonomous reader. SonoBase can be run on a GPU with 12-16 GB of VRAM, which enables it to be deployed on a range of portable and low-cost boards such as the nVidia Jetson. Deployment should follow application-specific prospective validation with appropriate regulatory approval. The open-release strategy (checkpoints, optimizer states, split indices, deduplication hashes, fine-tuning code) is intended to enable the ultrasound research and clinical community to extend the model to new anatomies, institutions, and clinical sites without retraining from scratch. Future work could include broader modality coverage (Doppler, elastography), more efficient multi-scale encoder designs, uncertainty quantification for flagging low-confidence predictions. The contribution of this work is infrastructure: a rigorously validated, fully open platform that enables application-specific clinical studies through released artifacts.


\section*{Methods}

\subsection*{SonoCorpus dataset assembly}

We built SonoCorpus by systematically collecting and curating publicly released ultrasound datasets from peer-reviewed publications and their associated repositories (Zenodo, Mendeley), public challenge websites, and project repositories; all 53 datasets are inventoried with their provenance in Supplementary Table~S1 and their licences in Supplementary Table~S1b, and each constituent dataset is cited here: the Benchmark and External datasets~\cite{BUSI,Brachial-Plexus,C-TRUS,CAMUS,HC18,PFUS,RegPro,TG3K,ACOUSLIC,BUS-BRA,DDTI,FUGC,KidneyUS,LUMINOUS,MMOTU-2d} and the Pretrain-tier datasets~\cite{Vitale2020abdominal,Ungi2020spine,AUL2023,BrEaST2024,BUID2023,YapUDIAT2018,BUSUC2023,BUS-UCLM,BUSIWHU2023,CardiacNet2024,GraphEcho2023,CDNet2022,CCAUS2022,CVA-Net,EchoCP,EchoNetDynamic2020,EchoNetPediatric2023,FALLMUD2018,FASS2023,FastUNet2022,GIST514_2023,JNU-IFM,LUSS2024,MISegNet2023,MUP,PPL2026,RVENet,S1breast2021,SegThy,STMUS2021,STU2019,TDSC-ABUS,TN3K2021,TNSCUI2020,ThyroidUSCineClip,UBPD2022,US-Nerve,US105_2017}. The final collection contains 53 datasets spanning 24 clinical applications, with data from 17 countries and broad diversity in anatomy, acquisition settings, and patient populations. SonoCorpus includes 456,963 images/frames and 1,626,085 segmentation masks covering three data formats: 2D images (40 datasets), video sequences (8 datasets), and 3D volumetric ultrasound (5 datasets). The dataset covers organs and lesions with large variation in shape, size, and appearance, including cardiac structures (left ventricle, epicardium, left atrium), fetal anatomy (head, abdomen), thyroid nodules, breast lesions, liver, kidney, prostate, peripheral nerves (brachial plexus), and muscles.

We curated detailed metadata when available from source documentation, including: site and country (documenting acquisition institution and geographic location for domain-shift evaluation), scanner type (manufacturer and model for vendor-shift evaluation), patient demographics (age, gender, BMI when available for fairness assessment), and acquisition protocol (modality, frequency, depth, frame rate for structured robustness evaluation). This metadata enables controlled, leakage-free data splits and targeted robustness evaluation under named distribution shifts.

For each source dataset, we retained original annotation protocols and converted annotations into the unified mask format used by SAM2. We used official train/validation/test splits when provided; otherwise, we created non-overlapping splits at the coarsest available grouping (patient $\rightarrow$ study $\rightarrow$ video/volume $\rightarrow$ frame). For video and 3D volumetric datasets, splitting was performed at the video/volume level to prevent data leakage from frame-level splitting.

\subsection*{Data integrity and deduplication}

We computed MD5 hashes of raw pixel data for each image in training and test sets. A hash table verified that each unique hash appeared exactly once within the training set and that no hash collision occurred between training and test partitions, confirming zero overlap. For datasets with patient identifiers, we verified that no patient contributed pixels to both training and test partitions. Hash values are released with project resources to enable future users to verify dataset similarity before fine-tuning. Full details are provided in Supplementary Section~S3.

We partitioned SonoCorpus into three subsets. The \textbf{Pretrain} set contains the majority of datasets with 95\%/5\% train/validation splits. The \textbf{Benchmark} set consists of eight representative datasets (Brachial-Plexus, BUSI, CAMUS, HC18, C-TRUS, PFUS, TG3K, RegPro) spanning all three modalities. For each, we used official splits when available or created 70\%/10\%/20\% train/val/test splits at the highest available grouping level. During training, we combined the training splits of Pretrain and Benchmark sets. The \textbf{External} set (ACOUSLIC, BUS-BRA, DDTI, FUGC, KidneyUS, LUMINOUS, MMOTU-3d) was completely held out from training; for each external dataset, 90\% of cases were reserved for testing and 10\% as a few-shot adaptation pool, disjoint from the test split, from which the $N$ labeled examples of the few-shot experiments are drawn.

\subsection*{SonoBase architecture}

SonoBase adapts SAM2~\cite{SAM2_2024}, a promptable segmentation foundation model for videos that extends SAM~\cite{SAM2023} with an explicit memory mechanism. SAM2 consists of an image encoder, prompt encoder, mask decoder, and memory module (memory attention and memory encoder). Given an ultrasound frame $F_t$ at time $t$, SonoBase extracts multi-scale features with our image-pyramid hybrid encoder. Following SAM2, the coarsest feature serves as the per-frame embedding for memory attention, while higher-resolution features are provided to the mask decoder for fine boundary recovery. The mask decoder predicts the segmentation mask conditioned on: the memory bank, the user prompt embedding (point or box), and the previous-frame mask logits. The memory encoder fuses the predicted mask with the unconditioned frame embedding to produce a memory representation appended to a first-in-first-out memory bank. For single images, the memory pathway is disabled; for 3D volumes, slices are treated as a temporal sequence.

\subsection*{Image-pyramid hybrid encoder}

Ultrasound targets span a large range of spatial scales, from millimeter nerve fascicles to centimeter-scale organ cross-sections. We address this with an image-pyramid hybrid encoder comprising three branches for multi-scale feature extraction. Given input frame $I \in \mathbb{R}^{H \times W}$, we construct a three-level image pyramid $\{I_1, I_2, I_3\}$ by resizing from low to high resolution. The lowest-resolution image $I_1$ is processed by a Hiera backbone~\cite{Hiera2023} ($B_1$), which is transformer-based and pretrained as part of SAM2, providing global semantic context at low computational cost. The higher-resolution images $I_2$ and $I_3$ are processed by two ConvNeXt~\cite{ConvNeXt2022} backbones ($B_2$, $B_3$) to preserve local texture and boundary details through convolutional inductive biases. The larger-capacity backbone is allocated to the low-resolution branch and lighter backbones to the high-resolution branches~\cite{piip}. In the primary variant (\textbf{Hiera-B++ConvNeXt-S+ConvNeXt-T}), the heavyweight Hiera-B+ backbone processes $I_1$, the lighter ConvNeXt-S handles $I_2$, and the lightest ConvNeXt-T is applied to $I_3$.

Cross-branch attention (CBA) units exchange information across scales at each hierarchical stage. Both Hiera and ConvNeXt produce four-stage feature hierarchies. We perform stage-wise fusion at corresponding stages. Feature maps from the three branches at stage $i$ are denoted $\{\mathbf{F}_l^i \in \mathbb{R}^{H_l W_l \times D_l}\}_{l=1}^{3}$. We update each branch feature by attending to all branches via multi-scale deformable attention~\cite{deformableDETR2021}:

\begin{equation}
  \operatorname{CBA}(\mathbf{z}_q) = \sum_{m=1}^{M} \mathbf{W}_m \left[ \sum_{l=1}^{3} \sum_{k=1}^{K} A_{mlqk} \cdot \mathbf{W}'_m \, \mathbf{F}_l^i(\phi_l(\hat{\mathbf{p}}_q) + \Delta \mathbf{p}_{mlqk}) \right]
  \label{eq:cba}
\end{equation}

\noindent where $\mathbf{z}_q$ is the query, $\hat{\mathbf{p}}_q \in [0,1]^2$ is the normalized reference point, $\phi_l$ rescales coordinates to the feature map level, $m$ indexes attention heads, $l$ indexes branches, $k$ indexes sampling points, and $\Delta \mathbf{p}_{mlqk}$ and $A_{mlqk}$ are predicted offsets and attention weights. The CBA output is fed to the next block of each branch.

Stage fusion merges features at each stage $j$ into a unified representation:

\begin{equation}
  \mathbf{S}_j = \sum_{l=1}^{3} w_l \, \mathrm{Up}(\mathrm{Linear}(\mathbf{F}_l^j))
  \label{eq:stagefusion}
\end{equation}

\noindent where $\mathrm{Linear}(\cdot)$ projects to a common channel dimension, $\mathrm{Up}(\cdot)$ upsamples via bilinear interpolation to the highest-resolution branch size, and $w_l$ is a learnable scalar initialized to 1. The resulting $\{\mathbf{S}_j\}_{j=1}^{4}$ serves as the encoder feature pyramid for SAM2.

\subsection*{Training protocol}

We initialized the Hiera branch and remaining SAM2 components (prompt encoder, mask decoder, memory module) from SAM2 pretrained weights, and initialized ConvNeXt branches from DINOv2~\cite{DINOv2_2024} weights. The training objective is:

\begin{equation}
  \mathcal{L} = 20 \, \mathcal{L}_{\mathrm{focal}} + \mathcal{L}_{\mathrm{dice}} + \mathcal{L}_{\mathrm{MAE}} + \mathcal{L}_{\mathrm{CE}}
  \label{eq:loss}
\end{equation}

\noindent where $\mathcal{L}_{\mathrm{focal}}$ and $\mathcal{L}_{\mathrm{dice}}$ supervise segmentation masks, $\mathcal{L}_{\mathrm{MAE}}$ is mean absolute error for IoU prediction, and $\mathcal{L}_{\mathrm{CE}}$ is cross-entropy for objectness prediction. We used AdamW ($\beta_1 = 0.9$, $\beta_2 = 0.999$, weight decay $0.1$) with cosine annealing. Learning rates: $3.0 \times 10^{-5}$ for the image encoder, $5.0 \times 10^{-5}$ for all remaining modules. Training ran for 20 epochs with batch size 8 per GPU on 8 NVIDIA H200 GPUs. Data augmentation: random flipping, random affine transforms, color jitter, random grayscale. During training, prompts were sampled from ground-truth masks (50\%), positive points inside the mask (25\%), or bounding boxes derived from the mask (25\%) to simulate interactive refinement.

\subsection*{Evaluation protocol}

We report mIoU under two prompt settings, using two prompt protocols. The segmentation benchmarks (Tables~\ref{tab:benchmark} and~\ref{tab:external}, Supplementary Tables~S2--S3) use the samplers SAM2 itself trains with: a point drawn uniformly from the ground-truth foreground, and a bounding box derived from the ground-truth mask with standard jitter. Uniform point sampling is stochastic, so these values are the mean of three prompt seeds, with the per-dataset seed spread reported alongside them. The clinical measurement and subgroup analyses instead use a deterministic prompt, for points, the mask pixel with maximum Euclidean distance to the mask boundary, so that a measurement attributed to one image is reproducible from that image alone. Iterative refinement follows an oracle protocol: at each correction step, a positive point is sampled from the false-negative region or a negative point from the false-positive region (Supplementary Section~S1).

\subsection*{Baseline and head-to-head protocols}

\paragraph{MedSAM2 and SAM2.} MedSAM2 uses the publicly released Hiera-Tiny checkpoint; SAM2 uses original pretrained weights with no ultrasound fine-tuning. Both are evaluated with the identical prompts and the identical scoring code as SonoBase.

\paragraph{Specialist baselines.} Task-specific models were trained on the same training split of each dataset that SonoBase's training corpus contains, run unprompted, and their masks imported into the same prediction archive so that the identical (case, frame, object) rows and the identical metric code produce every number (importer identity-tested against the archived SonoBase runs: maximum absolute difference 0). On the Benchmark tier, one nnU-Net ResEnc-M~\cite{nnUNet2021,nnUNetRevisited2024} (2D configuration, default 1{,}000-epoch trainer, single run, seed 42) was trained per dataset; video and volumetric datasets were trained and predicted frame by frame and slice by slice, so the specialist sees every frame whereas SonoBase is prompted once and propagates. For HC18 we additionally trained a DeepLabV3+ with a ResNet-50 encoder~\cite{DeepLabV3plus2018} (100 epochs, selection on validation Dice). For CAMUS we applied the public EchoNet-Dynamic left-ventricle segmentation weights~\cite{EchoNetDynamic2020}, trained on apical four-chamber views only, both zero-shot and after fine-tuning on the CAMUS training split (endocardium rows only). For ACOUSLIC, the three winning challenge solutions~\cite{ACOUSLIC} were retrained with their released code and default hyperparameters under subject-level five-fold cross-validation over the 300 public sweeps (240 training sweeps per fold; every test video in exactly one test fold), because their released weights were trained on the videos we test on; each was scored both on the annotated frames through our pipeline and, with its own frame selection, by the organizers' evaluation code. Paired comparisons against SonoBase use the deterministic prompt of Table~\ref{tab:fewshot}, bootstrap confidence intervals of the per-row and, for video datasets, per-video difference, and Benjamini--Hochberg correction over the specialist family; clinical comparisons enter one family with the ACOUSLIC pairs (Supplementary Tables~S3b, S5, S11, S14, S14b).

\paragraph{MedSAM3.} MedSAM3~\cite{MedSAM3_2025} is a detector fine-tune of SAM3 and does not expose SAM2's promptable box-to-mask head, so it cannot be driven by a box alone. We therefore evaluated it in its native text-plus-exemplar configuration (``T+I''): each dataset was assigned a single concept phrase, the ground-truth box was supplied as a geometry exemplar alongside the phrase, and the returned proposal with the highest IoU against that box was selected. This gives MedSAM3 strictly \emph{more} prompt information than SonoBase, which receives the box only; the comparison is conservative against our model rather than in its favor. We confirmed the box is actually consumed: supplying it raises BUSI Dice from 0.78 (text alone) to 0.896. We also note a property of the released version-1 LoRA that bears on interpretation: 916 adapter tensors load successfully but 1{,}378 adapter slots present in the model receive no released weights and therefore remain identity, including those on the geometry pathway. The box thus reaches an \emph{unadapted} geometry encoder, so MedSAM3's medical fine-tuning acts on the text and mask pathways rather than on box conditioning. Box handling is identical in the SB-SAM3 head-to-head, so the two SAM3-family comparisons share one protocol. BUSI is excluded because it appears in MedSAM3's training corpus and its inflated score there (0.896 versus 0.777 reported by the authors) is itself evidence of that overlap, giving a 14-dataset denominator. All MedSAM3 comparisons are reported as Dice to match the source implementation's metric.

\paragraph{SB-SAM3 head-to-head.} To test base-model dependence, we retrained the full SonoBase recipe on a SAM3.1 base using the same 46 pretraining datasets, the same splits, the same 20-epoch schedule, and the same loss. Five models are compared under a matched box-prompt protocol in which every model is handed the same box and returns a mask, scored with a single shared Dice implementation on frame-matched predictions: SAM2 (no fine-tuning), SAM3.1 (no fine-tuning), SB-SAM3 with a frozen encoder, SB-SAM3 with full-parameter fine-tuning, and SonoBase. The text-only arm is reported for the SAM3 lineage only, since SAM2-based models cannot accept text. In the prompt-composition ablation, ``T'' supplies text alone, ``TI'' supplies text and uses the ground-truth box only to select among returned proposals by IoU, and ``TI\_score'' supplies the box to the model as a geometry exemplar alongside the text.

\paragraph{SAM3-family evaluation protocol.} On video and volumetric datasets, all models in the SAM3-family comparisons are prompted on every frame and segment frames independently, because MedSAM3 exposes no mask-propagation mode; this matched per-frame protocol withholds SonoBase's memory propagation and is why values in Tables~\ref{tab:headtohead}--\ref{tab:medsam3} differ from the propagated values of Tables~\ref{tab:benchmark}--\ref{tab:external}.

\subsection*{Clinical measurement derivation}

\paragraph{Ejection fraction (CAMUS).} For each patient, we identified end-diastolic (ED) and end-systolic (ES) frames in apical four-chamber (A4C) and two-chamber (A2C) views. From segmented left ventricular endocardial contours, we computed volumes using Simpson's biplane method of discs~\cite{Lang2015}: each contour was divided into 20 equally spaced discs from mitral annulus to apex, disc volumes were summed for each view, and biplane volumes were averaged. Ejection fraction was computed as $\mathrm{EF} = (\mathrm{EDV} - \mathrm{ESV}) / \mathrm{EDV} \times 100$. Clinical reclassification was assessed at two treatment-relevant thresholds: $\mathrm{EF} \leq 40\%$ for HFrEF~\cite{McDonagh2021} and $\mathrm{EF} \leq 35\%$ for ICD candidacy~\cite{Heidenreich2022}.

\paragraph{Head circumference and gestational age (HC18).} For each image, we fit an ellipse to the largest connected component of the predicted skull mask and evaluated the root-mean-square (Euler) approximation to the ellipse perimeter, $\mathrm{HC} = \pi \sqrt{(A^2 + B^2)/2}$, where $A$ and $B$ are the full major and minor axis lengths converted to millimeters using dataset-provided pixel spacing. Applying the identical procedure to the ground-truth masks recovers the reference head circumference to 1.37~mm mean absolute error, which we report as the systematic floor of the measurement pipeline. Gestational age was derived from HC using the Hadlock formula~\cite{Hadlock1984}: $\mathrm{GA} = 8.96 + 0.540 \times \mathrm{HC}_{\mathrm{cm}} + 0.0003 \times \mathrm{HC}_{\mathrm{cm}}^3$.

\paragraph{Abdominal circumference (ACOUSLIC).} Ellipse fitting was applied to predicted fetal abdomen contours at the pixel spacing recorded in the image headers (0.28~mm/pixel), using the same root-mean-square perimeter approximation, and the per-video AC is the mean over annotated frames. Ground-truth AC was taken from the ACOUSLIC v1.1 release~\cite{ACOUSLIC}; the v1.0 circumference file reports values twice too large (full axes used in place of semi-axes) and was not used. Applying the identical procedure to the ground-truth masks recovers the reference AC to 7.39~mm mean absolute error when all annotated frames are averaged (4.5~mm on optimal-plane frames only), which we report as the floor of the measurement pipeline. Held-out linear recalibration of AC followed the protocol described for ejection fraction and was applied identically to all models.

\paragraph{Prostate volume (RegPro).} Volume was computed by voxel summation, multiplying the foreground voxel count by the voxel volume read from each case's NIfTI header. All RegPro cases have 0.8~mm isotropic spacing (voxel volume 0.512~mm$^3$); volumes are reported in milliliters.

\subsection*{Statistical analysis}

All pairwise model comparisons on clinical measurements were performed using two-sided Wilcoxon signed-rank tests on per-patient or per-image absolute errors. To control for multiple testing we applied the Benjamini-Hochberg procedure~\cite{BenjaminiHochberg1995} controlling the false discovery rate at $q = 0.05$, separately within each prompt family of eight tests (4 datasets $\times$ 2 comparisons). After correction, all comparisons remained significant except ACOUSLIC box-prompt SonoBase vs. SAM2 (corrected $p = 0.977$); ACOUSLIC box-prompt SonoBase vs. MedSAM2 is significant in MedSAM2's favor (corrected $p = 1.6 \times 10^{-5}$), a scale offset removed by held-out recalibration (Supplementary Table~S9d).

For the few-shot adaptation experiments, the primary endpoint was pre-specified as ACOUSLIC box-prompt AC MAE at $N = 5$ (SonoBase vs. MedSAM2), tested at $\alpha = 0.05$ without multiplicity adjustment. All secondary few-shot comparisons were corrected using BH-FDR at $q = 0.05$ across all test families.

Bootstrap 95\% confidence intervals were computed for all reported MAE values using 10,000 resamples with seed 42~\cite{EfronTibshirani1993}. Held-out linear recalibration (Supplementary Tables~S9b--S9d) fitted a linear map from predicted to reference values by least squares on training folds only, using 2,000 random 50/50 splits (split-half) and leave-one-out, applied identically to every model; because test labels enter the fit, it is reported as a secondary analysis alongside the raw values. Cohen's kappa~\cite{Cohen1960} was computed for EF classification agreement at 40\% and 35\% thresholds. Bland-Altman analysis~\cite{BlandAltman1986} reported mean bias and 95\% limits of agreement ($\pm 1.96 \times \mathrm{SD}$).

\subsection*{Few-shot adaptation protocol}

For each model (SonoBase, MedSAM2, SAM2), we fine-tuned only the mask decoder, freezing the image encoder, prompt encoder and memory modules, on $N$ labeled examples from the target dataset's adaptation pool (the 10\% of cases disjoint from its test split). $N$ values: 1, 2, 5, 10, 20, 30. For each $N$, training was repeated with 3 random seeds (42, 123, 456) controlling training subset selection. Hyperparameters were identical across all models: learning rate $10^{-4}$ with cosine decay, AdamW optimizer, batch size $\min(N, 4)$, epochs 50 ($N \leq 5$), 30 ($N \in \{10, 20\}$), or 20 ($N = 30$). Augmentation: random horizontal flip, rotation $\pm 15^{\circ}$, scale 0.9--1.1. Evaluation used the full test split. Results report mean $\pm$ std across seeds. All columns of Table~\ref{tab:fewshot}, including the $N=0$ reference, use a single deterministic evaluation protocol (native image resolution; exact ground-truth box or center click), so the $N=0 \rightarrow N=1$ step is a like-for-like measurement; these values therefore differ from the randomized-prompt averages of Tables~\ref{tab:benchmark}--\ref{tab:external}.

\subsection*{Catastrophic failure definition}

A catastrophic baseline failure is defined as a test case where a baseline model (SAM2 or MedSAM2) achieves IoU $< 10$ under point prompting, a threshold chosen to capture cases where a model is fundamentally unusable rather than merely suboptimal. Failure status is determined by the baseline scores alone. Among these failures we then report the fraction that SonoBase resolves, defined as SonoBase achieving IoU $> 50$ (clinically useful segmentation) on the same case.

\subsection*{Click efficiency and time estimation}

For click efficiency evaluation, we used the oracle interaction protocol (Supplementary Section~S1) and defined the 80\% mIoU threshold as clinically acceptable. Convergence curves use the same prompt protocol as Tables~\ref{tab:benchmark}--\ref{tab:external} --- 10\% jittered boxes or uniformly sampled points, averaged over three seeds --- evaluated at 0, 1, 3, 5, and 7 corrections; we report the first measured level that reaches the threshold, with the interpolated crossing as context. Time was estimated as 4 seconds per bounding box, 2 seconds per initial point, and 2 seconds per corrective click, following established interactive segmentation time benchmarks~\cite{MedSAM2024}.

\subsection*{Reproducibility and extensibility}

We release: (1) model checkpoints for all SonoBase variants, (2) optimizer states (Adam momentum and variance) to enable continual pretraining, (3) stored data split indices for exact reproduction, (4) MD5 hash values for deduplication verification, and (5) starter code for full-parameter and LoRA-based fine-tuning. Optimizer state release distinguishes this work from prior models that release only weights: users can continue pretraining on new datasets from the exact training state without cold-start problems.

\subsection*{Reporting compliance}

We completed the CLAIM checklist~\cite{CLAIM2024,CLAIMupdate2024} (Checklist for Artificial Intelligence in Medical Imaging, 2024 update) and the REFINE reporting checklist for foundation and large language models in medical research~\cite{REFINE2026}, addressing data availability, code sharing, generalization claims, and risk-of-bias considerations. Completed checklists are provided in Supplementary Section~S18.

\subsection*{Data availability}

SonoCorpus does not redistribute images or masks. It is released as a versioned manifest on Zenodo (\url{https://doi.org/10.5281/zenodo.22770825}): for each of the 53 constituent datasets, the source location, access conditions and licence as stated by the source (Supplementary Table~S1b), the curated metadata, the split indices used in this study, and MD5 checksums of every file, so that a copy obtained from the original source can be verified against the one used here. Per-dataset evaluation records generated for this study are deposited alongside. The record is maintained as a continuously updated resource, with new versions as sources move or datasets are added; the version used for this paper is fixed by its version DOI. One Pretrain-tier source (UBPD) was offline at the time of writing; its access date and checksums are recorded in the manifest. All constituent datasets are public and were de-identified by their originators before release; beyond the subject identifiers needed for subject-level splitting (Supplementary Section~S3), no DICOM headers or other patient-identifying metadata were used, and no new patient data were collected.

\subsection*{Code and model availability}

Code for SonoBase training, evaluation, and fine-tuning is available at \url{https://github.com/AlfredQin/sonobase} under the MIT licence. Pretrained weights and optimizer states are available at \url{https://huggingface.co/AlfredQin/sonobase} under the CC BY-NC 4.0 licence, reflecting the non-commercial terms of several constituent datasets (Supplementary Table~S1b).

\backmatter

\bmhead{Acknowledgements}
The work was funded by ADIA Lab, and MBZUAI through internal research grants.

\bmhead{Author contributions - CRediT}
Chao Qin (conceptualization, data curation, formal analysis, investigation, methodology, software, writing - original draft), Fahad Khan (methodology, project administration, supervision, resources, writing - review \& editing), Salman Khan (supervision, writing - review \& editing), Sarim Ather (supervision, validation, writing - review \& editing), Siddiq Anwar (supervision, validation, writing - review \& editing), Rao Anwer (methodology, supervision, writing - review \& editing), Shadab Khan (conceptualization, formal analysis, investigation, methodology, project administration, resources, software, supervision, validation, writing - original draft).

\bmhead{Competing interests}
The authors declare no competing interests.

\bibliography{bibliography}

\end{document}


\renewcommand{\thesection}{S\arabic{section}}
\renewcommand{\thetable}{S\arabic{table}}
\renewcommand{\thefigure}{S\arabic{figure}}

\begin{center}
{\LARGE\bfseries Supplementary Information}\\[10pt]
{\large\bfseries Open ultrasound foundation model for robust segmentation and clinical measurement across heterogeneous settings}\\[10pt]
Chao Qin$^{1,*}$, Fahad Shahbaz Khan$^{1}$, Salman Khan$^{1}$, Sarim Ather$^{2}$, Siddiq Anwar$^{3,4}$, Rao Muhammad Anwer$^{1}$, Shadab Khan$^{4,*}$\\[6pt]
{\small $^{1}$Mohamed bin Zayed University of Artificial Intelligence, Abu Dhabi, UAE. $^{2}$Sheikh Tahnoon Bin Mohammed Medical City (STMC), Al Ain, UAE. $^{3}$King's College Hospital London - Dubai, Dubai, UAE. $^{4}$ADIA Lab, Abu Dhabi, UAE. $^{*}$Corresponding author(s): chao.qin@mbzuai.ac.ae,  shadab@alumni.harvard.edu}
\end{center}
\vspace{6pt}

\section{Iterative Prompt Refinement Protocol}
\label{sec:iterative-refinement}

\subsection{Oracle Interaction Framework}

The iterative refinement pipeline employs an interactive oracle-based correction protocol to improve segmentation quality through targeted user feedback. The protocol operates as follows:

\subsubsection{Positive Point Sampling}

For each image, false negative regions (areas of ground truth not covered by model prediction) are identified. User points are sampled uniformly at random from these FN regions. These positive points guide the model to expand the segmentation to include missed structures.

\subsubsection{Negative Point Sampling}

False positive regions (areas of prediction not in ground truth) are similarly identified. Negative points are sampled uniformly from FP regions to refine boundaries and remove spurious predictions.

\subsubsection{Correction Iteration}

Each correction cycle provides a single point: a positive point sampled from the false-negative region, or a negative point from the false-positive region. The model processes this feedback and generates an updated segmentation. This cycle repeats until the user is satisfied (maximum 10 iterations in practice) or convergence is achieved.

\subsection{Iterative Refinement Results}

Starting performance and convergence are summarized below for Benchmark and External datasets.

\subsubsection{Benchmark Dataset Results}

SonoBase demonstrates strong initial performance and rapid convergence (all values under the
Tables~S2--S3 prompt protocol: jittered box or uniformly sampled point, mean of three seeds):
\begin{itemize}
    \item \textbf{Starting mIoU (0 corrections):} 74.4\% (point prompt), 78.4\% (box prompt)
    \item \textbf{After 7 corrections:} 78.8\% (point prompt), 81.2\% (box prompt)
    \item \textbf{Improvement:} +4.4 (point) and +2.8 (box) percentage points
\end{itemize}

In comparison, foundation models without ultrasound-specific pretraining show much lower baselines:
\begin{itemize}
    \item \textbf{SAM2 (Hiera-B+, no fine-tuning):} 28.1\% (point), 52.4\% (box) -- requires large improvements
    \item \textbf{MedSAM2:} 37.3\% (point), 54.0\% (box) -- still substantially below SonoBase baseline
\end{itemize}

\subsubsection{External Dataset Results}

External evaluation confirms strong generalization:
\begin{itemize}
    \item \textbf{Starting mIoU (0 corrections):} 64.5\% (point prompt), 79.8\% (box prompt)
    \item \textbf{After 7 corrections:} 85.1\% (point prompt), 87.5\% (box prompt)
    \item \textbf{Improvement:} +20.6 (point) and +7.8 (box) percentage points
\end{itemize}

The larger improvement on external data reflects the benefit of active correction when facing novel, out-of-distribution ultrasound appearances. SAM2 and MedSAM2 start with very low external performance and benefit substantially less from iterative correction.

\subsection{Figures}

\begin{figure}[!ht]
\centering
\includegraphics[width=\textwidth]{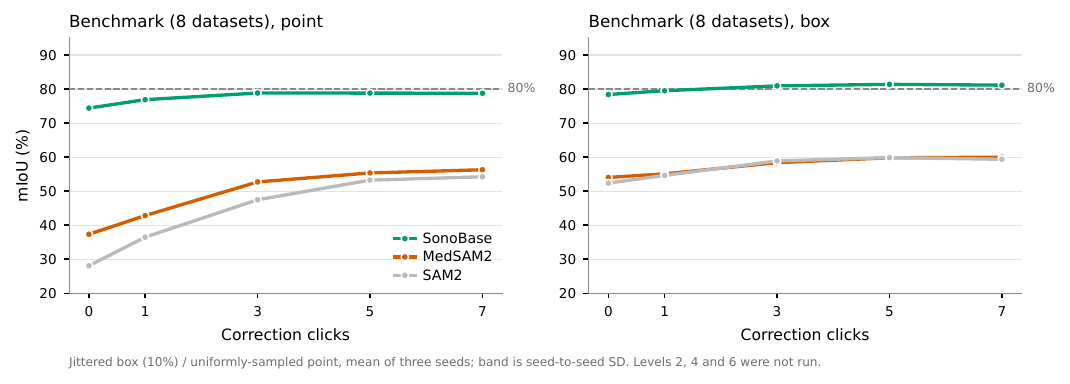}

\caption{\textbf{Iterative refinement on Benchmark datasets.} Macro mIoU over the
eight benchmark datasets against the number of oracle corrective clicks, evaluated at 0, 1, 3, 5 and
7. Left: point initialization; right: box initialization. Dashed line marks the 80\%
clinically-usable threshold. SonoBase crosses the threshold within three clicks from a box prompt;
neither baseline reaches it within the protocol on this tier. Curves are the mean of three prompt
seeds; seed-to-seed s.d.\ is $\leq 0.5$ mIoU across all cells (median 0.13) and reflects prompt-draw
variability only, so the model ordering is not an artifact of a particular prompt draw.}
\label{fig:S1}
\end{figure}

\begin{figure}[!ht]
\centering
\includegraphics[width=\textwidth]{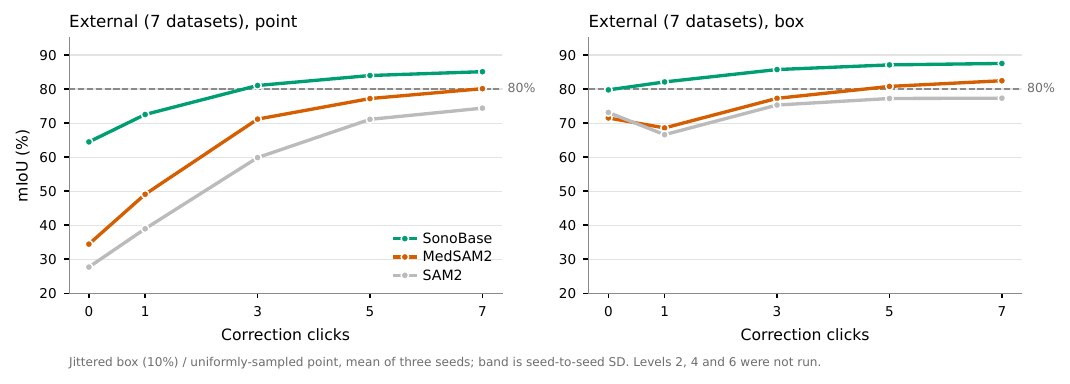}

\caption{\textbf{Iterative refinement on External datasets.} Macro mIoU over the seven
external datasets against oracle corrective clicks. Left: point initialization; right: box
initialization. Dashed line marks the 80\% threshold. SonoBase crosses first under both prompts;
MedSAM2 follows, and SAM2 does not cross under either. The first-click degradation of both baselines
under box initialization is visible in \textbf{b}. Curves are the mean of three prompt seeds;
seed-to-seed s.d.\ is $\leq 0.5$ mIoU across all cells and reflects prompt-draw variability only.}
\label{fig:S2}
\end{figure}

\newpage

\section{SonoCorpus Dataset Inventory}
\label{sec:dataset-inventory}

The SonoCorpus collection comprises 53 distinct ultrasound datasets covering diverse anatomical structures,
imaging modalities, clinical applications, and geographic origins. This comprehensive inventory ensures
reproducibility and enables future dataset curation.

\subsection{Dataset Inventory Table}

\begin{landscape}
\begin{footnotesize}
\setlength{\tabcolsep}{3pt}
\begin{longtable}{|l|p{0.16\linewidth}|l|p{0.11\linewidth}|p{0.20\linewidth}|r|r|l|}
\caption{\textbf{Complete SonoCorpus dataset inventory.} All 53 ultrasound datasets with anatomical
target, data modality, acquisition country, scanner manufacturers, deduplicated image/frame and mask
counts, and evaluation-tier assignment; licences are listed in Table~S1b. Countries and scanners were hand-curated from source
publications and dataset documentation; ``Not documented'' marks datasets whose source materials do
not record the field. Totals: 456,963 images/frames and 1,626,085 masks.
\textsuperscript{\dag}Volumetric at source but processed slice-flattened to one image per case.
\textsuperscript{\ddag}2D despite the dataset name; the source images are 2D JPG files.}
\label{tab:S1}\\
\hline
\textbf{Dataset} & \textbf{Anatomy} & \textbf{Modality} & \textbf{Country} & \textbf{Scanner} & \textbf{Images} & \textbf{Masks} & \textbf{Tier} \\
\hline
\endfirsthead
\multicolumn{8}{l}{\footnotesize\textit{Table S1 (continued)}}\\
\hline
\textbf{Dataset} & \textbf{Anatomy} & \textbf{Modality} & \textbf{Country} & \textbf{Scanner} & \textbf{Images} & \textbf{Masks} & \textbf{Tier} \\
\hline
\endhead
\hline
\endfoot
AbdomenUS & bone,  organ,  vessel & 2D image & Not documented & SonoSite & 59 & 87 & Pretrain \\
ASUS & spine & 2D image\textsuperscript{\dag} & Canada, United States & Telemed & 2,825 & 7,272 & Pretrain \\
AUL & lesion,  organ & 2D image & China & Hitachi, Aloka, Philips, SonoScape, GE, Mindray, Esaote, Toshiba, Siemens, SuperSonic Imagine (21 systems) & 735 & 1,367 & Pretrain \\
BrEaST & nodule & 2D image & Poland & Hitachi, Esaote, Samsung, Philips (5 systems) & 252 & 252 & Pretrain \\
BUID & nodule & 2D image & Iran & SuperSonic Imagine & 232 & 232 & Pretrain \\
BUS & nodule & 2D image & Spain & Siemens & 159 & 159 & Pretrain \\
BUS-UC & tumor & 2D image & Netherlands & Hitachi & 811 & 811 & Pretrain \\
BUS-UCLM & nodule & 2D image & Spain & Siemens & 263 & 263 & Pretrain \\
BUSI-WHU & nodule & 2D image & China & Not documented & 927 & 927 & Pretrain \\
CardiacNet & cardiac & 2D image\textsuperscript{\dag} & China & Philips, Hitachi & 1,550 & 5,988 & Pretrain \\
CardiacUDC & cardiac & 2D image & China & Philips, Hitachi & 1,409 & 5,633 & Pretrain \\
CDNet & cervix & 2D image & China & Esaote & 325 & 325 & Pretrain \\
Common-Carotid-Artery & vessel & 2D image & Poland & Mindray & 1,100 & 1,100 & Pretrain \\
EchoCP & chamber & 3D (slices) & China & Philips & 558 & 2,219 & Pretrain \\
EchoNet-Dynamic & cardiac & 2D image & United States & Philips, Siemens (5 systems) & 20,048 & 20,048 & Pretrain \\
EchoNet-Pediatric & cardiac & 2D image & United States & Philips, Siemens (3 systems) & 15,450 & 15,450 & Pretrain \\
FALLMUD & muscle & 2D image & France, United Kingdom & Aloka & 813 & 1,626 & Pretrain \\
FASS & organ & 2D image & Brazil & Siemens, GE, Philips & 1,563 & 6,159 & Pretrain \\
Fast-U-Net & measurement & 2D image & Not documented & Not documented & 406 & 406 & Pretrain \\
GIST514-DB & tumor & 2D image & China & Olympus & 514 & 514 & Pretrain \\
JNU-IFM & fetal & Video & China & Youkey & 5,202 & 8,945 & Pretrain \\
LUSS & artifact & 2D image & United Kingdom & GE (2 systems) & 561 & 3,766 & Pretrain \\
MI-SegNet & organ & 2D image & Germany & Siemens & 2,143 & 2,143 & Pretrain \\
MMOTU-2d & nodule & 2D image & China & Mindray & 1,469 & 1,469 & Pretrain \\
MUP & organ & 3D (slices) & United States & Not documented & 2,621 & 2,621 & Pretrain \\
PPL & lung & 2D image & China & Esaote, Hitachi & 2,756 & 2,756 & Pretrain \\
RVENet & cardiac & Video & Hungary & GE, Philips (5 systems) & 247,014 & 1,235,070 & Pretrain \\
S1 & nodule & 2D image & China & GE & 192 & 192 & Pretrain \\
SegThy & gland,  vessel & 3D (slices) & Germany & Siemens & 12,726 & 34,218 & Pretrain \\
STMUS-NDA & muscle & 2D image & Netherlands & Esaote & 8,153 & 8,153 & Pretrain \\
STU-Hospital & nodule & 2D image & China & GE & 42 & 42 & Pretrain \\
TDSC-ABUS & tumor & 3D (slices) & China & GE & 3,166 & 3,166 & Pretrain \\
ThyroidUSCineClip & thyroid & Video & United States & GE, Siemens & 17,412 & 17,412 & Pretrain \\
TN3K & nodule & 2D image & China & GE, Hitachi, Mindray & 3,493 & 3,493 & Pretrain \\
TNSCUI & nodule & 2D image & China & Mindray, Philips, Toshiba & 3,568 & 3,568 & Pretrain \\
UBPD & nerve,  vessel,  muscle & 2D image & China & Siemens, Philips & 946 & 2,780 & Pretrain \\
US-Nerve & nerve & Video & Not documented & Not documented & 2,295 & 2,295 & Pretrain \\
US105 & tumor & 2D image & Germany & GE, Toshiba & 105 & 105 & Pretrain \\
Brachial-Plexus & needle,  nerve & Video & India & Esaote, SonoSite, Butterfly & 39,025 & 32,732 & Benchmark \\
BUSI & nodule & 2D image & Egypt & GE (2 systems) & 646 & 646 & Benchmark \\
C-TRUS & organ & 2D image & Germany & Toshiba & 477 & 477 & Benchmark \\
CAMUS & cardiac & Video & France & GE & 19,232 & 57,696 & Benchmark \\
HC18 & fetal & 2D image & Netherlands & GE (2 systems) & 998 & 998 & Benchmark \\
PFUS & bone,  muscle,  organ & Video & Spain & Canon & 13,906 & 111,248 & Benchmark \\
RegPro & organ & 3D (slices) & United Kingdom & Hitachi & 4,688 & 4,688 & Benchmark \\
TG3K & gland & 2D image & China & GE & 3,585 & 3,585 & Benchmark \\
ACOUSLIC & fetal & Video & Sierra Leone, Tanzania & Telemed & 6,620 & 6,620 & External \\
BUS-BRA & nodule & 2D image & Brazil & GE, Toshiba (4 systems) & 1,875 & 1,875 & External \\
DDTI & nodule & 2D image & Colombia & Toshiba (2 systems) & 629 & 629 & External \\
FUGC & fetal & 2D image & China & GE (2 systems) & 440 & 880 & External \\
KidneyUS & organ & 2D image & Canada & Siemens, GE, Philips, Toshiba, SonoSite (9 systems) & 468 & 468 & External \\
LUMINOUS & muscle & 2D image & Canada & GE & 341 & 341 & External \\
MMOTU-3d & nodule & 2D image\textsuperscript{\ddag} & China & Mindray & 170 & 170 & External \\
\hline
\textbf{Total (53)} & & & 17 countries & & \textbf{456,963} & \textbf{1,626,085} & \\
\hline
\end{longtable}
\end{footnotesize}
\end{landscape}

The licence under which each constituent dataset is distributed, as stated by its source, is listed in
Table~S1b. SonoCorpus itself is a manifest and does not redistribute images or masks (Data Availability).

\begingroup
\renewcommand{\thetable}{S1b}
\small
\setlength{\tabcolsep}{4pt}
\begin{longtable}{@{}p{3.4cm}p{1.7cm}p{7.2cm}@{}}
\caption{\textbf{Licences of the 53 SonoCorpus constituent datasets.} Licence as stated by the distributing record, page, or file bundled with the data at the time of audit (14 September 2026). ``Not stated'' means that no licence text accompanies the data; where the source article carries a licence it is noted but does not extend to the data. SonoCorpus does not redistribute images or masks (Data Availability): each dataset is obtained from its original source under these terms, and the SonoCorpus manifest records source, access date and file checksums.}\label{tab:S1b}\\
\toprule \textbf{Dataset} & \textbf{Tier} & \textbf{Licence as stated by source} \\ \midrule \endfirsthead
\toprule \textbf{Dataset} & \textbf{Tier} & \textbf{Licence as stated by source} \\ \midrule \endhead
\bottomrule \endfoot
AbdomenUS & Pretrain & Not stated (Kaggle: original authors) \\
ASUS & Pretrain & Not stated \\
AUL & Pretrain & CC BY 4.0 \\
BrEaST & Pretrain & CC BY 4.0 \\
BUID & Pretrain & Not stated \\
BUS & Pretrain & Research-use licence agreement (application required) \\
BUS-UC & Pretrain & CC BY 4.0 \\
BUS-UCLM & Pretrain & CC BY-NC 3.0 (v1, as used; v3 is CC BY 4.0) \\
BUSI-WHU & Pretrain & CC BY 4.0 \\
CardiacNet & Pretrain & CC BY 4.0 \\
CardiacUDC & Pretrain & CC BY-NC 4.0 \\
CDNet & Pretrain & Not stated \\
Common-Carotid-Artery & Pretrain & CC BY 4.0 \\
EchoCP & Pretrain & CC BY-NC 4.0 \\
EchoNet-Dynamic & Pretrain & Research-use agreement (non-commercial) \\
EchoNet-Pediatric & Pretrain & Research-use agreement (non-commercial) \\
FALLMUD & Pretrain & Not stated \\
FASS & Pretrain & CC BY 4.0 \\
Fast-U-Net & Pretrain & Not stated (citation requested) \\
GIST514-DB & Pretrain & CC BY-NC-SA 4.0 \\
JNU-IFM & Pretrain & CC BY 4.0 \\
LUSS & Pretrain & CC BY 4.0 \\
MI-SegNet & Pretrain & Not stated \\
MMOTU-2d & Pretrain & Not stated (repository code Apache-2.0) \\
MUP & Pretrain & CC BY 4.0 \\
PPL & Pretrain & CC BY 4.0 \\
RVENet & Pretrain & Research-use agreement (non-commercial) \\
S1 & Pretrain & CC BY 4.0 (article supplement) \\
SegThy & Pretrain & CC BY (version not stated) \\
STMUS-NDA & Pretrain & CC BY 4.0 \\
STU-Hospital & Pretrain & Not stated \\
TDSC-ABUS & Pretrain & Challenge data-use agreement (participants only) \\
ThyroidUSCineClip & Pretrain & Research-use terms (non-commercial) \\
TN3K & Pretrain & MIT (dataset card) \\
TNSCUI & Pretrain & Challenge-only terms \\
UBPD & Pretrain & Not stated (portal offline at time of writing) \\
US-Nerve & Pretrain & Kaggle competition rules \\
US105 & Pretrain & Not stated \\
Brachial-Plexus & Benchmark & Non-commercial data-use statement \\
BUSI & Benchmark & Not stated (source article CC BY 4.0) \\
C-TRUS & Benchmark & Not stated \\
CAMUS & Benchmark & CC BY-NC-SA 4.0 \\
HC18 & Benchmark & CC BY 4.0 \\
PFUS & Benchmark & CC BY 4.0 \\
RegPro & Benchmark & CC BY-NC-SA 4.0 \\
TG3K & Benchmark & CC BY-NC-SA 3.0 (OpenCAS source) \\
ACOUSLIC & External & CC BY-NC-SA 4.0 \\
BUS-BRA & External & CC BY 4.0 \\
DDTI & External & Not stated ('open access' per source article) \\
FUGC & External & CC BY 4.0 \\
KidneyUS & External & CC BY-NC-SA 4.0 \\
LUMINOUS & External & Not stated (citation requested) \\
MMOTU-3d & External & Not stated (repository code Apache-2.0) \\
\end{longtable}
\endgroup
\addtocounter{table}{-1}

\newpage

\section{Data Integrity and Deduplication}
\label{sec:data-integrity}

\subsection{MD5 Hash Protocol}

All images and annotations in SonoCorpus undergo cryptographic verification via MD5 hashing. Each image
file is hashed during ingestion and the hash is recorded in a manifest. This enables:
\begin{itemize}
    \item Detection of file corruption during storage or transfer
    \item Verification that distributed versions are identical
    \item Rapid identification of duplicate files across datasets
\end{itemize}

The MD5 manifest is part of the SonoCorpus record on Zenodo (Data Availability), so that a copy obtained from an original source can be verified against the files used in this study.

\subsection{Patient-Level Split Verification}

To prevent data leakage between train/val/test splits, datasets undergo patient-level deduplication wherever patient identifiers are available; otherwise the split is made at the highest grouping level the public release supports (for TG3K, released as frames drawn from 16 videos, this is the frame level, which is why a single-dataset specialist scores near its ceiling there; Table~S3b):
\begin{enumerate}
    \item Patient identifiers (de-identified) are extracted from DICOM metadata or dataset documentation
    \item All frames/slices from the same patient are grouped
    \item Split assignment ensures no patient appears in more than one split
    \item Cross-dataset patient overlaps (rare) are documented and controlled
\end{enumerate}

This rigorous split curation prevents inflated performance estimates from test sets containing familiar patients.

\subsection{Cross-Dataset Overlap Verification}

Despite 53 datasets originating from different institutions, some datasets share common subjects or
overlapping acquisition protocols. A graph-based deduplication check identifies:
\begin{itemize}
    \item Datasets from the same clinical trial (grouped together)
    \item Multi-site studies with coordinated protocols (noted in metadata)
    \item Explicit overlaps (very rare; documented when detected)
\end{itemize}

These relationships are documented in the dataset inventory (Table S1).

\subsection{Split Indices Release}

The train/val/test split assignments for all datasets will be released with SonoCorpus, enabling complete
reproducibility and future benchmark studies. Split indices are stored as text files listing image
identifiers for each split.

\newpage

\section{Full Benchmark and External Results}
\label{sec:full-results}

This section provides comprehensive quantitative results across all dataset and model combinations.

\subsection{Benchmark Datasets: Complete Results}

\begin{table}[!ht]
\small
\setlength{\tabcolsep}{3pt}
\centering
\caption{\textbf{Segmentation results on Benchmark datasets (point and box prompts).}
mIoU and Dice percentages for all models. Hybrid: best-performing hybrid encoder
architecture (3-branch with cross-attention). Each dataset row shows results for point initialization (top) and box initialization (bottom).}
\label{tab:S2}
\begin{tabular}{|c|c|cc|cc|cc|}
\hline
\textbf{Dataset} & \textbf{Prompt} & \multicolumn{2}{c|}{\textbf{SonoBase Hybrid}} &
\multicolumn{2}{c|}{\textbf{MedSAM2}} & \multicolumn{2}{c|}{\textbf{SAM2 Hiera-B+}} \\
& & \textbf{mIoU} & \textbf{Dice} & \textbf{mIoU} & \textbf{Dice} & \textbf{mIoU} & \textbf{Dice} \\
\hline
BUSI & Point & \textbf{78.7$\pm$0.3} & \textbf{86.7$\pm$0.1} & 67.6$\pm$1.0 & 77.7$\pm$1.1 & 59.6$\pm$0.6 & 68.9$\pm$0.7 \\
& Box & \textbf{84.0$\pm$0.1} & \textbf{91.0$\pm$0.0} & 81.2$\pm$0.3 & 89.3$\pm$0.2 & 80.3$\pm$0.2 & 88.5$\pm$0.2 \\
Brachial-Plexus & Point & \textbf{60.5$\pm$0.8} & \textbf{72.5$\pm$1.0} & 11.5$\pm$1.1 & 15.5$\pm$1.2 & 12.6$\pm$1.1 & 18.3$\pm$1.5 \\
& Box & \textbf{62.1$\pm$0.3} & \textbf{74.2$\pm$0.3} & 26.2$\pm$0.8 & 34.7$\pm$1.1 & 41.5$\pm$0.8 & 51.4$\pm$0.8 \\
C-TRUS & Point & \textbf{59.3$\pm$0.6} & \textbf{72.5$\pm$0.4} & 24.2$\pm$1.8 & 35.7$\pm$2.3 & 10.3$\pm$0.5 & 15.6$\pm$0.5 \\
& Box & \textbf{67.1$\pm$0.4} & \textbf{79.3$\pm$0.3} & 46.4$\pm$0.6 & 61.2$\pm$0.6 & 49.8$\pm$0.3 & 64.1$\pm$0.2 \\
CAMUS & Point & \textbf{81.1$\pm$0.5} & \textbf{88.1$\pm$0.5} & 69.2$\pm$0.2 & 79.1$\pm$0.2 & 15.1$\pm$0.1 & 25.5$\pm$0.2 \\
& Box & \textbf{84.3$\pm$0.2} & \textbf{91.1$\pm$0.2} & 77.9$\pm$0.2 & 87.0$\pm$0.1 & 56.2$\pm$0.1 & 68.2$\pm$0.1 \\
HC18 & Point & \textbf{94.8$\pm$0.0} & \textbf{97.2$\pm$0.0} & 60.6$\pm$0.2 & 67.7$\pm$0.0 & 56.6$\pm$0.9 & 67.9$\pm$0.9 \\
& Box & \textbf{94.9$\pm$0.1} & \textbf{97.4$\pm$0.0} & 90.8$\pm$0.0 & 95.0$\pm$0.0 & 81.3$\pm$0.5 & 86.6$\pm$0.5 \\
PFUS & Point & \textbf{59.8$\pm$0.5} & \textbf{72.0$\pm$0.6} & 21.1$\pm$0.4 & 31.3$\pm$0.4 & 7.5$\pm$0.4 & 13.0$\pm$0.5 \\
& Box & \textbf{70.9$\pm$0.3} & \textbf{82.2$\pm$0.2} & 48.0$\pm$0.1 & 61.7$\pm$0.2 & 37.9$\pm$0.1 & 48.0$\pm$0.0 \\
RegPro & Point & \textbf{72.2$\pm$0.2} & \textbf{80.5$\pm$0.2} & 4.8$\pm$0.2 & 6.9$\pm$0.3 & 23.9$\pm$0.1 & 36.0$\pm$0.1 \\
& Box & \textbf{75.4$\pm$0.2} & \textbf{83.5$\pm$0.2} & 1.7$\pm$0.1 & 2.5$\pm$0.2 & 2.8$\pm$0.1 & 4.1$\pm$0.2 \\
TG3K & Point & \textbf{89.0$\pm$0.1} & \textbf{94.0$\pm$0.0} & 39.7$\pm$1.2 & 52.0$\pm$1.4 & 39.1$\pm$0.9 & 49.2$\pm$1.2 \\
& Box & \textbf{88.5$\pm$0.0} & \textbf{93.7$\pm$0.0} & 60.1$\pm$0.2 & 73.9$\pm$0.2 & 69.0$\pm$0.5 & 80.4$\pm$0.4 \\
\hline
\textbf{Average} & Point & \textbf{74.4$\pm$0.4} & \textbf{82.9$\pm$0.4} & 37.3$\pm$0.7 & 45.8$\pm$0.9 & 28.1$\pm$0.6 & 36.8$\pm$0.7 \\
& Box & \textbf{78.4$\pm$0.2} & \textbf{86.6$\pm$0.2} & 54.0$\pm$0.3 & 63.1$\pm$0.3 & 52.4$\pm$0.3 & 61.4$\pm$0.3 \\
\hline
\end{tabular}
\end{table}

\subsection{External Datasets: Complete Results}

\begin{table}[!ht]
\small
\setlength{\tabcolsep}{3pt}
\centering
\caption{\textbf{Segmentation results on External (out-of-distribution) datasets (point and box prompts).}
mIoU and Dice percentages for all models using best-performing configurations from Table S2.}
\label{tab:S3}
\begin{tabular}{|c|c|cc|cc|cc|}
\hline
\textbf{Dataset} & \textbf{Prompt} & \multicolumn{2}{c|}{\textbf{SonoBase Hybrid}} &
\multicolumn{2}{c|}{\textbf{MedSAM2}} & \multicolumn{2}{c|}{\textbf{SAM2 Hiera-B+}} \\
& & \textbf{mIoU} & \textbf{Dice} & \textbf{mIoU} & \textbf{Dice} & \textbf{mIoU} & \textbf{Dice} \\
\hline
ACOUSLIC & Point & \textbf{73.7$\pm$0.3} & \textbf{80.9$\pm$0.3} & 49.3$\pm$0.7 & 59.4$\pm$0.9 & 21.0$\pm$0.0 & 33.6$\pm$0.1 \\
& Box & \textbf{77.2$\pm$0.2} & \textbf{83.5$\pm$0.2} & 61.8$\pm$0.3 & 70.3$\pm$0.3 & 70.3$\pm$0.7 & 78.6$\pm$0.9 \\
BUS-BRA & Point & \textbf{81.1$\pm$0.0} & \textbf{88.8$\pm$0.0} & 44.3$\pm$0.5 & 55.6$\pm$0.5 & 57.6$\pm$0.2 & 66.3$\pm$0.2 \\
& Box & \textbf{84.4$\pm$0.1} & \textbf{91.3$\pm$0.1} & 76.1$\pm$0.2 & 85.8$\pm$0.1 & 83.9$\pm$0.1 & 90.9$\pm$0.1 \\
DDTI & Point & \textbf{74.0$\pm$0.2} & \textbf{83.4$\pm$0.2} & 25.8$\pm$0.8 & 37.0$\pm$0.8 & 27.2$\pm$0.3 & 37.7$\pm$0.4 \\
& Box & \textbf{82.6$\pm$0.1} & \textbf{90.3$\pm$0.1} & 76.2$\pm$0.3 & 86.3$\pm$0.2 & 78.9$\pm$0.4 & 87.9$\pm$0.3 \\
FUGC & Point & \textbf{41.1$\pm$0.2} & \textbf{54.1$\pm$0.2} & 10.8$\pm$0.4 & 17.5$\pm$0.6 & 16.1$\pm$0.0 & 27.3$\pm$0.0 \\
& Box & \textbf{63.0$\pm$0.2} & \textbf{76.0$\pm$0.1} & 54.8$\pm$0.2 & 69.5$\pm$0.2 & 56.1$\pm$0.2 & 70.9$\pm$0.2 \\
KidneyUS & Point & \textbf{61.0$\pm$1.7} & \textbf{70.2$\pm$1.3} & 39.4$\pm$0.4 & 52.9$\pm$0.5 & 27.2$\pm$0.4 & 40.4$\pm$0.5 \\
& Box & \textbf{87.9$\pm$0.1} & \textbf{93.4$\pm$0.1} & 79.7$\pm$0.0 & 88.5$\pm$0.0 & 81.6$\pm$0.3 & 89.5$\pm$0.2 \\
LUMINOUS & Point & \textbf{54.6$\pm$1.4} & \textbf{68.3$\pm$1.5} & 37.0$\pm$0.7 & 51.6$\pm$0.7 & 15.0$\pm$0.0 & 25.8$\pm$0.0 \\
& Box & \textbf{80.4$\pm$0.4} & \textbf{89.0$\pm$0.3} & 76.4$\pm$0.3 & 86.5$\pm$0.2 & 67.3$\pm$0.5 & 79.8$\pm$0.4 \\
MMOTU-3d & Point & \textbf{65.9$\pm$1.6} & \textbf{75.9$\pm$1.7} & 34.5$\pm$1.9 & 44.0$\pm$2.4 & 29.7$\pm$0.8 & 42.0$\pm$0.9 \\
& Box & \textbf{82.9$\pm$0.3} & \textbf{90.3$\pm$0.2} & 75.7$\pm$0.1 & 85.7$\pm$0.1 & 73.9$\pm$0.6 & 84.0$\pm$0.4 \\
\hline
\textbf{Average} & Point & \textbf{64.5$\pm$0.8} & \textbf{74.5$\pm$0.8} & 34.4$\pm$0.8 & 45.4$\pm$0.9 & 27.7$\pm$0.3 & 39.0$\pm$0.3 \\
& Box & \textbf{79.8$\pm$0.2} & \textbf{87.7$\pm$0.2} & 71.5$\pm$0.2 & 81.8$\pm$0.2 & 73.1$\pm$0.4 & 83.1$\pm$0.4 \\
\hline
\end{tabular}
\end{table}

\subsection{Task-Specific Specialist Baselines on the Benchmark Tier}

To test whether a single promptable model gives up accuracy relative to dedicated models, one nnU-Net
ResEnc-M (2D configuration, default 1,000-epoch trainer, single run) was trained per Benchmark dataset on
the same training split that SonoBase's training corpus contains, run unprompted on the test split, and its
masks imported into the paper's prediction archive so that the identical (case, frame, object) rows and the
identical metric code produce every number below (importer identity-tested against the archived SonoBase
runs: maximum absolute difference 0 on HC18, Brachial-Plexus, CAMUS and ACOUSLIC). Video and volumetric
datasets were trained and predicted frame by frame or slice by slice, so the specialist sees every frame,
whereas SonoBase is prompted once per object and propagates. Paired differences use the deterministic
(exact) prompt so that each row has a single SonoBase value; the jittered three-seed values of Tables~S2
and~1 are shown beside them.

\begin{landscape}
\begin{table*}[!ht]
\renewcommand{\thetable}{S3b}
\small
\setlength{\tabcolsep}{3pt}
\centering
\caption{\textbf{Per-dataset nnU-Net specialist versus prompted SonoBase on identical test rows (mIoU, \%).}
SonoBase cells give the jittered three-seed value of Table~1 followed by the deterministic exact-prompt value.
$\Delta$ is nnU-Net minus SonoBase (exact prompt) in IoU points with a bootstrap 95\% CI, computed per row and,
for video and volumetric datasets, per video or volume; negative values favor SonoBase. $^{*}$: $q<0.05$
after Benjamini--Hochberg correction over the family of paired Wilcoxon tests (the per-volume RegPro test is
not computed at $n=8$). $^{\ddagger}$HC18: the rank test is significant although the mean difference is zero,
because nnU-Net is marginally higher on most images while one gross nnU-Net failure (image 29; Table~S11)
gives SonoBase a large lead on a single image. TG3K's public split is frame-level within 16 videos, so a
single-dataset model memorizes near-duplicate frames; the Brachial-Plexus gap is the needle (nnU-Net 51.5
versus SonoBase box 21.1 IoU on needle rows), which first-frame box propagation loses, while the nerve
differs by three points.}
\label{tab:S3b}
\begin{tabular}{|l|r|c|c|c|c|c|}
\hline
\textbf{Dataset} & \textbf{Rows (videos)} & \textbf{nnU-Net} & \textbf{SonoBase box} & \textbf{SonoBase point} &
\textbf{$\Delta$ box [95\% CI]} & \textbf{$\Delta$ point [95\% CI]} \\
 & & \textbf{(unprompted)} & \textbf{jitter / exact} & \textbf{jitter / exact} & & \\
\hline
BUSI & 130 & 76.8 & 84.0 / 84.8 & 78.7 / 78.8 & $-8.0$ [$-11.7$, $-4.9$]$^{*}$ & $-1.9$ [$-5.4$, $1.1$] \\
Brachial-Plexus & 6,905 (44) & 66.7 & 62.1 / 62.8 & 60.5 / 59.5 & $+3.9$ [$3.4$, $4.5$]$^{*}$ & $+7.2$ [$6.6$, $7.8$]$^{*}$ \\
\quad per video & & & & & $+1.1$ [$-3.7$, $5.6$] & $+4.1$ [$-1.0$, $9.4$] \\
C-TRUS & 97 & 59.5 & 67.1 / 68.5 & 59.3 / 61.4 & $-9.0$ [$-12.1$, $-5.8$]$^{*}$ & $-1.9$ [$-4.4$, $0.7$] \\
CAMUS & 11,595 (200) & 83.7 & 84.3 / 85.8 & 81.1 / 83.6 & $-2.1$ [$-2.3$, $-2.0$]$^{*}$ & $+0.1$ [$0.0$, $0.2$]$^{*}$ \\
\quad per video & & & & & $-2.2$ [$-2.6$, $-1.8$]$^{*}$ & $+0.1$ [$-0.3$, $0.5$] \\
HC18 & 201 & 95.7 & 94.9 / 95.7 & 94.8 / 94.8 & $0.0$ [$-0.8$, $0.5$]$^{*\ddagger}$ & $+0.9$ [$0.4$, $1.4$]$^{*}$ \\
PFUS & 25,136 (22) & 66.1 & 70.9 / 72.2 & 59.8 / 64.3 & $-6.1$ [$-6.3$, $-5.9$]$^{*}$ & $+1.9$ [$1.7$, $2.0$]$^{*}$ \\
\quad per video & & & & & $-6.4$ [$-7.9$, $-4.8$]$^{*}$ & $+1.8$ [$0.0$, $3.5$] \\
RegPro & 497 (8) & 63.5 & 75.4 / 75.4 & 72.2 / 72.3 & $-12.0$ [$-13.8$, $-10.1$]$^{*}$ & $-8.8$ [$-10.8$, $-6.9$]$^{*}$ \\
\quad per volume & & & & & $-11.5$ [$-17.7$, $-5.7$] & $-8.0$ [$-15.5$, $-0.5$] \\
TG3K & 718 & 98.5 & 88.5 / 90.4 & 89.0 / 88.8 & $+8.2$ [$7.8$, $8.6$]$^{*}$ & $+9.7$ [$9.3$, $10.3$]$^{*}$ \\
\hline
\textbf{Macro (8)} & & \textbf{76.3} & \textbf{78.4 / 79.5} & \textbf{74.4 / 75.4} & & \\
\hline
\end{tabular}

\end{table*}
\end{landscape}
\addtocounter{table}{-1}

Against the box prompt the specialist loses on five datasets, is within a point on HC18 and, at the video
level, on Brachial-Plexus, and wins on TG3K; against the point prompt it is at parity on BUSI, C-TRUS and CAMUS, ahead by one to two points on HC18 and
PFUS, ahead on Brachial-Plexus ($+7.2$ per row; $+4.1$ [$-1.0$, $9.4$] per video) and TG3K, and behind on
RegPro. The specialist's worst cases are localization
failures (extra or wrong regions on HC18, PFUS and RegPro), which is what a prompt removes; where the prompt
carries less information (a point) or the object is thin and moving (the needle), the per-frame specialist
catches up or passes. Specialist results on the clinical endpoints are reported beside the corresponding
tables (Tables~S5, S11, S14 and S14b) and their bootstrap intervals in Table~S21.

\newpage

\section{Clinical Measurement Analyses}
\label{sec:clinical-validation}

This section provides complete clinical measurement results across four key measurement scenarios:
left ventricular ejection fraction (CAMUS), fetal head circumference and gestational age (HC18),
abdominal circumference (ACOUSLIC), and prostate volume (RegPro).

\subsection{Ejection Fraction Estimation (CAMUS)}
\label{subsec:ef-results}

Ejection fraction (EF) is calculated as $(LV_{end-diastole} - LV_{end-systole}) / LV_{end-diastole}$.
Clinical accuracy is critical for cardiac function assessment.

\subsubsection{Overall EF Performance (Point Prompt)}

\begin{table}[!ht]
\footnotesize
\setlength{\tabcolsep}{3pt}
\centering
\caption{\textbf{Ejection fraction overall results (point prompt).} Mean absolute error (MAE),
Pearson correlation (r), bias, limits of agreement (LoA), and reclassification errors at clinical
thresholds (40\% and 35\%). $n=100$ patients.}
\label{tab:S4}
\begin{tabular}{|c|cccc|ccc|}
\hline
\textbf{Model} & \textbf{MAE} & \textbf{Std} & \textbf{r} & \textbf{Bias} & \textbf{LoA} &
\textbf{Reclass @40\%} & \textbf{Reclass @35\%} \\
\hline
SonoBase & \textbf{8.86} & 5.66 & \textbf{0.861} & +8.37 & [-4.10, 20.84] & \textbf{18.0\%} & \textbf{13.0\%} \\
MedSAM2 & 12.80 & 12.79 & 0.463 & -9.77 & [-39.66, 20.13] & 35.0\% & 24.0\% \\
SAM2 (no FT) & 45.01 & 12.82 & 0.094 & -45.01 & [-70.14, -19.88] & 72.0\% & 80.0\% \\
\textit{GT masks (control)} & \textit{8.06} & --- & \textit{0.943} & --- & --- & --- & --- \\
\hline
\end{tabular}
\end{table}

SonoBase achieves superior accuracy with MAE 8.86\% and correlation 0.861,
substantially outperforming MedSAM2 (MAE 12.80\%, r=0.463) and SAM2 (MAE 45.01\%, r=0.094). Reclassification errors (misclassification
relative to clinical thresholds) are 18.0\% at the 40\% threshold and 13.0\% at the 35\% threshold
commonly used for systolic dysfunction diagnosis.

\subsubsection{Overall EF Performance (Box Prompt)}

\begin{table}[!ht]
\scriptsize
\setlength{\tabcolsep}{2pt}
\centering
\caption{\textbf{Ejection fraction overall results (box prompt), ASE disc-pairing primary.} Same
metrics as Table S4 with bounding-box initialization. The lower block reports the view-averaged
pairing convention as a sensitivity comparator: for SonoBase the two conventions agree
to within 0.07 pp of MAE and 0.002 of $r$; the baselines diverge further (up to 1.60 pp and 0.027 for
SAM2). The model ranking is preserved under both conventions, so no conclusion depends on the choice.
Specialist rows (unprompted; Methods): a CAMUS-trained nnU-Net ResEnc-M is statistically indistinguishable
from box-prompted SonoBase (paired Wilcoxon on absolute EF errors, $q = 0.39$); the EchoNet-Dynamic
segmentation network fine-tuned on the CAMUS training split is also indistinguishable ($q = 0.95$);
$^{\dagger}$the same network's public weights, trained on apical four-chamber views only, applied zero-shot
to both CAMUS views ($n=99$; one patient without a usable two-chamber volume) segment the endocardium at
Dice 0.85 (four-chamber) and 0.79 (two-chamber) yet give unusable EF ($q = 2.7\times10^{-10}$ versus
SonoBase), an in-house instance of the single-task transfer failure described in the Introduction.}
\label{tab:S5}
\begin{tabularx}{\textwidth}{|X|cccc|ccc|}
\hline
\textbf{Model} & \textbf{MAE} & \textbf{Std} & \textbf{r} & \textbf{Bias} & \textbf{LoA} &
\textbf{Reclass @40\%} & \textbf{Reclass @35\%} \\
\hline
\multicolumn{8}{|l|}{\textit{ASE disc-pairing (primary)}} \\
\hline
SonoBase & \textbf{6.63} & 5.01 & \textbf{0.867} & +5.53 & [-6.65, 17.71] & \textbf{16.0\%} & \textbf{13.0\%} \\
MedSAM2 & 10.35 & 10.89 & 0.565 & -7.39 & [-33.06, 18.29] & 27.0\% & 18.0\% \\
SAM2 (no FT) & 17.41 & 15.55 & 0.368 & -10.03 & [-51.44, 31.38] & 41.0\% & 42.0\% \\
\hline
\multicolumn{8}{|l|}{\textit{Specialists, unprompted, CAMUS-trained (see caption)}} \\
\hline
nnU-Net & 6.28 & 4.40 & 0.853 & +3.97 & [-8.93, 16.86] & 16.0\% & 11.0\% \\
EchoNet (fine-tuned) & 7.43 & 6.10 & 0.713 & +2.11 & [-16.33, 20.55] & 16.0\% & 16.0\% \\
EchoNet (zero-shot)$^{\dagger}$ & 16.92 & 19.47 & 0.227 & -1.24 & [-51.85, 49.36] & 34.3\% & 27.3\% \\
\hline
\multicolumn{8}{|l|}{\textit{View-averaged volumes (sensitivity comparator)}} \\
\hline
SonoBase & 6.70 & 4.94 & 0.869 & +5.60 & [-6.48, 17.69] & 18.0\% & 13.0\% \\
MedSAM2 & 9.99 & 10.36 & 0.560 & -6.89 & [-31.70, 17.92] & 26.0\% & 18.0\% \\
SAM2 (no FT) & 15.81 & 11.96 & 0.395 & -7.88 & [-43.63, 27.88] & 42.0\% & 40.0\% \\
\hline
\end{tabularx}
\end{table}

Box prompts improve SonoBase performance (MAE 6.63\%, r=0.867), lowering the 40\%-threshold
reclassification error from 18.0\% to 16.0\% while the 35\%-threshold error is unchanged at 13.0\%.
Notably, SAM2 improves substantially with box initialization (from
MAE 45.01\% to 17.41\%), indicating the model can leverage structural guidance but lacks ultrasound-specific
pretraining for point-based refinement.

\subsubsection{EF Stratified by Image Quality (Point Prompt)}

\begin{table}[!ht]
\small
\setlength{\tabcolsep}{3pt}
\centering
\caption{\textbf{Ejection fraction stratified by image quality (point prompt).} Performance
varies with sonographer experience and patient characteristics. Quality scores (Good/Medium/Poor)
assigned by expert cardiologists. Point initialization results shown.}
\label{tab:S6}
\begin{tabular}{|c|c|cc|cc|cc|}
\hline
\textbf{Quality} & \textbf{Model} & \textbf{MAE} & \textbf{Std} & \textbf{r} & \textbf{Bias} & \textbf{Reclass @40\%} & \textbf{Reclass @35\%} \\
\hline
\multirow{3}{*}{Good} & SonoBase & 9.25 & 5.10 & 0.890 & +9.03 & 23.1\% & 12.8\% \\
& MedSAM2 & 13.90 & 15.16 & 0.358 & -10.77 & 25.6\% & 25.6\% \\
& SAM2 & 45.53 & 13.06 & 0.153 & -45.53 & 71.8\% & 82.1\% \\
\hline
\multirow{3}{*}{Medium} & SonoBase & 8.49 & 5.92 & 0.838 & +7.92 & 11.1\% & 11.1\% \\
& MedSAM2 & 11.88 & 12.41 & 0.495 & -8.73 & 40.0\% & 17.8\% \\
& SAM2 & 45.23 & 12.21 & -0.119 & -45.23 & 77.8\% & 82.2\% \\
\hline
\multirow{3}{*}{Poor} & SonoBase & 8.92 & 6.45 & 0.853 & +8.03 & 25.0\% & 18.8\% \\
& MedSAM2 & 12.69 & 6.34 & 0.732 & -10.24 & 43.8\% & 37.5\% \\
& SAM2 & 43.11 & 14.52 & -0.018 & -43.11 & 56.2\% & 68.8\% \\
\hline
\end{tabular}
\end{table}

Performance gracefully degrades with image quality, but SonoBase maintains strong accuracy (MAE 8.92\%
even in poor quality). This robustness is crucial for clinical deployment where image quality varies.

\subsubsection{EF Stratified by Image Quality (Box Prompt)}

\begin{table}[!ht]
\small
\setlength{\tabcolsep}{3pt}
\centering
\caption{\textbf{Ejection fraction stratified by image quality (box prompt).} Same stratification
as Table S6 but with bounding box initialization. Box prompts improve SonoBase and SAM2 at every quality level; MedSAM2 is the exception, degrading on poor-quality images.}
\label{tab:S7}
\begin{tabular}{|c|c|cc|cc|cc|}
\hline
\textbf{Quality} & \textbf{Model} & \textbf{MAE} & \textbf{Std} & \textbf{r} & \textbf{Bias} & \textbf{Reclass @40\%} & \textbf{Reclass @35\%} \\
\hline
\multirow{3}{*}{Good} & SonoBase & 7.03 & 4.57 & 0.898 & +6.54 & 20.5\% & 12.8\% \\
& MedSAM2 & 9.10 & 9.45 & 0.560 & -4.96 & 17.9\% & 15.4\% \\
& SAM2 & 14.62 & 14.02 & 0.335 & -1.12 & 30.8\% & 28.2\% \\
\hline
\multirow{3}{*}{Medium} & SonoBase & 6.13 & 5.26 & 0.852 & +4.98 & 13.3\% & 13.3\% \\
& MedSAM2 & 10.14 & 12.29 & 0.522 & -8.02 & 33.3\% & 17.8\% \\
& SAM2 & 19.31 & 17.41 & 0.369 & -14.88 & 46.7\% & 51.1\% \\
\hline
\multirow{3}{*}{Poor} & SonoBase & 7.02 & 5.52 & 0.849 & +4.61 & 12.5\% & 12.5\% \\
& MedSAM2 & 13.94 & 9.79 & 0.657 & -11.53 & 31.2\% & 25.0\% \\
& SAM2 & 18.86 & 13.22 & 0.489 & -18.10 & 50.0\% & 50.0\% \\
\hline
\end{tabular}
\end{table}

\subsubsection{Gray Zone Analysis (EF 30--50\%)}

The gray zone (EF between 30--50\%) represents the most clinically important range where precise
quantification distinguishes normal from impaired systolic function.

\begin{table}[!ht]
\small
\centering
\caption{\textbf{Ejection fraction in gray zone (30--50\%, point and box prompts).}
Performance on $n=52$ subjects in the 30--50\% EF range. This range is clinically critical for systolic
dysfunction assessment.}
\label{tab:S8}
\begin{tabular}{|c|c|ccc|}
\hline
\textbf{Model} & \textbf{Prompt} & \textbf{MAE (\%)} & \textbf{Reclass @40\%} & \textbf{Reclass @35\%} \\
\hline
\multirow{2}{*}{SonoBase} & Point & \textbf{9.36} & \textbf{25.0}\% & \textbf{13.5}\% \\
& Box & \textbf{6.74} & \textbf{21.2}\% & \textbf{13.5}\% \\
\hline
\multirow{2}{*}{MedSAM2} & Point & 12.56 & 44.2\% & 36.5\% \\
& Box & 9.98 & 36.5\% & 21.2\% \\
\hline
\multirow{2}{*}{SAM2} & Point & 42.88 & 71.2\% & 86.5\% \\
& Box & 15.40 & 40.4\% & 44.2\% \\
\hline
\end{tabular}
\end{table}

SonoBase achieves the best gray-zone performance of the three models with box initialization
(MAE 6.74\%, 13.5\% reclassification at 35\%). SonoBase's gray-zone error exceeds its full-cohort error (6.74\% versus 6.63\%), while both baselines are marginally more accurate here than overall.
This range remains the hardest for clinical decision-making regarding therapy initiation.

\subsubsection{Cohen's Kappa for Classification Accuracy}

\begin{table}[!ht]
\small
\centering
\caption{\textbf{Cohen's kappa for EF classification agreement.} Unweighted Cohen's kappa between
model-derived and reference ejection fraction, computed separately as a binary classification at each
of two treatment thresholds: EF $\leq 40\%$ (HFrEF) and EF $\leq 35\%$ (ICD candidacy). The reference
is the ejection fraction distributed with CAMUS, derived from the expert-annotated reference contours;
CAMUS ships no categorical grading of systolic function, so no multi-grade classification is reported.
$n = 100$ patients.}
\label{tab:S9}
\begin{tabular}{|c|c|cc|}
\hline
\textbf{Model} & \textbf{Threshold} & \textbf{Point} & \textbf{Box} \\
\hline
\multirow{2}{*}{SonoBase} & $\kappa$ @40\% & 0.422 & 0.515 \\
& $\kappa$ @35\% & 0.428 & 0.457 \\
\hline
\multirow{2}{*}{MedSAM2} & $\kappa$ @40\% & 0.337 & 0.439 \\
& $\kappa$ @35\% & 0.428 & 0.558 \\
\hline
\multirow{2}{*}{SAM2} & $\kappa$ @40\% & 0.007 & 0.195 \\
& $\kappa$ @35\% & 0.005 & 0.128 \\
\hline
\end{tabular}
\end{table}

Uncorrected agreement is moderate rather than near-expert: SonoBase reaches $\kappa = 0.457$ at the 35\% threshold with box prompts, and MedSAM2 is nominally higher at 0.558 despite a 3.7 pp larger mean absolute error. Kappa penalizes SonoBase's systematic positive bias, which shifts predictions across the threshold in one direction, whereas MedSAM2's larger but more symmetric error preserves the marginal class balance. Because the bias is directional, it is correctable by calibration, unlike random error, which the following analysis quantifies.

\subsubsection{Bias-Corrected EF by Held-Out Linear Recalibration}

We fit a linear correction on one random half of the cohort and evaluated on the held-out half
(split-half; 2,000 random 50/50 splits), with leave-one-out (LOO) recalibration reported alongside as
a stability check. The two procedures agree closely, indicating the correction is not an artifact of
any particular split. In-sample recalibration would be circular and is not reported.

\begin{table}[!ht]
\renewcommand{\thetable}{S9b}
\small
\centering
\caption{\textbf{Bias-corrected EF classification agreement and error (box prompt, $n=100$).}
Split-half linear recalibration (fit on one random half, evaluated on the held-out half, 2,000
splits) with leave-one-out shown as a stability check. Calibration raises SonoBase's kappa and lowers
its MAE to well within the inter-observer band, while lowering MedSAM2's kappa at the 35\% threshold:
its higher uncorrected value reflected cancellation between a large negative bias and the class
balance rather than closer agreement. Confidence intervals are wide at $n=100$ and overlap; the
calibrated ordering should be read as directional.}
\label{tab:S9b}
\begin{tabular}{|l|c|cc|c|c|}
\hline
\textbf{Model} & \textbf{Threshold} & \textbf{Raw $\kappa$} & \textbf{Calibrated $\kappa$} & \textbf{95\% CI} & \textbf{LOO $\kappa$} \\
\hline
\textbf{SonoBase} & 35\% & 0.457 & \textbf{0.546} & $[0.267, 0.769]$ & 0.557 \\
\textbf{SonoBase} & 40\% & 0.515 & \textbf{0.639} & $[0.458, 0.810]$ & 0.645 \\
MedSAM2 & 35\% & 0.558 & 0.237 & $[-0.037, 0.598]$ & 0.229 \\
MedSAM2 & 40\% & 0.439 & 0.447 & $[0.150, 0.674]$ & 0.488 \\
SAM2 & 35\% & 0.128 & 0.067 & $[-0.070, 0.260]$ & 0.061 \\
SAM2 & 40\% & 0.195 & 0.217 & $[0.000, 0.452]$ & 0.272 \\
\hline
\end{tabular}
\end{table}

\addtocounter{table}{-1}
\begin{table}[!ht]
\renewcommand{\thetable}{S9c}
\small
\centering
\caption{\textbf{EF mean absolute error before and after leave-one-out calibration (box prompt).}
Calibration improves every model, most dramatically SAM2, which carried the largest systematic bias.}
\label{tab:S9c}
\begin{tabular}{|l|c|c|}
\hline
\textbf{Model} & \textbf{Raw MAE (\%)} & \textbf{Calibrated MAE (\%)} \\
\hline
\textbf{SonoBase} & 6.63 & \textbf{4.95} \\
MedSAM2 & 10.35 & 7.82 \\
SAM2 & 17.41 & 8.72 \\
\hline
\end{tabular}
\end{table}

\addtocounter{table}{-1}
Calibration works for SonoBase because its error is a genuine offset (bias $+5.53\%$, $r = 0.867$),
so a linear correction removes most of it. MedSAM2's correlation is only $r = 0.565$; removing its
$-7.39\%$ bias exposes the underlying scatter.

\begin{landscape}
\begin{table*}[!ht]
\renewcommand{\thetable}{S9d}
\scriptsize
\setlength{\tabcolsep}{4pt}
\centering
\caption{\textbf{Abdominal circumference error before and after held-out linear recalibration
(ACOUSLIC, $n=270$ videos).} Same protocol as Table~S9b: a linear map $\mathrm{GT}=a+b\cdot\mathrm{pred}$
fitted on training folds only and applied identically to every model; split-half = 2,000 random 50/50 splits
(mean MAE and the 2.5--97.5th percentile of split MAEs, a spread across splits); LOO = leave-one-out.
Few-shot rows are mean $\pm$ SD over three seeds, calibrated within each seed. Secondary analysis: test
labels enter the fit. Under box prompts the uncorrected MedSAM2 advantage is a scale offset that calibration
removes (paired Wilcoxon on LOO errors $p=0.998$); under point prompts SonoBase remains best
($p=2.5\times10^{-6}$); in the few-shot arm SonoBase and fine-tuned SAM2 tie ($p=0.86$) and MedSAM2 remains
behind ($p=2.1\times10^{-18}$). $^{\dagger}$SAM2 point prompt is a $\sim$2$\times$ collapse (bias $+294$~mm); its
calibrated value is a rescaled collapse, not a result.}
\label{tab:S9d}
\begin{tabular}{|l|l|c|c|c|c|}
\hline
\textbf{Arm} & \textbf{Model} & \textbf{Raw MAE (mm)} & \textbf{Bias (mm)} & \textbf{Split-half MAE [2.5, 97.5]} & \textbf{LOO MAE} \\
\hline
Zero-shot, box & SonoBase & 18.92 & $-17.75$ & 13.94 [12.62, 15.23] & 13.89 \\
 & MedSAM2 & \textbf{14.77} & $-7.54$ & 13.97 [12.52, 15.43] & 13.94 \\
 & SAM2 & 18.11 & $-14.98$ & 13.84 [12.54, 15.14] & \textbf{13.77} \\
\hline
Zero-shot, point & SonoBase & \textbf{30.50} & $-20.94$ & 31.33 [28.86, 34.28] & \textbf{31.36} \\
 & MedSAM2 & 40.34 & $-11.35$ & 35.02 [32.24, 37.72] & 34.92 \\
 & SAM2$^{\dagger}$ & 296.34 & $+294.53$ & 37.04 [34.22, 39.90] & 36.94 \\
\hline
Few-shot $N{=}5$, box & SonoBase & 17.17 $\pm$ 5.52 & $-14.7$ & 12.46 $\pm$ 1.00 & \textbf{12.39 $\pm$ 0.98} \\
 & MedSAM2 & 29.80 $\pm$ 16.83 & $+10.2$ & 21.30 $\pm$ 6.38 & 21.23 $\pm$ 6.35 \\
 & SAM2 (fine-tuned) & \textbf{12.79 $\pm$ 0.74} & $-2.8$ & 12.55 $\pm$ 0.37 & 12.50 $\pm$ 0.38 \\
\hline
\end{tabular}
\end{table*}
\end{landscape}
\addtocounter{table}{-1}

\subsection{Head Circumference and Gestational Age Estimation (HC18)}
\label{subsec:hc-results}

Accurate fetal head circumference (HC) measurement enables precise gestational age (GA) estimation,
critical for perinatal management.

\subsubsection{HC Results (Point Prompt)}

\begin{table}[!ht]
\small
\centering
\caption{\textbf{Head circumference results (point prompt).} Accuracy measured in absolute
error (MAE in mm) and percentage within clinical thresholds ($<$3mm, $<$5mm). Pearson correlation with
ground truth measurements. $n=201$ images.}
\label{tab:S10}
\begin{tabular}{|c|cccc|cc|}
\hline
\textbf{Model} & \textbf{MAE (mm)} & \textbf{Std} & \textbf{r} & \textbf{$<$3mm} & \textbf{$<$5mm} & \textbf{n} \\
\hline
SonoBase & \textbf{2.49} & 2.92 & \textbf{0.998} & \textbf{73.1\%} & \textbf{87.1\%} & 201 \\
MedSAM2 & 66.62 & 69.11 & 0.615 & 15.4\% & 20.9\% & 201 \\
SAM2 (no FT) & 77.68 & 77.66 & 0.556 & 13.4\% & 17.4\% & 201 \\
\textit{GT ellipse fit (floor)} & \textit{1.37} & --- & --- & --- & --- & 201 \\
\hline
\end{tabular}
\end{table}

SonoBase achieves remarkable accuracy with MAE 2.49~mm (better than typical ultrasound caliper measurement
variability) and excellent correlation (r=0.998). Foundation models without ultrasound pretraining fail
catastrophically on this task.

\subsubsection{HC Results (Box Prompt)}

\begin{table}[!ht]
\small
\centering
\caption{\textbf{Head circumference results (box prompt).} Box initialization provides
additional structural guidance, slightly improving SonoBase accuracy and substantially improving baseline models.
Specialist rows (unprompted, trained on the HC18 training split, scored on the same 201 images; Methods):
both are significantly less accurate than box-prompted SonoBase (paired Wilcoxon on absolute errors,
$q = 2.5\times10^{-4}$ for DeepLabV3+ and $7.4\times10^{-5}$ for nnU-Net) despite comparable overlap
(Dice 0.976 and 0.977 versus 0.978). Medians are 1.36 (SonoBase), 1.80 (DeepLabV3+) and 1.78~mm (nnU-Net);
nnU-Net's mean and standard deviation reflect one gross localization failure (image 29: predicted area six
times the skull, 108~mm error; 2.31~mm without it), and errors above 10~mm number 1, 3 and 4 of 201 for
SonoBase, DeepLabV3+ and nnU-Net.}
\label{tab:S11}
\begin{tabular}{|l|cccc|cc|}
\hline
\textbf{Model} & \textbf{MAE (mm)} & \textbf{Std} & \textbf{r} & \textbf{$<$3mm} & \textbf{$<$5mm} & \textbf{n} \\
\hline
SonoBase & \textbf{1.81} & 1.63 & \textbf{0.9994} & \textbf{80.6\%} & \textbf{96.0\%} & 201 \\
MedSAM2 & 3.14 & 4.17 & 0.9972 & 61.7\% & 85.6\% & 201 \\
SAM2 (no FT) & 5.60 & 7.60 & 0.9914 & 45.8\% & 70.1\% & 201 \\
\hline
\multicolumn{7}{|l|}{\textit{Specialists, unprompted, HC18-trained}} \\
\hline
DeepLabV3+ (R50) & 2.30 & 2.29 & 0.9989 & 72.6\% & 90.5\% & 201 \\
nnU-Net ResEnc-M & 2.83 & 7.74 & 0.9928 & 71.6\% & 90.5\% & 201 \\
\hline
\textit{GT ellipse fit (floor)} & \textit{1.37} & --- & --- & --- & --- & 201 \\
\hline
\end{tabular}
\end{table}

With box prompts, all models show improved performance. SonoBase maintains the lowest MAE (1.81~mm)
with 96.0\% of measurements within the 5mm clinical tolerance, and is more accurate than both HC18-trained
specialists on the same images.

\subsubsection{Gestational Age Estimation}

Gestational age is computed from HC using the standard Hadlock equation. Estimation accuracy directly
reflects HC measurement precision.

\begin{table}[!ht]
\small
\centering
\caption{\textbf{Gestational age estimation from head circumference.} Point and box prompts.
MAE in days, percentage within clinical tolerance ($\leq$3 days, $\leq$7 days), and percentage $>$14 days.}
\label{tab:S12}
\begin{tabular}{|c|c|cc|ccc|}
\hline
\textbf{Model} & \textbf{Prompt} & \textbf{MAE (days)} & \textbf{Std} &
\textbf{$\leq$3 days} & \textbf{$\leq$7 days} & \textbf{$>$14 days} \\
\hline
\multirow{2}{*}{SonoBase} & Point & \textbf{1.65} & 2.03 & \textbf{85.1\%} & \textbf{95.5\%} & \textbf{0.0\%} \\
& Box & \textbf{1.20} & 1.39 & \textbf{93.0\%} & \textbf{99.0\%} & \textbf{0.0\%} \\
\hline
\multirow{2}{*}{MedSAM2} & Point & 33.37 & 35.24 & 21.4\% & 32.8\% & 52.2\% \\
& Box & 1.99 & 2.68 & 83.1\% & 94.5\% & 0.5\% \\
\hline
\multirow{2}{*}{SAM2} & Point & 62.76 & 83.56 & 18.9\% & 31.8\% & 59.2\% \\
& Box & 3.97 & 7.10 & 68.7\% & 88.6\% & 6.5\% \\
\hline
\end{tabular}
\end{table}

SonoBase achieves exceptional GA estimation accuracy: MAE 1.20 days (box prompt) with 93.0\% within
$\pm 3$ days, exceeding the clinically recommended precision for fetal assessment.

\subsection{Abdominal Circumference (ACOUSLIC)}
\label{subsec:ac-results}

Fetal abdominal circumference measurement is essential for growth assessment and diagnosis of structural
abnormalities.

\subsubsection{AC Results (Point Prompt)}

\begin{table}[!ht]
\small
\centering
\caption{\textbf{Abdominal circumference results (point prompt).} MAE in mm, Pearson correlation,
and comparison to baseline. $n=270$ videos (per-video AC is the mean over annotated frames). Ground truth is the
ACOUSLIC v1.1 reference at the header pixel spacing (0.28~mm/px); the ellipse-fit pipeline has a measurement floor
of 7.39~mm on the ground-truth masks when all annotated frames are averaged.}
\label{tab:S13}
\begin{tabular}{|c|cc|cc|}
\hline
\textbf{Model} & \textbf{MAE (mm)} & \textbf{Std} & \textbf{r} & \textbf{n} \\
\hline
SonoBase & \textbf{30.50} & 43.50 & \textbf{0.460} & 270 \\
MedSAM2 & 40.34 & 37.33 & 0.323 & 270 \\
SAM2 (no FT) & 296.34 & 64.23 & 0.226 & 270 \\
\hline
\end{tabular}
\end{table}

Point initialization shows challenging performance on AC due to the ambiguity of internal abdominal
boundaries in ultrasound images. SonoBase shows moderate correlation (r=0.460) with MAE 30.50~mm, significantly below MedSAM2 (Wilcoxon $p = 2.6\times10^{-6}$).

\subsubsection{AC Results (Box Prompt)}

\begin{table}[!ht]
\small
\centering
\caption{\textbf{Abdominal circumference results (box prompt).} Box initialization substantially
improves AC measurement by providing clear anatomical guidance for peritoneal boundary detection.
MedSAM2 records the lowest raw error; the difference versus SonoBase is significant (Wilcoxon $p = 7.8\times10^{-6}$, BH $q = 1.6\times10^{-5}$) and is a scale offset removed by held-out recalibration (Table~S9d).
Specialist rows: the three winning ACOUSLIC-AI solutions retrained under subject-level five-fold
cross-validation on the 300 public sweeps (in-distribution; Methods), scored with the same frame-averaged
protocol on the same videos; $p$ is the paired Wilcoxon against SonoBase box (BH $q$ over the specialist
family: $8.0\times10^{-16}$, $0.026$ and $0.33$). A2's segmentation head leaves 17\% of the annotated
frames empty, so its per-video mean uses 18.0 instead of 21.7 frames and 264 of 270 videos. Raw scale
throughout; no recalibration was applied to the specialist rows. Details in Table~S14b.}
\label{tab:S14}
\begin{tabular}{|l|cc|cc|}
\hline
\textbf{Model} & \textbf{MAE (mm)} & \textbf{Std} & \textbf{r} & \textbf{Wilcoxon p} \\
\hline
SonoBase & 18.92 & 17.02 & 0.913 & \multirow{2}{*}{$7.8\times10^{-6}$} \\
MedSAM2 & \textbf{14.77} & 13.87 & 0.909 & \\
SAM2 (no FT) & 18.11 & 14.93 & 0.918 & -- \\
\hline
\multicolumn{5}{|l|}{\textit{Challenge-winning specialists, retrained in-distribution (five-fold CV), unprompted}} \\
\hline
A3 (nnU-Net hybrid) & \textbf{10.79} & 8.88 & 0.966 & $9.4\times10^{-17}$ \\
A2 (ConvNeXt-V2 U-Net; $n=264$) & 18.20 & 19.41 & 0.888 & $0.020$ \\
A1 (U-Net ensemble) & 20.04 & 22.30 & 0.854 & $0.27$ \\
\hline
\end{tabular}
\end{table}

\begin{landscape}
\begin{table*}[!ht]
\renewcommand{\thetable}{S14b}
\scriptsize
\setlength{\tabcolsep}{3pt}
\centering
\caption{\textbf{ACOUSLIC: in-distribution challenge specialists versus zero-shot SonoBase under two protocols.}
The three winning solutions of the ACOUSLIC-AI challenge (A1 Ninalga, nine-member VGG U-Net ensemble; A2 Kondo
\emph{et al.}, ConvNeXt-V2 U-Net with three-slice input; A3 Akumu \emph{et al.}, ResNet-50 frame selection plus
2D nnU-Net) were retrained with their released code and default hyperparameters under GroupKFold(5) by subject
over all 300 public sweeps (240 training sweeps per fold; every one of our 270 test videos in exactly one test
fold), because their released weights were trained on the videos we test on. \emph{Annotated-frame protocol}:
each segmentation network applied to every expert-annotated frame and scored on the identical 5,868 rows /
270 videos as the archived SonoBase run; mIoU and Dice over rows, $\Delta$ mIoU is specialist minus SonoBase
box (exact prompt) with bootstrap 95\% CI per row and per video; per-video AC is the mean of the ellipse fits
over the annotated frames (corrected basis: v1.1 truth, 0.28~mm/px; ground-truth floor 7.39~mm). \emph{Native
protocol}: each solution's own frame selection on the full 840-frame stack, scored by the organizers'
evaluation code (score $=0.5(1-\mathrm{NAE})+0.25\,\mathrm{Dice}_{\mathrm{soft}}+0.25\,\mathrm{WFSS}$; the
challenge report's hidden-test values for the same solutions are 0.71, 0.65 and 0.64 for A1, A2 and A3).
The two protocols disagree for a stated reason: averaging AC over every annotated frame penalizes any model
that follows the visible boundary on oblique sections, whereas the native metric lets the model choose its
frame. A3 used one frame-selection checkpoint per fold rather than the authors' checkpoint ensemble. SonoBase
never saw a Sierra Leone sweep; zero-shot rows are given for reference. $^{a}$264 of 270 videos (six videos with every
annotated frame empty); 18.0 frames per video instead of 21.7. In the native-protocol runs A2 selected no frame in 8 of 270
videos, A1 and A3 in none.}
\label{tab:S14b}
\resizebox{\linewidth}{!}{%
\begin{tabular}{|l|cc|c|c|c|c|ccc|}
\hline
 & \multicolumn{6}{c|}{\textbf{Annotated-frame protocol (identical rows)}} & \multicolumn{3}{c|}{\textbf{Native protocol}} \\
\textbf{Model} & \textbf{mIoU} & \textbf{Dice} & \textbf{Empty rows} & \textbf{$\Delta$ mIoU per row / per video} & \textbf{AC MAE (mm) [95\% CI]} & \textbf{Bias (mm)} & \textbf{Score} & \textbf{NAE mean / median} & \textbf{WFSS} \\
\hline
A3 (5-fold CV) & \textbf{89.4} & \textbf{94.3} & 0 & $+11.8$ [$11.1$, $12.5$] / $+9.1$ [$7.3$, $11.0$] & \textbf{10.79} [9.77, 11.88] & $-7.2$ & 0.843 & 0.114 / 0.029 & 0.754 \\
A1 (5-fold CV) & 82.4 & 87.7 & 200 (3.4\%) & $+4.8$ [$4.0$, $5.7$] / $+2.3$ [$-0.1$, $4.7$] & 20.04 [17.55, 22.85] & $-18.7$ & \textbf{0.892} & 0.088 / 0.032 & 0.853 \\
A2 (5-fold CV) & 69.9 & 74.6 & 1,005 (17\%) & $-7.7$ [$-8.7$, $-6.6$] / $-9.7$ [$-12.9$, $-6.6$] & 18.20 [15.99, 20.67]$^{a}$ & $-16.7$ & 0.813 & 0.126 / 0.038 & 0.683 \\
\hline
SonoBase box, zero-shot & 77.6 & 83.7 & 0 & ref. & 18.92 [16.99, 21.01] & $-17.8$ & --- & --- & --- \\
SonoBase point, zero-shot & 72.9 & 80.0 & 0 & --- & 30.50 [25.54, 36.03] & $-20.9$ & --- & --- & --- \\
\hline
\end{tabular}}

\end{table*}
\end{landscape}
\addtocounter{table}{-1}

With box prompts all three models reach $r > 0.90$. MedSAM2's raw error is lowest (14.77~mm) and significantly
below SonoBase's (18.92~mm; $p = 7.8\times10^{-6}$). The gap is a scale offset rather than contour accuracy:
SonoBase's masks have the highest per-frame overlap and the lowest per-frame circumference error, but it is about 3.5\% inside the expert boundary, and averaging the $\sim$22 annotated frames per video cancels MedSAM2's random
error while leaving SonoBase's systematic under-measurement intact. After identical held-out linear recalibration
of all three models the difference disappears (13.9, 13.9, and 13.8~mm; $p = 0.998$; Table~S9d), and under the
challenge organizers' optimal-plane protocol the three models tie (13.1, 12.8, and 12.9~mm).

\subsection{Per-Structure CAMUS Segmentation Accuracy}
\label{subsec:per-structure}

The CAMUS dataset includes three cardiac structures: left ventricular endocardium (LV Endo), left
ventricular epicardium (LV Epi), and left atrium (LA). Per-structure analysis reveals differential
model strengths.

\begin{table}[!ht]
\small
\setlength{\tabcolsep}{3.5pt}
\centering
\caption{\textbf{Per-structure cardiac segmentation on CAMUS.} Dice coefficient and mIoU for
three structures and overall average. Point and box prompts. All metrics in percentages.}
\label{tab:S15}
\begin{tabular}{|c|c|cc|cc|cc|cc|}
\hline
\textbf{Model} & \textbf{Prompt} & \multicolumn{2}{c|}{\textbf{LV Endo}} &
\multicolumn{2}{c|}{\textbf{LV Epi}} & \multicolumn{2}{c|}{\textbf{LA}} &
\multicolumn{2}{c|}{\textbf{Overall}} \\
& & Dice & mIoU & Dice & mIoU & Dice & mIoU & Dice & mIoU \\
\hline
\multirow{2}{*}{SonoBase} & Point & 93.87 & 88.60 & 87.41 & 77.90 & 91.12 & 84.21 & 90.80 & 83.57 \\
& Box & 94.87 & 90.35 & 88.26 & 79.18 & 93.49 & 88.00 & 92.21 & 85.84 \\
\hline
\multirow{2}{*}{MedSAM2} & Point & 87.87 & 78.98 & 79.29 & 66.65 & 86.16 & 77.31 & 84.44 & 74.31 \\
& Box & 91.13 & 83.91 & 81.75 & 69.58 & 91.34 & 84.33 & 88.08 & 79.27 \\
\hline
\multirow{2}{*}{SAM2} & Point & 29.15 & 17.51 & 30.18 & 18.08 & 18.88 & 10.91 & 26.07 & 15.50 \\
& Box & 85.48 & 75.03 & 41.74 & 26.99 & 81.35 & 71.12 & 69.52 & 57.72 \\
\hline
\end{tabular}
\end{table}

SonoBase achieves exceptional per-structure accuracy with LV endocardium Dice $>$94\% and overall
mIoU 85.84\% with box prompts. The LV endocardium (the most clinically important boundary for EF
calculation) is the most accurately segmented structure for SonoBase under both prompts, and for most
model-prompt combinations overall.

\subsection{Prostate Volume (RegPro Dataset, n=8 subjects)}
\label{subsec:prostate}

Prostate volume measurement is critical for diagnosis of benign prostatic hyperplasia and risk
stratification. Accurate segmentation enables reliable volume computation via ellipsoid or summation formulas.

\subsubsection{Prostate Volume (Point Prompt)}

\begin{table}[!ht]
\small
\centering
\caption{\textbf{Prostate volume results (point prompt).} Mean absolute error (mL), relative
error (\%), and Pearson correlation against expert volumes on the RegPro cohort ($n=8$).
Bootstrap 95\% CIs in Table S18.}
\label{tab:S16}
\begin{tabular}{|c|cc|c|c|}
\hline
\textbf{Model} & \textbf{MAE (mL)} & \textbf{Rel Err (\%)} & \textbf{r} & \textbf{n} \\
\hline
SonoBase & \textbf{6.28} & \textbf{15.95\%} & \textbf{0.960} & 8 \\
MedSAM2 & 42.07 & 101.80\% & -0.236 & 8 \\
SAM2 (no FT) & 145.59 & 367.62\% & 0.138 & 8 \\
\hline
\end{tabular}
\end{table}

SonoBase achieves the most accurate volume estimation of the three models (r=0.960, MAE 6.28~mL).
This precision is clinically actionable for prostate disease management.

\subsubsection{Prostate Volume (Box Prompt)}

\begin{table}[!ht]
\small
\centering
\caption{\textbf{Prostate volume results (box prompt).} Box prompts provide structural guidance
that further improves volume accuracy. SonoBase shows minimal error with excellent correlation.}
\label{tab:S17}
\begin{tabular}{|c|cc|c|c|}
\hline
\textbf{Model} & \textbf{MAE (mL)} & \textbf{Rel Err (\%)} & \textbf{r} & \textbf{n} \\
\hline
SonoBase & \textbf{3.62} & \textbf{9.27\%} & \textbf{0.962} & 8 \\
MedSAM2 & 41.35 & 99.77\% & -0.087 & 8 \\
SAM2 (no FT) & 41.03 & 98.95\% & -0.115 & 8 \\
\hline
\end{tabular}
\end{table}

Box prompts further reduce SonoBase's volume error to 3.62~mL (9.27\% relative), within the
inter-observer variability of 11.4\%, demonstrating clinical utility for prostate disease
management decisions.

\subsubsection{Bootstrap Confidence Intervals}

\begin{table}[!ht]
\small
\centering
\caption{\textbf{Bootstrap 95\% confidence intervals for prostate volume MAE ($n=8$).} Bootstrap
resampling with 10,000 iterations. The SonoBase and baseline intervals do not overlap, separating
the models by an order of magnitude.}
\label{tab:S18}
\begin{tabular}{|c|c|c|}
\hline
\textbf{Model} & \textbf{MAE (mL)} & \textbf{Bootstrap 95\% CI (mL)} \\
\hline
SonoBase (Point) & 6.28 & [4.61, 8.09] \\
SonoBase (Box) & 3.62 & [2.28, 5.00] \\
MedSAM2 (Point) & 42.07 & [36.17, 48.15] \\
MedSAM2 (Box) & 41.35 & [35.35, 47.77] \\
SAM2 (Point) & 145.59 & [110.48, 181.14] \\
SAM2 (Box) & 41.03 & [34.99, 47.50] \\
\hline
\end{tabular}
\end{table}

Bootstrap CIs confirm that SonoBase estimates are substantially more precise than baseline models,
with tight confidence bounds.

\newpage

\section{Statistical Testing}
\label{sec:statistical-testing}

Comprehensive statistical testing validates significant differences between SonoBase and baseline models
using non-parametric methods appropriate for skewed error distributions.

\subsection{Wilcoxon Signed-Rank Tests (Point Prompt)}

\begin{table}[!ht]
\small
\setlength{\tabcolsep}{3pt}
\centering
\caption{\textbf{Wilcoxon signed-rank tests comparing SonoBase to baselines (point prompt).}
Eight paired comparisons across four datasets × two baseline models. Benjamini-Hochberg FDR
correction applied within the point-prompt family of eight tests at $q = 0.05$. All tests remain
significant after correction.}
\label{tab:S19}
\begin{tabular}{|c|c|c|ccc|}
\hline
\textbf{Dataset} & \textbf{Comparison} & \textbf{n} & \textbf{p-value} &
\textbf{Corrected} & \textbf{Significant} \\
\hline
CAMUS & SonoBase vs MedSAM2 & 100 & $0.0212$ & $0.0212$ & Yes \\
HC18 & SonoBase vs MedSAM2 & 201 & $9.15\times10^{-33}$ & $2.44\times10^{-32}$ & Yes \\
ACOUSLIC & SonoBase vs MedSAM2 & 270 & $2.60\times10^{-6}$ & $4.16\times10^{-6}$ & Yes \\
RegPro & SonoBase vs MedSAM2 & 8 & $0.00781$ & $0.00893$ & Yes \\
CAMUS & SonoBase vs SAM2 & 100 & $4.67\times10^{-18}$ & $9.34\times10^{-18}$ & Yes \\
HC18 & SonoBase vs SAM2 & 201 & $6.45\times10^{-33}$ & $2.44\times10^{-32}$ & Yes \\
ACOUSLIC & SonoBase vs SAM2 & 270 & $7.69\times10^{-46}$ & $6.15\times10^{-45}$ & Yes \\
RegPro & SonoBase vs SAM2 & 8 & $0.00781$ & $0.00893$ & Yes \\
\hline
\end{tabular}
\end{table}

\subsection{Wilcoxon Signed-Rank Tests (Box Prompt)}

\begin{table}[!ht]
\small
\setlength{\tabcolsep}{3pt}
\centering
\caption{\textbf{Wilcoxon signed-rank tests comparing SonoBase to baselines (box prompt).}
Eight paired comparisons. Benjamini-Hochberg FDR correction applied within the box-prompt family of
eight tests at $q = 0.05$. SonoBase has the lower error in every significant row except ACOUSLIC vs MedSAM2,
which is significant in MedSAM2's favor (corrected $p = 1.6\times10^{-5}$), a scale offset that held-out
recalibration removes (Table~S9d); ACOUSLIC vs SAM2 is non-significant (corrected $p = 0.977$).}
\label{tab:S20}
\begin{tabular}{|c|c|c|ccc|}
\hline
\textbf{Dataset} & \textbf{Comparison} & \textbf{n} & \textbf{p-value} &
\textbf{Corrected} & \textbf{Significant} \\
\hline
CAMUS & SonoBase vs MedSAM2 & 100 & $0.0127$ & $0.0145$ & Yes \\
HC18 & SonoBase vs MedSAM2 & 201 & $3.06\times10^{-7}$ & $8.15\times10^{-7}$ & Yes \\
ACOUSLIC & SonoBase vs MedSAM2 & 270 & $7.80\times10^{-6}$ & $1.56\times10^{-5}$ & Yes (MedSAM2 lower) \\
RegPro & SonoBase vs MedSAM2 & 8 & $0.00781$ & $0.0104$ & Yes \\
CAMUS & SonoBase vs SAM2 & 100 & $9.74\times10^{-9}$ & $3.90\times10^{-8}$ & Yes \\
HC18 & SonoBase vs SAM2 & 201 & $1.04\times10^{-23}$ & $8.29\times10^{-23}$ & Yes \\
ACOUSLIC & SonoBase vs SAM2 & 270 & $0.977$ & $0.977$ & \textbf{No} \\
RegPro & SonoBase vs SAM2 & 8 & $0.00781$ & $0.0104$ & Yes \\
\hline
\end{tabular}
\end{table}

\subsection{Bootstrap Confidence Intervals for All Clinical Measurements}

\begin{table}[!ht]
\small
\centering
\caption{\textbf{Bootstrap 95\% confidence intervals for all clinical MAE measurements.}
Point and box prompts across all datasets and models. Bootstrap resampling with 10,000 iterations.}
\label{tab:S21}
\begin{tabular}{|c|c|c|c|}
\hline
\textbf{Dataset} & \textbf{Model/Prompt} & \textbf{MAE} & \textbf{Bootstrap 95\% CI} \\
\hline
\multirow{6}{*}{CAMUS EF} & SonoBase (Point) & 8.86 & [7.78, 9.99] \\
& SonoBase (Box) & 6.63 & [5.68, 7.62] \\
& MedSAM2 (Point) & 12.80 & [10.45, 15.42] \\
& MedSAM2 (Box) & 10.35 & [8.38, 12.61] \\
& SAM2 (Point) & 45.01 & [42.41, 47.53] \\
& SAM2 (Box) & 17.41 & [14.45, 20.52] \\
\hline
\multirow{6}{*}{HC18 HC} & SonoBase (Point) & 2.49 & [2.12, 2.93] \\
& SonoBase (Box) & 1.81 & [1.60, 2.04] \\
& MedSAM2 (Point) & 66.62 & [57.36, 76.37] \\
& MedSAM2 (Box) & 3.14 & [2.62, 3.78] \\
& SAM2 (Point) & 77.68 & [67.24, 88.45] \\
& SAM2 (Box) & 5.60 & [4.61, 6.69] \\
\hline
\multirow{6}{*}{ACOUSLIC AC} & SonoBase (Point) & 30.50 & [25.54, 36.03] \\
& SonoBase (Box) & 18.92 & [16.99, 21.01] \\
& MedSAM2 (Point) & 40.34 & [35.96, 44.78] \\
& MedSAM2 (Box) & 14.77 & [13.19, 16.50] \\
& SAM2 (Point) & 296.34 & [288.40, 303.70] \\
& SAM2 (Box) & 18.11 & [16.38, 19.95] \\
\hline
\multirow{6}{*}{RegPro Vol} & SonoBase (Point) & 6.28 & [4.61, 8.09] \\
& SonoBase (Box) & 3.62 & [2.28, 5.00] \\
& MedSAM2 (Point) & 42.07 & [36.17, 48.15] \\
& MedSAM2 (Box) & 41.35 & [35.35, 47.77] \\
& SAM2 (Point) & 145.59 & [110.48, 181.14] \\
& SAM2 (Box) & 41.03 & [34.99, 47.50] \\
\hline
\multicolumn{4}{|l|}{\textit{Specialists, unprompted (Tables~S5, S11, S14)}} \\
\hline
\multirow{3}{*}{CAMUS EF} & nnU-Net ResEnc-M, CAMUS-trained & 6.28 & [5.44, 7.14] \\
& EchoNet-Dynamic, CAMUS fine-tuned & 7.43 & [6.30, 8.65] \\
& EchoNet-Dynamic, zero-shot ($n=99$) & 16.92 & [13.36, 21.01] \\
\hline
\multirow{2}{*}{HC18 HC} & DeepLabV3+ (ResNet-50), HC18-trained & 2.30 & [1.99, 2.62] \\
& nnU-Net ResEnc-M, HC18-trained & 2.83 & [2.08, 4.04] \\
\hline
\multirow{3}{*}{ACOUSLIC AC} & A3, five-fold CV & 10.79 & [9.77, 11.88] \\
& A2, five-fold CV ($n=264$) & 18.20 & [15.99, 20.67] \\
& A1, five-fold CV & 20.04 & [17.55, 22.85] \\
\hline
\end{tabular}
\end{table}

\subsection{Multiple Hypothesis Correction}

The Benjamini-Hochberg FDR procedure is applied at $q=0.05$ separately within each prompt family:
the eight point-prompt tests in Table S19 and the eight box-prompt tests in Table S20. All comparisons
marked ``Yes'' in the Significant column remain significant after BH correction. The one non-significant
comparison is ACOUSLIC box-prompt SonoBase vs SAM2 (corrected $p=0.977$). The ACOUSLIC box-prompt SonoBase vs
MedSAM2 comparison is significant in MedSAM2's favor (corrected $p=1.6\times10^{-5}$); it reflects a systematic
under-measurement by SonoBase's box-prompted contours rather than lower contour accuracy, and disappears under
held-out recalibration (Table~S9d). The ellipse-fitting measurement floor on this dataset is 7.39~mm.

\newpage

\section{Catastrophic Failure Analysis}
\label{sec:catastrophic-failure}

Despite strong overall performance, segmentation models can occasionally produce grossly incorrect
predictions. We define a \emph{catastrophic failure resolved} as a test case on which at least one
baseline scores IoU $< 10$ (no usable segmentation) while SonoBase scores IoU $> 50$ (clinically
useful), under point prompting with no corrective clicks. Cases are counted per patient, image or
volume; for video and 3D datasets a case's IoU is the mean over its frames or slices, so that a
long sequence does not outweigh a single image.

\subsection{Failure Rates Across Datasets}

\begin{table}[!ht]
\small
\centering
\caption{\textbf{Catastrophic-failure resolution per dataset.} A resolved catastrophic failure is a
test case on which at least one baseline scores IoU $< 10$ while SonoBase scores IoU $> 50$, under
point prompting with no corrections. Counted per case over all fifteen evaluation datasets: for video
and 3D datasets a case's IoU is the mean over its frames or slices, so denominators match the test-set
sizes reported in Tables S2--S3. Of the 1{,}324 cases where at least one baseline collapsed, SonoBase
recovered a usable segmentation in 1{,}073, a conditional rescue rate of 81.0\%.}
\label{tab:S22}
\begin{tabularx}{\textwidth}{|X|cc|c|}
\hline
\textbf{Dataset} & \textbf{Resolved} & \textbf{Images} & \textbf{Rate} \\
\hline
RegPro (3D transrectal prostate)      &     7 &      8 & \textbf{87.5\%} \\
PFUS (pelvic floor, video)             &    18 &     22 & 81.8\% \\
Brachial-Plexus (nerve, video)        &    30 &     44 & 68.2\% \\
C-TRUS (colon wall, transabdominal)   &    55 &     97 & 56.7\% \\
HC18 (fetal head)                     &    82 &    201 & 40.8\% \\
DDTI (thyroid nodule)                 &   229 &    567 & 40.4\% \\
MMOTU-3d (ovarian tumor)              &    45 &    153 & 29.4\% \\
TG3K (thyroid gland)                  &   164 &    718 & 22.8\% \\
BUS-BRA (breast lesion)               &   302 &  1{,}688 & 17.9\% \\
KidneyUS (multi-vendor kidney)        &    46 &    422 & 10.9\% \\
FUGC (cervix)                         &    35 &    396 & 8.8\% \\
LUMINOUS (lumbar muscle)              &    23 &    307 & 7.5\% \\
BUSI (breast lesion)                  &     9 &    130 & 6.9\% \\
CAMUS (cardiac video)                 &    13 &    200 & 6.5\% \\
ACOUSLIC (fetal abdomen, POCUS)       &    15 &    270 & 5.6\% \\
\hline
\multicolumn{1}{|l|}{\textbf{Pooled (15 datasets)}} & \textbf{1{,}073} & \textbf{5{,}223} & \textbf{20.5\%} \\
\hline
\end{tabularx}
\end{table}

Resolved failures concentrate in the datasets where the baselines have no usable representation at
all: RegPro (87.5\%, 3D transrectal prostate), PFUS (81.8\%, pelvic-floor video), Brachial-Plexus
(68.2\%, difficult acoustic access) and C-TRUS (56.7\%, variable probe angles). Conversely, on
well-standardized acquisitions where the baselines are already competent in ACOUSLIC (5.6\%), CAMUS
(6.5\%), BUSI (6.9\%), there is little left to resolve. The rate therefore measures the percentage of instances where a baseline collapses and SonoBase does not, and is bounded above by the baseline collapse rate itself; the conditional rescue rate of 81.0\% is the quantity that is invariant to how often the baselines fail. Of the 1,324 failures, 1,214 (91.7\%) involve a single baseline and 110 (2.1\% of all cases) both baselines together; SonoBase resolves 90 of the 110 joint failures (81.8\%), the same rate as for
single-baseline failures, so the rescue rate is not inflated by cases where only the weaker baseline collapsed.

\subsection{Representative Failure Examples}

\begin{figure}[!ht]
\centering
\includegraphics[width=\textwidth]{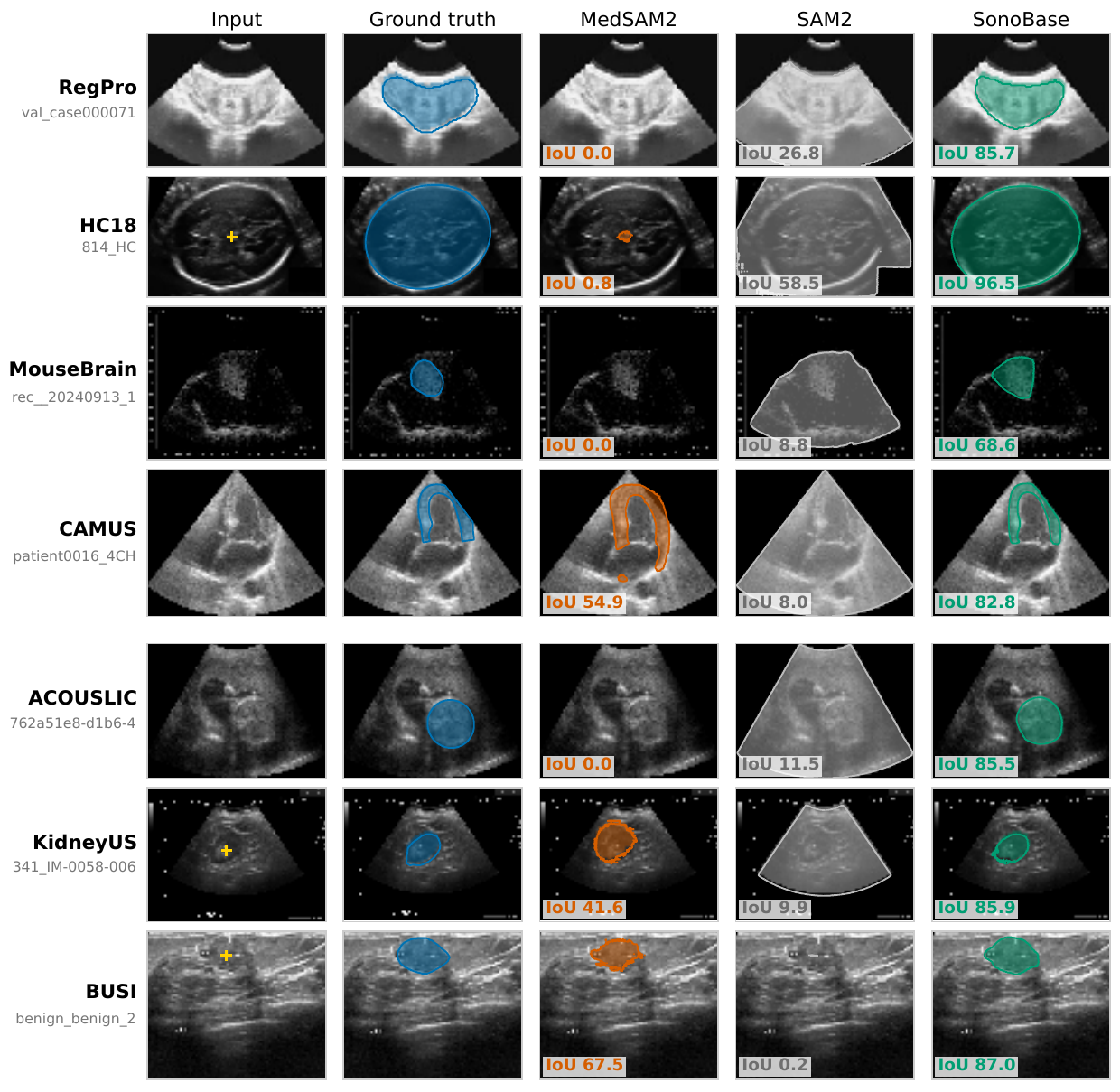}
\caption{\textbf{Representative catastrophic failure examples.} One case per
dataset: an image on which at least one
baseline produces no usable segmentation (IoU $<10$) while SonoBase produces a clinically useful one
(IoU $>50$), under point prompting with no corrections. Among qualifying images we show the one whose
SonoBase IoU is the median, so each row is a typical resolved case rather than the most favorable
one available. Columns are the input frame, ground truth (blue) and each model's prediction, with the
per-image IoU of the displayed frame; the yellow cross marks the point prompt where the displayed
frame is the prompted frame. Failures involve low tissue contrast, acoustic shadowing, unusual
anatomy and severe domain shift. Which baseline collapses varies by row, MedSAM2 alone on RegPro, HC18, and ACOUSLIC, SAM2 alone on CAMUS, KidneyUS and BUSI, and both only on MouseBrainTumor and this is consistent with the general pattern that most failures are single-baseline rather than joint.}
\label{fig:S3}
\end{figure}

\newpage

\section{Click Efficiency and Convergence Analysis}
\label{sec:click-efficiency}

Interactive segmentation evaluation quantifies the number of user clicks required to achieve target
accuracy (e.g., 80\% mIoU). This metric directly translates to clinical workflow efficiency.

\subsection{Full Convergence Results}

\begin{table}[!ht]
\scriptsize
\setlength{\tabcolsep}{3pt}
\centering
\caption{\textbf{Click convergence for all models under the Tables~S2--S3 prompt protocol.} Macro
mIoU at 0, 1, 3, 5, and 7 oracle corrections (10\% jittered box or uniformly sampled point; mean of
three seeds; seed-to-seed s.d.\ $\leq 0.5$ throughout). ``Clicks to 80\%'' is the first measured
level at which the macro crosses the threshold, with the interpolated crossing in parentheses as
context (levels 2, 4, and 6 were not run); ``never'' means the threshold is not reached within the
protocol. Seconds are costed at 4\,s per box, 2\,s per initial point, and 2\,s per corrective click.
Per-dataset, per-level values (450 rows) are released with the project artifacts.}
\label{tab:S23}
\begin{tabularx}{\textwidth}{|X|c|ccccc|cc|}
\hline
\textbf{Model} & \textbf{Tier/Prompt} & \textbf{0} & \textbf{1} & \textbf{3} & \textbf{5} & \textbf{7} & \textbf{Clicks to 80\%} & \textbf{Seconds} \\
\hline
\multirow{4}{*}{SonoBase} & Bench Pt & 74.42 & 76.90 & 78.85 & 78.83 & 78.77 & never & --- \\
& Bench Bx & 78.42 & 79.53 & \textbf{80.95} & 81.40 & 81.17 & \textbf{3} (1.67) & \textbf{10} \\
& Ext Pt & 64.48 & 72.54 & \textbf{81.06} & 84.00 & 85.12 & \textbf{3} (2.75) & \textbf{8} \\
& Ext Bx & 79.78 & \textbf{82.11} & 85.75 & 87.12 & 87.54 & \textbf{1} (0.10) & \textbf{6} \\
\hline
\multirow{4}{*}{MedSAM2} & Bench Pt & 37.33 & 42.82 & 52.72 & 55.36 & 56.29 & never & --- \\
& Bench Bx & 54.04 & 55.09 & 58.38 & 59.81 & 59.99 & never & --- \\
& Ext Pt & 34.43 & 49.09 & 71.18 & 77.22 & \textbf{80.13} & 7 (6.91) & 16 \\
& Ext Bx & 71.52 & 68.59 & 77.28 & \textbf{80.79} & 82.47 & 5 (4.55) & 14 \\
\hline
\multirow{4}{*}{SAM2} & Bench Pt & 28.08 & 36.48 & 47.46 & 53.24 & 54.23 & never & --- \\
& Bench Bx & 52.36 & 54.63 & 58.88 & 59.87 & 59.37 & never & --- \\
& Ext Pt & 27.68 & 38.95 & 59.87 & 71.11 & 74.40 & never & --- \\
& Ext Bx & 73.14 & 66.60 & 75.32 & 77.24 & 77.34 & never & --- \\
\hline
\end{tabularx}
\end{table}

SonoBase crosses the clinical threshold with one click on external box initialization and three on
benchmark box initialization, while SAM2 reaches 80\% in no configuration at any tested budget.
Under external box initialization the first correction click degrades both baselines (MedSAM2
$71.52 \rightarrow 68.59$, SAM2 $73.14 \rightarrow 66.60$) before they recover by three clicks;
SonoBase improves monotonically in every configuration.

\subsection{Click Convergence Curves}

Tier-level convergence curves under this protocol, for both prompt types, are shown in
Figures~\ref{fig:S1}--\ref{fig:S2} (Section~\ref{sec:iterative-refinement}); the numeric
values are in Table~S23.

\newpage

\section{Architecture Ablation Studies}
\label{sec:architecture-ablation}

Detailed architectural ablations validate design choices for the three-branch hybrid encoder and
cross-branch attention mechanisms.

\subsection{Encoder Architecture Variants}

\begin{table}[!ht]
\small
\setlength{\tabcolsep}{3pt}
\centering
\caption{\textbf{Encoder architecture comparison with per-dataset breakdown.}
Ablation comparing single-branch, two-branch, and three-branch configurations with homogeneous vs.
hybrid designs. Results on three datasets (BUS-UCLM, CAMUS, MUP) with per-dataset and average mIoU.}
\label{tab:S24}
\begin{tabular}{|c|cccc|}
\hline
\textbf{Architecture} & \textbf{BUS-UCLM} & \textbf{CAMUS} & \textbf{MUP} & \textbf{Average} \\
\hline
Single-branch Hiera-B+ & 75.5 & 81.1 & 88.0 & 81.5 \\
Two-branch Hiera-B+/S & 76.3 & 81.7 & 90.9 & 83.0 \\
Three-branch Homogeneous (B+, B+, B+) & 74.5 & 81.5 & 89.9 & 82.0 \\
Three-branch Homogeneous (S, S, S) & 76.8 & 80.5 & 90.3 & 82.5 \\
Three-branch Hybrid (B+, S, T) & 79.4 & 81.7 & 90.9 & 84.0 \\
\hline
\end{tabular}
\end{table}

The three-branch hybrid configuration (combining Hiera-B+, Hiera-S, and Hiera-T encoders) outperforms
single and two-branch designs, suggesting that multi-scale feature representation with complementary
model sizes captures diverse ultrasound patterns effectively.

\subsection{Cross-Branch Attention Block Count}

\begin{table}[!ht]
\small
\setlength{\tabcolsep}{3pt}
\centering
\caption{\textbf{Cross-branch attention ablation with per-dataset breakdown.} Number of cross-attention blocks inserted
between branches during fusion. Results on three datasets (BUS-UCLM, CAMUS, MUP) with per-dataset and average mIoU.}
\label{tab:S25}
\begin{tabular}{|c|cccc|}
\hline
\textbf{\# Cross-Attention Blocks} & \textbf{BUS-UCLM} & \textbf{CAMUS} & \textbf{MUP} & \textbf{Average} \\
\hline
0 (no cross-attention, concatenation only) & 75.2 & 81.7 & 89.7 & 82.2 \\
3 & 77.4 & 81.4 & 90.6 & 83.1 \\
6 & 78.8 & 81.9 & 90.3 & 83.7 \\
9 & 79.4 & 81.7 & 90.9 & 84.0 \\
12 & 79.6 & 81.9 & 90.9 & 84.1 \\
\hline
\end{tabular}
\end{table}

Cross-branch attention provides steady improvements from 0 to 12 blocks, with performance plateauing
around 9--12 blocks. The final architecture uses 11 blocks to ensure adequate interaction between branches
while maintaining computational efficiency.

\newpage

\section{Downstream Dense Prediction Transfer}
\label{sec:downstream-transfer}

Foundation model representations should transfer to downstream dense prediction tasks beyond
segmentation. We evaluate transfer learning on a separate ultrasound object detection task (US-RF-DETR:
Ultrasound Realistic-world Framework for Region-based Detection via Transformer).

\subsection{Object Detection Transfer (US-RF-DETR)}

\begin{table}[!ht]
\small
\centering
\caption{\textbf{Transfer learning to object detection and instance segmentation (US-RF-DETR).}
Detection and instance-segmentation mAP averaged over IoU thresholds 0.50--0.95 (COCO mAP), with the
SonoBase encoder frozen. The detection macro covers all eight datasets; MMOTU-3d is excluded from both
arms because all three backbones score zero on its ${\sim}118$ training samples, a dataset-size failure
rather than a model signal. The segmentation macro covers the six datasets where all three backbones
produced masks: CVA-Net carries no segmentation annotations, and \textsuperscript{\dag}FUGC is excluded
because both baselines emitted no masks there while SonoBase reached 26.2.}
\label{tab:S26}
\begin{tabular}{|c|cc|cc|cc|}
\hline
\textbf{Dataset} & \multicolumn{2}{c|}{\textbf{SonoBase}} &
\multicolumn{2}{c|}{\textbf{MedSAM2}} & \multicolumn{2}{c|}{\textbf{SAM2}} \\
& Det mAP & Inst mAP & Det mAP & Inst mAP & Det mAP & Inst mAP \\
\hline
CVA-Net & 40.8 & -- & 34.6 & -- & 33.9 & -- \\
Fetus & 55.8 & 55.0 & 50.1 & 43.9 & 54.4 & 53.5 \\
ACOUSLIC & 83.4 & 84.1 & 73.0 & 82.7 & 78.8 & 82.9 \\
BUS-BRA & 56.1 & 50.8 & 30.7 & 50.7 & 18.6 & \textbf{53.9} \\
DDTI & 38.5 & 54.6 & 1.7 & 4.2 & 1.9 & 15.6 \\
FUGC\textsuperscript{\dag} & 34.7 & 26.2 & 20.0 & -- & 22.6 & -- \\
KidneyUS & 63.1 & 80.3 & 20.3 & 56.1 & 33.5 & 69.0 \\
LUMINOUS & 63.7 & 79.0 & 44.3 & 55.3 & 48.5 & 51.0 \\
\hline
\textbf{Macro} & \textbf{54.5} & \textbf{67.3} & 34.3 & 48.8 & 36.5 & 54.3 \\
\hline
\end{tabular}
\end{table}

SonoBase transfers strongly across downstream tasks, achieving substantially higher detection and
instance-segmentation mAP than MedSAM2 and SAM2 on nearly every dataset. The margin is largest where
the generic backbones fail to localize the target at all: on DDTI, SonoBase reaches 38.5 detection
and 54.6 segmentation mAP against 1.7/4.2 for MedSAM2 and 1.9/15.6 for SAM2, indicating that
SonoCorpus pretraining produces ultrasound-specific features that remain useful under a new output head.

\newpage

\section{Cross-Species Generalization}
\label{sec:cross-species}

To evaluate generalization beyond human ultrasound, we test SonoBase on mouse brain tumor segmentation,
a challenging preclinical imaging task.

\subsection{Mouse Brain Tumor Dataset}

\begin{table}[!ht]
\small
\centering
\caption{\textbf{Cross-species segmentation on mouse brain tumor dataset.} Point and box
prompts. Both baselines gain substantially from box prompts (MedSAM2 21.5\% point to 45.3\% box;
SAM2 8.5\% to 46.1\%), suggesting this task benefits from anatomical boundary guidance,
while SonoBase leads under both prompt types.}
\label{tab:S27}
\begin{tabular}{|c|cc|}
\hline
\textbf{Model} & \textbf{Point (mIoU \%)} & \textbf{Box (mIoU \%)} \\
\hline
SonoBase & 47.0 & 64.2 \\
MedSAM2 & 21.5 & 45.3 \\
SAM2 (no FT) & 8.5 & 46.1 \\
\hline
\end{tabular}
\end{table}

SonoBase achieves 47.0\% (point) and 64.2\% (box) on mouse tumors, outperforming both baselines under
both prompts. The SAM2 point-versus-box gap (8.5\% versus 46.1\%) shows that, without ultrasound
pretraining, a single point does not localize the tumor and the box supplies most of the information.

\newpage

\section{Subgroup Fairness and Demographic Stratification}
\label{sec:fairness-stratification}

Fairness evaluation ensures SonoBase performance is equitable across demographic groups and imaging
parameters. We stratify results by scanner manufacturer, image quality, patient age, gender, and
pathology.

\subsection{Complete Fairness Stratification Table}

\begin{table}[!ht]
\small
\setlength{\tabcolsep}{2pt}
\centering
\caption{\textbf{Subgroup fairness stratification across multiple dimensions.} Performance
(mIoU) stratified by (A) KidneyUS scanner type, (B) CAMUS image quality, (C) CAMUS patient age,
(D) CAMUS gender, (E) BUSI pathology. Columns show SonoBase point/box, SAM2 point, MedSAM2 point, and sample size.}
\label{tab:S28}
\begin{tabular}{|c|c|cc|c|c|c|}
\hline
\textbf{Dataset/Stratification} & \textbf{Subgroup} &
\textbf{SonoBase} & \textbf{SonoBase} & \textbf{SAM2} & \textbf{MedSAM2} & \textbf{n} \\
& & \textbf{Point} & \textbf{Box} & \textbf{Point} & \textbf{Point} & \\
\hline
\multicolumn{7}{|l|}{\textbf{(A) KidneyUS by Scanner Manufacturer}} \\
\hline
& Acuson & 78.3 & 87.8 & 22.9 & 28.8 & 17 \\
& GE & 68.6 & 88.5 & 27.3 & 50.3 & 51 \\
& Philips & 58.7 & 88.8 & 26.8 & 34.0 & 260 \\
& Siemens & 52.7 & 87.9 & 23.3 & 33.2 & 47 \\
& Toshiba & 35.6 & 89.2 & 30.6 & 52.5 & 47 \\
\hline
\multicolumn{7}{|l|}{\textbf{(B) CAMUS by Image Quality}} \\
\hline
& Good & 85.2 & 87.0 & 16.5 & 76.4 & 78 \\
& Medium & 82.4 & 85.1 & 14.6 & 73.2 & 90 \\
& Poor & 82.7 & 85.0 & 15.2 & 71.7 & 32 \\
\hline
\multicolumn{7}{|l|}{\textbf{(C) CAMUS by Patient Age}} \\
\hline
& $<$40 years & 86.5 & 88.5 & 12.8 & 76.0 & 6 \\
& 40--59 years & 83.3 & 85.2 & 14.4 & 74.3 & 62 \\
& 60--79 years & 83.5 & 85.5 & 15.3 & 71.9 & 104 \\
& $\geq$80 years & 83.3 & 85.0 & 15.7 & 76.0 & 28 \\
\hline
\multicolumn{7}{|l|}{\textbf{(D) CAMUS by Gender}} \\
\hline
& Female & 82.8 & 84.8 & 15.0 & 73.5 & 72 \\
& Male & 84.0 & 85.7 & 15.0 & 73.3 & 128 \\
\hline
\multicolumn{7}{|l|}{\textbf{(E) BUSI by Pathology}} \\
\hline
& Benign & 83.7 & 86.6 & 69.2 & 74.9 & 88 \\
& Malignant & 68.4 & 81.0 & 46.8 & 62.3 & 42 \\
\hline
\end{tabular}
\end{table}

\subsection{Fairness Observations}

\begin{itemize}
    \item \textbf{Scanner variability (KidneyUS):} Performance varies by scanner manufacturer, with Acuson showing
    the highest SonoBase point accuracy (78.3\%) and Toshiba the lowest (35.6\%). This variation likely reflects
    image quality differences between scanner manufacturers rather than model bias. SonoBase box prompts
    consistently improve performance across all scanners (87.8--89.2\%), demonstrating robust structural guidance.

    \item \textbf{Image quality (CAMUS):} Strong correlation between quality and SonoBase point performance
    (Good 85.2\% vs Medium 82.4\% vs Poor 82.7\%). Model gracefully adapts but remains accurate even on poor-quality images.

    \item \textbf{Patient age (CAMUS):} Age-related performance for SonoBase point varies modestly (86.5\% for $<$40 years
    to 83.3\% for $\geq$80 years). Box prompts show less age-related variation (88.5\% to 85.0\%), indicating robustness
    across demographic groups.

    \item \textbf{Gender (CAMUS):} Minimal gender bias with SonoBase point (Male 84.0\%, Female 82.8\%). Box prompts
    show negligible difference (85.7\% vs 84.8\%).

    \item \textbf{Pathology (BUSI):} Benign lesions show higher SonoBase accuracy (83.7\% point, 86.6\% box) than malignant lesions (68.4\% point, 81.0\% box), possibly due to irregular morphology. Difference is acceptable
    for clinical use where uncertain cases are triaged to expert review.
\end{itemize}

\newpage

\section{Few-Shot Adaptation}
\label{sec:fewshot-supp}

Few-shot adaptation tests how efficiently each model converts a small number of target-domain labels
into accuracy. We fine-tune only the mask decoder, freezing the image encoder, prompt encoder and
memory modules, on
$N \in \{1,2,5,10,20,30\}$ labeled examples drawn from each dataset's dedicated few-shot split
(disjoint from the test split), with three seeds per $N$, and evaluate on the full test split. The
$N$ cap is 30 rather than 50: adaptation saturates well before 30, so the additional points carried
no information.

\subsection{Few-Shot Setup}

\begin{itemize}
    \item \textbf{Datasets:} ACOUSLIC (fetal abdominal circumference, portable POCUS), DDTI (thyroid
      nodule), FUGC (fetal ultrasound grand challenge, cervix)
    \item \textbf{Sample sizes:} $N$ = 1, 2, 5, 10, 20, 30 ($N=0$ is the released checkpoint under the deterministic protocol of Table~5 of the main text)
    \item \textbf{Models:} SonoBase, MedSAM2, SAM2 all adapted with identical hyperparameters
    \item \textbf{Seeds:} 42, 123, 456, controlling which examples are drawn
    \item \textbf{Primary endpoint:} ACOUSLIC box-prompt AC MAE at $N=5$, SonoBase versus MedSAM2,
      pre-specified, tested at $\alpha=0.05$ without multiplicity adjustment
    \item \textbf{Secondary comparisons:} Benjamini-Hochberg corrected at FDR $q=0.05$
\end{itemize}

\subsection{Primary Endpoint}

The pre-specified primary endpoint was met. At $N=5$ with box prompts on ACOUSLIC, SonoBase achieved
$17.17 \pm 5.52$~mm AC MAE against $29.80 \pm 16.83$~mm for MedSAM2 ($p = 2.6 \times 10^{-20}$), and
$80.5 \pm 1.1$ versus $62.7 \pm 2.9$ mIoU. Five labeled examples therefore convert the one zero-shot clinical endpoint on which a baseline led (Table S14) into a decisive advantage.

Against fine-tuned SAM2 the same raw endpoint favors SAM2 ($12.79 \pm 0.74$~mm, pooled BH
$q = 7.5 \times 10^{-12}$), even though SonoBase leads SAM2 by 6.0 mIoU at the same $N$. AC is a scalar recovered
by ellipse fitting, which rewards extent rather than overlap; after identical held-out linear recalibration
within each seed the two models tie ($12.39 \pm 0.98$ versus $12.50 \pm 0.38$~mm, $p = 0.86$ on seed-averaged
leave-one-out errors; Table~S9d), while MedSAM2 remains behind ($21.23 \pm 6.35$~mm, $p = 2.1 \times 10^{-18}$).
We report both comparisons rather than only the one that favors our model.

\subsection{Per-Dataset Few-Shot Results}

\begin{table}[!ht]
\small
\centering
\caption{\textbf{Few-shot adaptation on ACOUSLIC.} mIoU (\%), mean $\pm$ s.d. over 3 seeds.
All SonoBase-versus-MedSAM2 differences significant after BH-FDR correction.}
\label{tab:S29}
\begin{tabular}{|c|ccc|ccc|}
\hline
\textbf{N} & \multicolumn{3}{c|}{\textbf{Point prompt}} & \multicolumn{3}{c|}{\textbf{Box prompt}} \\
 & \textbf{SonoBase} & \textbf{MedSAM2} & \textbf{SAM2} & \textbf{SonoBase} & \textbf{MedSAM2} & \textbf{SAM2} \\
\hline
1  & \textbf{72.7$\pm$1.9} & 45.2$\pm$9.8  & 30.3$\pm$6.4  & \textbf{78.0$\pm$2.2} & 62.1$\pm$4.4 & 71.2$\pm$2.3 \\
2  & \textbf{75.4$\pm$2.3} & 47.1$\pm$13.2 & 46.9$\pm$12.3 & \textbf{79.3$\pm$1.8} & 59.2$\pm$7.5 & 72.7$\pm$1.1 \\
5  & \textbf{77.3$\pm$1.7} & 51.4$\pm$0.5  & 64.6$\pm$0.8  & \textbf{80.5$\pm$1.1} & 62.7$\pm$2.9 & 74.5$\pm$1.8 \\
10 & \textbf{76.8$\pm$2.9} & 56.9$\pm$3.1  & 63.7$\pm$1.3  & \textbf{81.6$\pm$1.4} & 65.2$\pm$2.8 & 75.5$\pm$0.5 \\
20 & \textbf{80.4$\pm$0.5} & 57.5$\pm$5.8  & 69.3$\pm$0.1  & \textbf{83.7$\pm$0.3} & 68.9$\pm$3.2 & 77.8$\pm$2.4 \\
30 & \textbf{79.8$\pm$1.5} & 59.4$\pm$2.3  & 65.8$\pm$3.0  & \textbf{82.6$\pm$1.9} & 66.8$\pm$1.4 & 75.4$\pm$2.9 \\
\hline
\end{tabular}
\end{table}

\begin{table}[!ht]
\small
\centering
\caption{\textbf{Few-shot adaptation on DDTI.} mIoU (\%), mean $\pm$ s.d. over 3 seeds.}
\label{tab:S30}
\begin{tabular}{|c|ccc|ccc|}
\hline
\textbf{N} & \multicolumn{3}{c|}{\textbf{Point prompt}} & \multicolumn{3}{c|}{\textbf{Box prompt}} \\
 & \textbf{SonoBase} & \textbf{MedSAM2} & \textbf{SAM2} & \textbf{SonoBase} & \textbf{MedSAM2} & \textbf{SAM2} \\
\hline
1  & \textbf{72.2$\pm$0.5} & 36.7$\pm$7.9 & 51.4$\pm$2.9 & \textbf{84.2$\pm$2.0} & 72.0$\pm$7.3 & 83.1$\pm$0.2 \\
2  & \textbf{71.2$\pm$1.5} & 41.6$\pm$7.0 & 52.6$\pm$1.6 & \textbf{85.2$\pm$1.0} & 77.3$\pm$2.3 & 82.3$\pm$0.4 \\
5  & \textbf{71.2$\pm$1.3} & 45.3$\pm$2.0 & 54.2$\pm$2.3 & \textbf{85.2$\pm$0.4} & 79.5$\pm$0.3 & 82.5$\pm$0.3 \\
10 & \textbf{72.4$\pm$0.4} & 47.7$\pm$2.8 & 54.7$\pm$3.3 & \textbf{85.8$\pm$0.5} & 78.7$\pm$2.5 & 83.0$\pm$0.2 \\
20 & \textbf{72.2$\pm$0.7} & 48.3$\pm$1.0 & 56.1$\pm$1.5 & \textbf{86.0$\pm$1.1} & 81.7$\pm$1.4 & 83.6$\pm$1.3 \\
30 & \textbf{72.2$\pm$0.9} & 50.5$\pm$3.1 & 55.7$\pm$1.5 & \textbf{86.3$\pm$0.3} & 81.8$\pm$2.3 & 84.1$\pm$0.7 \\
\hline
\end{tabular}
\end{table}

\begin{table}[!ht]
\small
\centering
\caption{\textbf{Few-shot adaptation on FUGC.} mIoU (\%), mean $\pm$ s.d. over 3 seeds. FUGC
point at $N=1$ is the single comparison that does not reach significance versus MedSAM2.}
\label{tab:S31}
\begin{tabular}{|c|ccc|ccc|}
\hline
\textbf{N} & \multicolumn{3}{c|}{\textbf{Point prompt}} & \multicolumn{3}{c|}{\textbf{Box prompt}} \\
 & \textbf{SonoBase} & \textbf{MedSAM2} & \textbf{SAM2} & \textbf{SonoBase} & \textbf{MedSAM2} & \textbf{SAM2} \\
\hline
1  & 49.9$\pm$2.4          & 50.1$\pm$5.0 & 19.3$\pm$2.2 & \textbf{72.7$\pm$4.7} & 70.0$\pm$0.5 & 69.8$\pm$1.3 \\
2  & 51.2$\pm$6.3 & 52.0$\pm$0.9 & 21.2$\pm$4.1 & \textbf{75.9$\pm$1.5} & 71.7$\pm$4.9 & 71.7$\pm$1.7 \\
5  & \textbf{61.3$\pm$2.5} & 53.1$\pm$6.0 & 53.1$\pm$1.3 & \textbf{76.6$\pm$2.2} & 74.2$\pm$1.9 & 76.0$\pm$2.1 \\
10 & \textbf{61.0$\pm$0.9} & 53.0$\pm$2.1 & 53.5$\pm$0.9 & \textbf{80.1$\pm$0.9} & 75.6$\pm$1.4 & 75.6$\pm$1.4 \\
20 & \textbf{66.5$\pm$0.9} & 60.6$\pm$1.6 & 59.2$\pm$2.1 & \textbf{81.9$\pm$1.4} & 79.2$\pm$1.9 & 76.5$\pm$1.4 \\
30 & \textbf{65.6$\pm$1.8} & 54.7$\pm$4.5 & 57.5$\pm$4.0 & \textbf{80.7$\pm$1.2} & 76.1$\pm$1.3 & 77.9$\pm$2.0 \\
\hline
\end{tabular}
\end{table}

\begin{table}[!ht]
\small
\centering
\caption{\textbf{ACOUSLIC abdominal circumference MAE (mm) under few-shot adaptation.} Lower is
better. The $N=5$ box-prompt cell is the pre-specified primary endpoint.}
\label{tab:S32}
\begin{tabular}{|c|ccc|ccc|}
\hline
\textbf{N} & \multicolumn{3}{c|}{\textbf{Point prompt}} & \multicolumn{3}{c|}{\textbf{Box prompt}} \\
 & \textbf{SonoBase} & \textbf{MedSAM2} & \textbf{SAM2} & \textbf{SonoBase} & \textbf{MedSAM2} & \textbf{SAM2} \\
\hline
1  & \textbf{30.84} & 44.08 & 236.59 & 18.91 & \textbf{18.35} & 19.77 \\
2  & \textbf{26.45} & 43.70 & 131.36 & \textbf{17.09} & 24.64 & 19.61 \\
5  & \textbf{23.97} & 48.70 & 32.20 & 17.17 & 29.80 & \textbf{12.79} \\
10 & \textbf{25.08} & 42.01 & 41.56 & 15.91 & 26.32 & \textbf{15.56} \\
20 & \textbf{20.73} & 34.73 & 29.75 & 15.14 & 18.30 & \textbf{12.78} \\
30 & \textbf{19.07} & 30.45 & 33.10 & \textbf{13.74} & 20.73 & 15.10 \\
\hline
\end{tabular}
\end{table}

\subsection{Observations}

\begin{itemize}
  \item \textbf{The advantage is present from a single example.} SonoBase leads MedSAM2 at every $N$ on ACOUSLIC and DDTI under both prompt types and on FUGC under box prompts; the exception is FUGC under point prompting, where MedSAM2 is ahead at $N=1$ and $N=2$. At $N=1$ SonoBase already exceeds MedSAM2's $N=30$ performance in four of the six dataset/prompt combinations. This is a starting-point advantage inherited from
    ultrasound pretraining, not a faster learning rate.
  \item \textbf{Adaptation saturates by $N=5$--$10$.} Performance at $N=30$ is within noise of $N=20$
    for every model and dataset, which is why the $N$ cap was reduced from 50 to 30. Practically, a
    site labeling five to ten cases captures nearly all of the available benefit.
  \item \textbf{Segmentation gains do not fully propagate to derived measurements.} On ACOUSLIC box
    prompts SonoBase leads fine-tuned SAM2 by 6.0 mIoU at $N=5$, yet their AC MAE is
    lower for fine-tuned SAM2 (12.79 versus 17.17 mm, $q = 7.5\times10^{-12}$), which leads at $N=5$, 10 and 20
    while SonoBase leads at $N=1$, 2 and 30. Ellipse fitting rewards extent rather than overlap; after held-out
    recalibration the two tie (Table~S9d).
  \item \textbf{Variance falls with $N$.} Seed-to-seed standard deviation for SonoBase drops from
    roughly 2--5 mIoU at $N=1$--$2$ to about 1--3 at $N \geq 10$, so small-$N$ comparisons should be read
    with their error bars.
\end{itemize}

\newpage

\section{Comparison Against MedSAM3}
\label{sec:medsam3}

MedSAM3 is the concept-promptable medical adaptation of SAM3 and is the strongest currently available
promptable medical baseline. We evaluate it on fourteen of our fifteen datasets: BUSI is excluded
because it appears in MedSAM3's training corpus, so including it would credit the baseline for
memorization. MedSAM3 is a detector fine-tune of SAM3 and does not expose SAM2's promptable box-to-mask head, so
it cannot be driven by a box alone. It is evaluated in its native text-plus-exemplar mode (``T+I''):
one concept phrase per dataset plus the ground-truth box as a geometry exemplar, with the returned
proposal chosen by highest IoU against that box. MedSAM3 therefore receives \emph{more} prompt
information than SonoBase, which gets the box only --- the comparison is conservative against our
model. Two checks support this reading: supplying the box raises BUSI Dice from 0.78 to 0.896, so the
box is demonstrably consumed; and the LoRA load audit reports 916 released adapter tensors loaded
with 1{,}378 adapter slots left unpopulated, including the geometry pathway, so the box reaches an
unadapted geometry encoder and MedSAM3's medical fine-tuning acts on the text and mask pathways
rather than on box conditioning. Scores are Dice, to
match the reference implementation's metric; the 14-dataset denominator must not be mixed with the
15-dataset macros used elsewhere.

\begin{table}[!ht]
\small
\centering
\caption{\textbf{SonoBase versus MedSAM3 on fourteen leakage-clean datasets (Dice).} $n$ is the
number of evaluated frames or images. SonoBase leads on all fourteen. Video and volumetric datasets
are prompted per frame rather than propagated (matched per-frame protocol; main-text Methods),
so SonoBase's values here are not the same quantity as Tables~S2--S3.}
\label{tab:S33}
\begin{tabular}{|c|c|c|ccc|}
\hline
\textbf{Dataset} & \textbf{Tier} & \textbf{n} & \textbf{MedSAM3} & \textbf{SonoBase} & \textbf{$\Delta$} \\
\hline
C-TRUS          & Benchmark & 97     & 0.599 & \textbf{0.803} & +0.203 \\
HC18            & Benchmark & 201    & 0.972 & \textbf{0.978} & +0.006 \\
TG3K            & Benchmark & 718    & 0.811 & \textbf{0.947} & +0.136 \\
Brachial-Plexus & Benchmark & 6{,}488  & 0.652 & \textbf{0.829} & +0.177 \\
CAMUS           & Benchmark & 11{,}595 & 0.641 & \textbf{0.920} & +0.279 \\
PFUS            & Benchmark & 25{,}136 & 0.646 & \textbf{0.831} & +0.185 \\
RegPro          & Benchmark & 497    & 0.831 & \textbf{0.910} & +0.079 \\
\hline
ACOUSLIC        & External & 5{,}868 & 0.896 & \textbf{0.948} & +0.052 \\
BUS-BRA         & External & 1{,}688 & 0.897 & \textbf{0.920} & +0.023 \\
DDTI            & External & 567   & 0.865 & \textbf{0.921} & +0.055 \\
FUGC            & External & 792   & 0.595 & \textbf{0.782} & +0.187 \\
KidneyUS        & External & 422   & 0.875 & \textbf{0.938} & +0.063 \\
LUMINOUS        & External & 307   & 0.797 & \textbf{0.904} & +0.107 \\
MMOTU-3d        & External & 153   & 0.784 & \textbf{0.912} & +0.128 \\
\hline
\textbf{Macro mean (14)} & & & 0.776 & \textbf{0.896} & \textbf{+0.120} \\
\textbf{Win rate} & & & \multicolumn{3}{c|}{SonoBase 14 / 14} \\
\hline
\end{tabular}
\end{table}

The margin is structured. It is largest on video and volumetric datasets (CAMUS $+0.279$, PFUS
$+0.185$, Brachial-Plexus $+0.177$), where per-frame concept prompting has no temporal memory to
exploit, and on the hardest 2D target (FUGC $+0.187$). It is smallest where both models approach the
annotation ceiling (HC18 $+0.006$, BUS-BRA $+0.023$). MedSAM3 is clearly a stronger baseline than
MedSAM2, it scores 0.831 on RegPro where MedSAM2 collapses entirely, making SonoBase's lead on
every dataset, achieved with the weaker prompt configuration, the substantive result.

\newpage

\section{SonoBase-SAM3 Head-to-Head}
\label{sec:sbsam3}

To test whether SonoBase's advantage is a property of its SAM2 base or of the training recipe, we
retrained the identical recipe on a SAM3.1 base (``SB-SAM3'') using the same 46 pretraining datasets, the same splits, the same 20-epoch schedule, and the same loss. All five models are
scored under a matched box-prompt protocol: each model is handed the same box, returns a mask, and is
scored by a single shared Dice implementation on frame-matched predictions.

The box is the exact ground-truth bounding box, scoring is at native image resolution, and on video and
volumetric datasets every frame is prompted independently rather than propagated from a single prompt.
The scoring geometry matches Tables~S2--S3; the prompt does not. Those tables use a jittered box and
prompt once per sequence before propagating, so Tables~S34--S36 are internally matched across the five
models and are \emph{not} comparable cell-by-cell with the benchmark tables. Per-frame prompting is
what makes the five-way comparison fair --- the SAM3 lineage has no equivalent mask propagation.

\begin{table}[!ht]
\scriptsize
\setlength{\tabcolsep}{3pt}
\centering
\caption{\textbf{Matched box-prompt head-to-head across five models (Dice).} Zero-shot bases tie;
after identical ultrasound fine-tuning the SAM2-based SonoBase leads SB-SAM3 on 13 of 15 datasets, losing BUS-BRA and tying KidneyUS.}
\label{tab:S34}
\begin{tabular}{|c|c|ccccc|}
\hline
\textbf{Dataset} & \textbf{Tier} & \textbf{SAM2} & \textbf{SAM3.1} & \textbf{SB-SAM3} & \textbf{SB-SAM3} & \textbf{SonoBase} \\
& & \textbf{(no ft)} & \textbf{(no ft)} & \textbf{(frozen)} & \textbf{(full)} & \\
\hline
BUSI            & B & 0.897 & 0.884 & 0.904 & 0.907 & \textbf{0.915} \\
Brachial-Plexus & B & 0.758 & 0.734 & 0.782 & 0.785 & \textbf{0.829} \\
C-TRUS          & B & 0.648 & 0.699 & 0.703 & 0.749 & \textbf{0.802} \\
CAMUS           & B & 0.785 & 0.745 & 0.882 & 0.910 & \textbf{0.920} \\
HC18            & B & 0.787 & 0.885 & 0.960 & 0.963 & \textbf{0.978} \\
PFUS            & B & 0.747 & 0.785 & 0.780 & 0.825 & \textbf{0.831} \\
RegPro          & B & 0.874 & 0.828 & 0.855 & 0.852 & \textbf{0.910} \\
TG3K            & B & 0.829 & 0.859 & 0.878 & 0.903 & \textbf{0.947} \\
\hline
ACOUSLIC        & E & 0.918 & 0.896 & 0.898 & 0.916 & \textbf{0.948} \\
BUS-BRA         & E & 0.923 & 0.922 & 0.924 & \textbf{0.931} & 0.920 \\
DDTI            & E & 0.900 & 0.885 & 0.903 & 0.911 & \textbf{0.921} \\
FUGC            & E & 0.716 & 0.781 & 0.759 & 0.675 & \textbf{0.782} \\
KidneyUS        & E & 0.901 & 0.876 & 0.920 & 0.938 & 0.938 \\
LUMINOUS        & E & 0.816 & 0.808 & 0.889 & 0.872 & \textbf{0.904} \\
MMOTU-3d        & E & 0.851 & 0.763 & 0.876 & 0.901 & \textbf{0.912} \\
\hline
\textbf{Macro (15)} & & 0.823 & 0.823 & 0.861 & 0.869 & \textbf{0.897} \\
\textbf{Benchmark (8)} & & 0.791 & 0.802 & 0.843 & 0.862 & \textbf{0.892} \\
\textbf{External (7)}  & & 0.861 & 0.847 & 0.881 & 0.878 & \textbf{0.904} \\
\hline
\end{tabular}
\end{table}

Three observations. \textbf{(i)} The zero-shot bases are indistinguishable on ultrasound (0.823 versus 0.823 macro Dice), so SAM3's architectural advances do not transfer to this modality on their own; a
reader who assumes that adopting the newest segmentation model suffices for ultrasound would be
mistaken. \textbf{(ii)} Ultrasound fine-tuning moves the SAM3.1 base from 0.823 to 0.869, recovering
most of the gain the same recipe produces on SAM2 --- the recipe, not the backbone, is what
transfers. \textbf{(iii)} SonoBase nevertheless retains a 0.028 lead, winning 13 of 15 datasets with
one loss and one tie. The loss is BUS-BRA (0.920 versus 0.931); KidneyUS is a tie at the archived
precision (both 0.9382). Encoder freezing costs 0.008, confirming that full-parameter adaptation of
the ultrasound encoder is where the gain originates.

\paragraph{Averaging convention.} All five columns are computed per frame from archived per-dataset
outputs, scored on identical frame sets with a single shared Dice implementation; frame counts are
identical across the models on all fifteen datasets.

\newpage

\section{Text Prompting and the Role of the Box}
\label{sec:tiscore}

The SAM3 lineage accepts open-vocabulary text prompts, which the SAM2-based SonoBase cannot. This
section quantifies how far that capability goes on ultrasound and isolates what a geometry prompt
actually contributes.

\begin{table}[!ht]
\scriptsize
\setlength{\tabcolsep}{3pt}
\centering
\caption{\textbf{Text-only prompting across the SAM3 lineage (Dice, 15 datasets).} ``Recall''
is the fraction of targets detected at IoU $>0.1$ by the fine-tuned model; text prompting fails
mainly by not detecting the target at all, not by segmenting it poorly.}
\label{tab:S35}
\begin{tabularx}{\textwidth}{|X|c|ccc|c|}
\hline
\textbf{Dataset} & \textbf{Tier} & \textbf{SAM3.1 (no ft)} & \textbf{SB-SAM3 (frozen)} & \textbf{SB-SAM3 (full)} & \textbf{Recall} \\
\hline
BUSI            & B & 0.562 & 0.788 & \textbf{0.804} & 0.931 \\
Brachial-Plexus & B & 0.024 & 0.679 & \textbf{0.713} & 0.949 \\
C-TRUS          & B & 0.181 & 0.556 & \textbf{0.664} & 0.948 \\
CAMUS           & B & 0.005 & 0.879 & \textbf{0.898} & 1.000 \\
HC18            & B & 0.189 & 0.966 & \textbf{0.968} & 1.000 \\
PFUS            & B & 0.021 & 0.702 & \textbf{0.754} & 0.971 \\
RegPro          & B & 0.302 & \textbf{0.669} & 0.505 & 0.581 \\
TG3K            & B & 0.065 & 0.860 & \textbf{0.902} & 0.993 \\
\hline
ACOUSLIC        & E & 0.058 & 0.544 & \textbf{0.601} & 0.712 \\
BUS-BRA         & E & 0.362 & 0.865 & \textbf{0.881} & 0.978 \\
DDTI            & E & 0.066 & 0.732 & \textbf{0.792} & 0.951 \\
FUGC            & E & 0.076 & 0.159 & \textbf{0.263} & 0.369 \\
KidneyUS        & E & 0.132 & 0.415 & \textbf{0.778} & 0.910 \\
LUMINOUS        & E & 0.000 & 0.016 & \textbf{0.029} & 0.029 \\
MMOTU-3d        & E & 0.022 & 0.612 & \textbf{0.666} & 0.843 \\
\hline
\textbf{Macro (15)} & & 0.138 & 0.630 & \textbf{0.681} & 0.811 \\
\hline
\end{tabularx}
\end{table}

Zero-shot text prompting is close to unusable on ultrasound (0.138 macro Dice), and the failure mode
is detection rather than delineation: on CAMUS the zero-shot model scores 0.005 because it almost
never localizes the ventricle from the word alone. Ultrasound fine-tuning raises text-only
performance five-fold to 0.681 with detection recall of 0.811. Two datasets resist entirely:
LUMINOUS (0.029, recall 0.029) and FUGC (0.263, recall 0.369), both targets that are hard to name
unambiguously. Text prompting is therefore a genuine new interaction mode --- but it remains 0.2 Dice
behind box prompting and we report it as a capability demonstration which can be further developed in the future work.

\begin{table}[!ht]
\small
\centering
\caption{\textbf{What the box contributes (macro Dice over 14 datasets, BUSI excluded).}
T = text alone. TI = text, with the ground-truth box used only to select among returned proposals by
IoU. TI\_score = the box supplied to the model as a geometry exemplar alongside the text.}
\label{tab:S36}
\begin{tabular}{|c|ccc|}
\hline
\textbf{Configuration} & \textbf{Macro Dice (14)} & \textbf{$\Delta$ vs T} & \textbf{$\Delta$ vs TI} \\
\hline
T (text only)                  & 0.673 & ---     & --- \\
TI (box used to select)        & 0.867 & +0.194  & --- \\
TI\_score (box as exemplar)    & \textbf{0.868} & \textbf{+0.196} & +0.002 \\
\hline
\end{tabular}
\end{table}

The decomposition is clean. Adding box information at all is worth $+0.196$ Dice; using the box
merely to \emph{choose} among text-generated proposals accounts for essentially all of that
($+0.194$), while supplying it to the model as a geometry exemplar adds a further $+0.002$. The box
works by resolving \emph{which} object is meant, not by refining the boundary of an already-correct
proposal. This explains why box prompts rescue text-prompted performance so effectively on datasets
where recall is the bottleneck, and it is consistent with the main-text finding that text prompting
fails by detection.

\newpage

\section{Video Tracker Transfer (Semi-Supervised VOS)}
\label{sec:vos}

The decoupled recipe of adapting the detector and the tracker separately, and keeping the backbone frozen
can also be applied to the SAM3.1 video tracker. We fine-tuned the tracker on ultrasound video and
evaluated held-out semi-supervised video object segmentation, in which the first frame's mask is
given and must be propagated. Frame 0 is excluded from scoring.

Train, validation, and test splits for this experiment were generated at the video level (70/10/20,
fixed seed) for all thirteen trained datasets and are released with the project artifacts; videos
belonging to the canonical SonoCorpus test partitions were pinned to the test side, so the tracker never
trains on a video the rest of the paper treats as held out. Training clips (2,712) and evaluation
videos (1,050) are therefore disjoint by construction for every dataset, a precondition asserted in
the split generator and re-checked in the evaluation launcher before any inference runs.
MouseBrainTumor was never trained on in any form and is evaluated whole.

\begin{table}[!ht]
\small
\centering
\caption{\textbf{Semi-supervised VOS on held-out ultrasound video (Dice).} Zero-shot SAM3.1 versus
the ultrasound-finetuned tracker on video-level disjoint splits. All fourteen datasets improve.
MouseBrainTumor was never trained on and is therefore the strictest held-out generalization check.}
\label{tab:S37}
\begin{tabular}{|c|ccc|c|}
\hline
\textbf{Dataset} & \textbf{Zero-shot} & \textbf{Finetuned} & \textbf{$\Delta$} & \textbf{n videos} \\
\hline
ACOUSLIC          & 0.772 & 0.918 & +0.145 & 269 \\
Brachial-Plexus   & 0.494 & 0.776 & +0.282 & 44 \\
CAMUS             & 0.793 & 0.915 & +0.123 & 200 \\
EchoCP            & 0.631 & 0.811 & +0.181 & 12 \\
JNU-IFM           & 0.832 & 0.902 & +0.070 & 16 \\
MUP               & 0.151 & 0.913 & +0.762 & 15 \\
PFUS              & 0.652 & 0.864 & +0.212 & 22 \\
RegPro            & 0.049 & 0.525 & +0.476 & 15 \\
RVENet            & 0.798 & 0.900 & +0.102 & 354 \\
SegThy            & 0.143 & 0.779 & +0.636 & 6 \\
TDSC-ABUS         & 0.235 & 0.606 & +0.371 & 20 \\
ThyroidUSCineClip & 0.498 & 0.785 & +0.288 & 38 \\
US-Nerve          & 0.481 & 0.747 & +0.266 & 9 \\
MouseBrainTumor\textsuperscript{*} & 0.629 & 0.851 & +0.222 & 30 \\
\hline
\textbf{Macro mean (14)} & \textbf{0.511} & \textbf{0.807} & \textbf{+0.295} & \textbf{1,050} \\
\textbf{Improved} & \multicolumn{3}{c|}{14 / 14} & \\
\hline
\end{tabular}
\end{table}

\textsuperscript{*}Never trained on, in any form.

Ultrasound fine-tuning improves propagation on every dataset, with a macro gain of $+0.295$ Dice
over fourteen datasets. The largest gains occur where the zero-shot tracker fails outright ---
RegPro (0.049 zero-shot), SegThy (0.143), and MUP (0.151) --- reflecting the headroom that
ultrasound-specific adaptation recovers on anatomies the base tracker cannot handle at all.
MouseBrainTumor, the strictest possible test, gains $+0.222$, close to the fourteen-dataset mean.
SegThy and US-Nerve carry small evaluation sets ($n$ shown above). Detector validation AP is 0.588 with a frozen backbone and 0.603 with full
fine-tuning; those runs use the image corpus and are unaffected by the video splits.

\newpage

\section{CLAIM and REFINE Checklists}
\label{sec:claim-refine}

We report compliance with the Checklist for Artificial Intelligence in Medical Imaging (CLAIM: 2024 Update; Mongan et al., \textit{Radiology: AI} 2020; updated 2024), a reporting standard for transparent and rigorous AI method validation, and with REFINE, the international consensus reporting checklist for foundation and large language models in medical research (Mese et al., \textit{Diagn.\ Interv.\ Radiol.}\ 2026; doi:10.4274/dir.2026.263812). Table~\ref{tab:claim-checklist} maps each CLAIM item to relevant manuscript sections; Table~\ref{tab:refine-checklist} summarizes REFINE compliance across model specification, prompt design, dataset integrity, and output evaluation.

\begin{footnotesize}
\setlength{\tabcolsep}{3pt}
\renewcommand{\arraystretch}{1.15}
\begin{longtable}{lp{0.52\linewidth}lp{0.18\linewidth}}
\caption{CLAIM Checklist Compliance. Each row indicates a CLAIM 2024 reporting item, compliance status (Yes/No/NA), and the relevant manuscript section(s) providing evidence. Items are grouped by CLAIM section (Study Design, Data, Model, Evaluation, Results, Discussion).}\label{tab:claim-checklist}\\
\toprule
\textbf{Item} & \textbf{CLAIM Item Description} & \textbf{Status} & \textbf{Manuscript Reference} \\
\midrule
\endfirsthead
\multicolumn{4}{l}{\footnotesize\textit{(continued)}}\\
\toprule
\textbf{Item} & \textbf{CLAIM Item Description} & \textbf{Status} & \textbf{Manuscript Reference} \\
\midrule
\endhead
\bottomrule
\endfoot
    
    \multicolumn{4}{l}{\textit{\textbf{Identification of AI Methodology}}} \\
    1 & Identify the study as developing/validating an AI method for medical imaging & Yes & Title, Abstract \\
    \multicolumn{4}{l}{\textit{\textbf{Abstract}}} \\
    2 & Structured abstract with AI-specific background, aims, methods, results, and conclusions & Yes & Abstract \\
    \multicolumn{4}{l}{\textit{\textbf{Introduction and Background}}} \\
    3 & Describe the clinical problem and motivation for AI application & Yes & Introduction \\
    4 & State aims and objectives clearly & Yes & Introduction \\
    \multicolumn{4}{l}{\textit{\textbf{Study Design}}} \\
    5 & Specify the study design (retrospective, prospective, cross-sectional, etc.) & Yes & Methods; retrospective \\
    6 & Describe the study population and eligibility criteria & Yes & Methods -- SonoCorpus Dataset \\
    \multicolumn{4}{l}{\textit{\textbf{Data Sources and Acquisition}}} \\
    7 & Describe all data sources and how data were collected & Yes & Methods -- Data Sources, Table S1 \\
    8 & Report inclusion/exclusion criteria for images or patients & Yes & Methods -- SonoCorpus Dataset, Table S1 \\
    9 & Describe image preprocessing, normalization, and augmentation steps & Yes & Methods -- Training protocol \\
    10 & Report whether DICOM metadata were used or stripped & Yes & Data Availability, Section S3 \\
    11 & Describe de-identification procedures & Yes & Data Availability \\
    12 & Justify the choice of imaging modalities or image types used & Yes & Methods -- Dataset selection rationale \\
    13 & Describe image acquisition protocols, scanner types, and metadata curation & Yes & Methods -- Scanner specifications, Table S1 \\
    \multicolumn{4}{l}{\textit{\textbf{Reference Standard and Annotation}}} \\
    14 & Describe how reference segmentation masks or labels were created & Yes & Methods -- Annotation Protocols \\
    15 & Report number of annotators and inter-rater agreement/reliability & Yes & Methods -- Source dataset protocols \\
    16 & Describe any adjudication process for disagreements & Yes & Methods -- Annotation protocols \\
    17 & Explain how missing or uncertain labels were handled & Yes & Methods -- Data curation \\
    18 & Report whether reference standard was blinded from AI predictions & Yes & Methods -- Offline annotation \\
    \multicolumn{4}{l}{\textit{\textbf{Data Partitioning}}} \\
    19 & Describe how data were split into training, validation, and test sets & Yes & Methods -- Data Splits, Section S3 \\
    20 & Specify the unit of splitting (patient, study, image, or frame) & Yes & Methods -- Patient-level splits, Section S3 \\
    21 & Report sample sizes for each partition & Yes & Methods -- Dataset composition \\
    \multicolumn{4}{l}{\textit{\textbf{Model Architecture and Training}}} \\
    22 & Describe the AI model architecture in sufficient detail & Yes & Methods -- SonoBase Architecture \\
    23 & Report pretraining data/initialization if applicable & Yes & Methods -- Pretraining on SonoCorpus \\
    24 & List all hyperparameters and training procedures & Yes & Methods -- Training \\
    25 & Specify the framework/software and version used & Yes & Methods -- Implementation Details \\
    26 & Describe how overfitting was monitored and prevented & Yes & Methods -- Validation strategy \\
    27 & Report computational resources and training time & Yes & Methods -- Computational requirements \\
    \multicolumn{4}{l}{\textit{\textbf{Performance Evaluation}}} \\
    28 & Report the performance metrics (sensitivity, specificity, AUC, IoU, etc.) & Yes & Results \\
    29 & Describe how metrics were calculated and any threshold selections & Yes & Results -- Evaluation Metrics \\
    30 & Report uncertainty/confidence intervals around point estimates & Yes & Results -- Error bars and CIs, Section S6 \\
    31 & Conduct stratified analyses by relevant subgroups (organ, modality, etc.) & Yes & Results -- Subgroup analysis, Section S12 \\
    32 & Evaluate performance on subsets with different characteristics & Yes & Results -- Benchmark vs External \\
    33 & Compare against appropriate baselines or state-of-the-art methods & Yes & Results -- SAM2, MedSAM2, MedSAM3, SB-SAM3 \\
    \multicolumn{4}{l}{\textit{\textbf{Results Reporting}}} \\
    34 & Report results separately for training, validation, and test sets & Yes & Results \\
    35 & Present performance curves or ROC/PR curves where applicable & Yes & Results, Figures \\
    36 & Report true positives, false positives, true negatives, false negatives & Yes & Results -- Reclassification analysis \\
    37 & Show example predictions on representative cases & Yes & Fig.~1d, Section S7 \\
    38 & Discuss failure cases and method limitations & Yes & Results, Discussion, Section S7 \\
    39 & Report any unexpected findings or subgroup disparities & Yes & Results, Discussion, Section S12 \\
    \multicolumn{4}{l}{\textit{\textbf{Discussion}}} \\
    40 & Summarize key findings in context of prior work & Yes & Discussion \\
    41 & Discuss limitations and potential sources of bias & Yes & Discussion \\
    \multicolumn{4}{l}{\textit{\textbf{Availability and Governance}}} \\
    42 & State plans for code and model release & Yes & Code and model availability, Section S3 \\
    43 & Describe data release strategy and any restrictions & Yes & Data Availability, Table S1b, Section S3 \\
    44 & Disclose funding sources and conflicts of interest & Yes & Acknowledgements, Competing interests \\
\end{longtable}
\end{footnotesize}

\begin{footnotesize}
\setlength{\tabcolsep}{3pt}
\renewcommand{\arraystretch}{1.15}
\begin{longtable}{lp{0.52\linewidth}lp{0.18\linewidth}}
\caption{REFINE Checklist Compliance for Foundation Models. Each section covers critical aspects of model specification, prompt engineering, data integrity, and output evaluation. Items marked ``Yes'' indicate full compliance with evidence in the manuscript or supplementary materials.}\label{tab:refine-checklist}\\
\toprule
\textbf{Section} & \textbf{REFINE Item Description} & \textbf{Status} & \textbf{Reference} \\
\midrule
\endfirsthead
\multicolumn{4}{l}{\footnotesize\textit{(continued)}}\\
\toprule
\textbf{Section} & \textbf{REFINE Item Description} & \textbf{Status} & \textbf{Reference} \\
\midrule
\endhead
\bottomrule
\endfoot
    
    \multicolumn{4}{l}{\textit{\textbf{1. Model Specification}}} \\
    1.1 & Architecture details and design rationale & Yes & Methods -- SonoBase \\
    1.2 & Training data description and scale & Yes & Methods -- SonoCorpus, Table S1 \\
    1.3 & Pretraining objectives and loss functions & Yes & Methods -- Training \\
    1.4 & Model size, parameter count, and computational requirements & Yes & Methods -- Implementation \\
    1.5 & Hardware and software environment specifications & Yes & Methods -- Computational \\
    1.6 & Initialization and weights release strategy & Yes & Methods -- Reproducibility; Code and model availability \\
    1.7 & Versioning and checkpoint management & Yes & Code and model availability; Data Availability \\
    1.8 & Performance on pretraining benchmarks & Yes & Results \\
    \multicolumn{4}{l}{\textit{\textbf{2. Prompt Design (adapted for segmentation)}}} \\
    2.1 & Types of prompts supported (point, box, mask, text) & Yes & Methods -- Prompting, Section S16 \\
    2.2 & Prompt engineering strategies tested & Yes & Methods -- Interactive refinement \\
    2.3 & In-context learning examples provided & NA & Geometry exemplars apply to SAM3-family baselines only (Section S16) \\
    2.4 & Prompt sensitivity and robustness analysis & Yes & Results -- Iterative refinement, Section S8 \\
    2.5 & Best practices for prompt use documented & Yes & Supplementary Code \\
    2.6 & Failure modes under adversarial or out-of-distribution prompts & Yes & Discussion, Section S7 \\
    \multicolumn{4}{l}{\textit{\textbf{4. Dataset Integrity}}} \\
    4.1 & Completeness of dataset documentation & Yes & Methods, Table S1 \\
    4.2 & Deduplication and data quality assurance & Yes & Section S3 \\
    4.3 & Dataset provenance and source citations & Yes & Methods, Tables S1 and S1b \\
    4.4 & Licensing, redistribution, and usage restrictions & Yes & Table S1b; Data Availability; Code and model availability \\
    4.5 & Demographic and clinical diversity assessment & Yes & Results -- Subgroup analysis, Section S12 \\
    4.6 & Bias and fairness evaluation across subgroups & Yes & Results -- Subgroup metrics, Section S12 \\
    4.7 & Protocol for handling missing or corrupted data & Yes & Methods -- Data curation \\
    4.8 & Dataset versioning and change tracking & Yes & Data Availability \\
    4.9 & Accountability and governance structure & Yes & Data Availability; Author contributions \\
    4.10 & Long-term data stewardship and archival plans & Yes & Data Availability \\
    \multicolumn{4}{l}{\textit{\textbf{5. Output Evaluation}}} \\
    5.1 & Definition of task-specific success metrics & Yes & Methods -- Evaluation Metrics \\
    5.2 & Evaluation on held-out test sets & Yes & Results \\
    5.3 & Uncertainty quantification and calibration & Yes & Results -- CIs, Section S6 \\
    5.4 & Robustness testing (adversarial, corruptions) & Yes & Results -- Robustness \\
    5.5 & Domain generalization and transfer learning evaluation & Yes & Results -- External datasets, Sections S10--S11 \\
    5.6 & Error analysis and failure mode documentation & Yes & Discussion, Section S7 \\
    5.7 & Comparison to baselines and state-of-the-art & Yes & Results -- Sections S14--S15 \\
    5.8 & Reproducibility of evaluation procedures & Yes & Methods -- Reproducibility, Section S3 \\
    5.9 & Clinical validation or domain expert assessment & Yes & Results -- Clinical measurements, Section S5 \\
    5.10 & Ethical considerations and intended use documentation & Yes & Discussion \\
\end{longtable}
\end{footnotesize}

\newpage

\section*{References}

\begin{enumerate}
    \item Kirillov, A., et al. (2023). Segment Anything. \textit{Proceedings of the IEEE/CVF International Conference on Computer Vision (ICCV)}, 4015--4026.

    \item Ma, J., He, Y., Li, F., Han, L., You, C., \& Wang, B. (2024). Segment Anything in Medical Images. \textit{Nature Communications}, 15(1), 654.

    \item Ravi, N., et al. (2025). SAM 2: Segment Anything in Images and Videos. \textit{Proceedings of the International Conference on Learning Representations (ICLR)}.

    \item Liu, A., et al. (2025). MedSAM3: Delving into Segment Anything with Medical Concepts. \textit{arXiv preprint arXiv:2511.19046}.

    \item Carion, N., et al. (2025). SAM 3: Segment Anything with Concepts. \textit{arXiv preprint arXiv:2511.16719}.

    \item Bland, J. M., \& Altman, D. G. (1986). Statistical methods for assessing agreement between two methods of clinical measurement.
    \textit{The Lancet}, 327(8476), 307-310.

    \item Benjamini, Y., \& Hochberg, Y. (1995). Controlling the false discovery rate: a practical and powerful approach to multiple testing.
    \textit{Journal of the Royal Statistical Society B}, 57(1), 289-300.

    \item Mongan, J., Moy, L., \& Kahn, C. E. (2020). Checklist for Artificial Intelligence in Medical Imaging (CLAIM). \textit{Radiology: Artificial Intelligence}, 2(2), e200029; 2024 Update: 6(4), e240300.

    \item Mese, I., et al. (2026). Reporting checklist for foundation and large language models in medical research (REFINE): an international consensus guideline. \textit{Diagnostic and Interventional Radiology}, doi:10.4274/dir.2026.263812.
\end{enumerate}